\pdfoutput=1
\documentclass[11pt, letterpaper, logo, onecolumn, copyright, numbering]{paper_improved}

\usepackage{soul}
\usepackage{xcolor}

\usepackage{booktabs}
\usepackage{array}
\usepackage{colortbl}
\usepackage{xcolor}
\usepackage{caption}
\usepackage{tabularray}
\usepackage[utf8]{inputenc} 
\usepackage[T1]{fontenc}    
\usepackage{url}            
\usepackage{booktabs}       
\usepackage{amsfonts}       
\usepackage{nicefrac}       
\usepackage{microtype}      
\usepackage{xcolor}         
\usepackage{graphicx}

\usepackage{natbib}
\usepackage{verbatim}
\usepackage{inconsolata}
\usepackage{tabularx}
\usepackage{float}
\usepackage{multirow}
\usepackage{listings}
\usepackage{xcolor}
\usepackage{subfigure}
\usepackage{subcaption}
\usepackage{amsmath}
\usepackage{amssymb}
\usepackage{array}
\usepackage{svg}
\usepackage{wrapfig}

\definecolor{lightgray}{rgb}{0.95, 0.95, 0.95}
\definecolor{darkgray}{rgb}{0.4, 0.4, 0.4}
\definecolor{backcolour}{rgb}{0.95,0.95,0.92}
\definecolor{myblue}{rgb}{0.2, 0.4, 0.8} 
\definecolor{mygreen}{rgb}{0.2, 0.6, 0.2} 

\usepackage{CJKutf8}
\usepackage{soul}
\usepackage{url}
\usepackage[utf8]{inputenc}
\usepackage{caption}
\usepackage{graphicx}
\usepackage{xcolor}

\usepackage{booktabs}
\usepackage{latexsym}
\usepackage{amsmath}
\usepackage{subfigure}
\usepackage{microtype}
\usepackage[switch]{lineno}
\usepackage[most]{tcolorbox}
\usepackage{alltt}

\usepackage{multirow}
\usepackage{makecell}
\usepackage{hhline}
\usepackage{booktabs}
\usepackage{bm}
\usepackage{amssymb}
\usepackage{amsmath}
\usepackage{url}
\usepackage{graphicx}
\usepackage{float}
\usepackage{subfigure}
\usepackage{paralist}
\usepackage[ruled, lined, linesnumbered, commentsnumbered, longend]{algorithm2e}
\usepackage{xcolor}
\usepackage{colortbl}
\usepackage{pifont}
\usepackage{tabularx}
\usepackage[T2A,LAE,T1]{fontenc}
\usepackage{CJKutf8}
\usepackage{enumitem}
\usepackage{longtable}
\setlist[itemize]{noitemsep, topsep=0pt}
\usepackage{hyperref}

\definecolor{LightCyan}{rgb}{0.88,1,1}
\newcommand{\boldtitle}[1]{\noindent\textbf{#1}\xspace\xspace}

\newcommand{\cmark}{\ding{51}}
\newcommand{\xmark}{\ding{55}}

\newtcbox{\hlprimarytab}{on line, rounded corners, box align=base, colback=white!10,colframe=white,size=fbox,arc=3pt, before upper=\strut, top=-2pt, bottom=-4pt, left=-2pt, right=-2pt, boxrule=0pt}

\newcommand{\huggingface}{\raisebox{-1.5pt}{\includegraphics[height=1.05em]{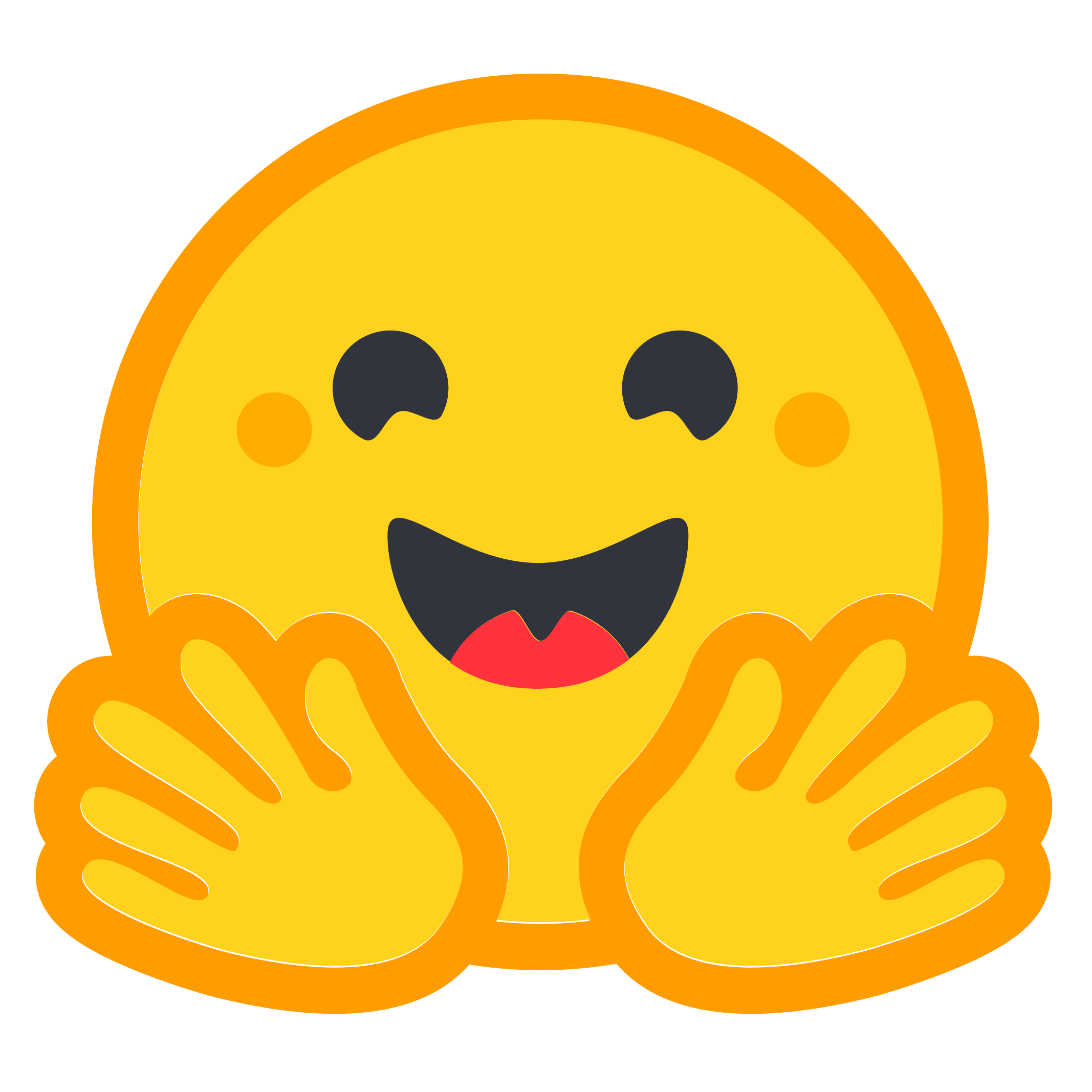}}\xspace}
\newcommand{\github}{\raisebox{-1.5pt}{\includegraphics[height=1.05em]{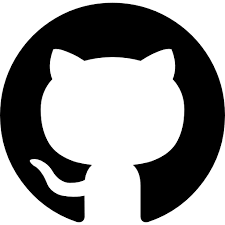}}\xspace}
\newcommand{\modelmetadata}[2][]{{\small {\sffamily \bfseries #1} #2}\par}

\newcommand{\marcomoe}{Marco-MoE\xspace}
\newcommand{\marcosmall}{Marco-Nano-Base\xspace}
\newcommand{\marcomedium}{Marco-Mini-Base\xspace}
\newcommand{\marcomediumglobal}{Marco-Mini-Global-Base\xspace}
\newcommand{\marcosmallins}{Marco-Nano-Instruct\xspace}
\newcommand{\marcomediumins}{Marco-Mini-Instruct\xspace}

\def\eg{{e.g.,}\xspace}
\def\ie{{i.e.,}\xspace}
\def\versus{{\em v.s.}\xspace}

\definecolor{lightblue}{HTML}{bdd6fb}
\definecolor{boxgray}{gray}{0.9}
\definecolor{bgyellow}{HTML}{fcebde}
\definecolor{bgred}{HTML}{d77470}
\definecolor{bggrey}{HTML}{dcc0e5}

\definecolor{forestgreen}{rgb}{0.13, 0.55, 0.13}

\tcbset{
    aibox/.style={
        colback=white, 
        colframe=blue!75!black, 
        colbacktitle=blue!85!black, 
        coltitle=white, 
        fonttitle=\bfseries,
        enhanced,  
        drop shadow=black!50!white,  
        boxrule=1pt,  
        boxsep=10pt,  
        left=10pt,  
        right=10pt,  
        top=6pt,  
        bottom=6pt,  
        title code={\path[tcb fill frame] ([xshift=-10pt]frame.west) -- (frame.north west) -- (frame.north east) -- ([xshift=10pt]frame.east) -- cycle;},  
        attach boxed title to top left={xshift=10pt, yshift*=-\tcboxedtitleheight/2},
        boxed title style={boxrule=0pt, frame code={}}  
    }
}

\newtcolorbox{AIbox}[2][]{aibox, title=#2, #1}

\let\cite\citep

\title{\marcomoe: Open Multilingual Mixture-of-Expert Language Models with Efficient Upcycling}

\author[*,1]{Fan Jiang, Yu Zhao, Chenyang Lyu, Tianqi Shi, Yichao Du, Feihu Jiang, Longyue Wang\textsuperscript{*}, Weihua Luo \\

\textbf{Alibaba International Digital Commerce} \\
\textsuperscript{*} Corresponding Author: wanglongyue.wly@alibaba-inc.com 
}

\begin{abstract}
We present \marcomoe, a suite of fully open multilingual sparse Mixture-of-Experts (MoE) models. \marcomoe features a highly sparse design in which only around 5\% of the total parameters are activated per input token. This extreme sparsity, combined with upcycling from dense models, enables efficient pre-training on 5T tokens. Our models surpass similarly-sized competitors on English and multilingual benchmarks, achieving a best-in-class performance-to-compute ratio. We further post-train these models to create \marcomoe-\textsc{Instruct} variants, which surpass the performance of competing models possessing $3$--$14\times$ more activated parameters. Our analysis reveals that \marcomoe learns structured expert activation patterns shared across related languages, while maintaining highly specialized utilization for linguistically isolated ones. We further show that \marcomoe allows for scalable language expansion without the interference typical of dense models. To support the community, we disclose our full training datasets, recipes, and model weights.
\end{abstract}

\begin{document}

\maketitle

\modelmetadata[\quad\github GitHub:]{
\href{https://github.com/AIDC-AI/Marco-LLM}{\texttt{Marco-LLM}}
}

\modelmetadata[\quad\huggingface Marco-MoE Base:]{
\href{https://huggingface.co/AIDC-AI/Marco-Nano-Base}{\texttt{Marco-Nano-Base}}\quad
\href{https://huggingface.co/AIDC-AI/Marco-Mini-Base}{\texttt{Marco-Mini-Base}}\quad
\href{https://huggingface.co/AIDC-AI/Marco-Mini-Global-Base}{\texttt{Marco-Mini-Global-Base}}
}

\modelmetadata[\quad\huggingface Marco-MoE Instruct:]{
\href{https://huggingface.co/AIDC-AI/Marco-Nano-Instruct}{\texttt{Marco-Nano-Instruct}}\quad
\href{https://huggingface.co/AIDC-AI/Marco-Mini-Instruct}{\texttt{Marco-Mini-Instruct}}
}

\begin{figure*}[t]
    \centering
    \includegraphics[width=\textwidth]{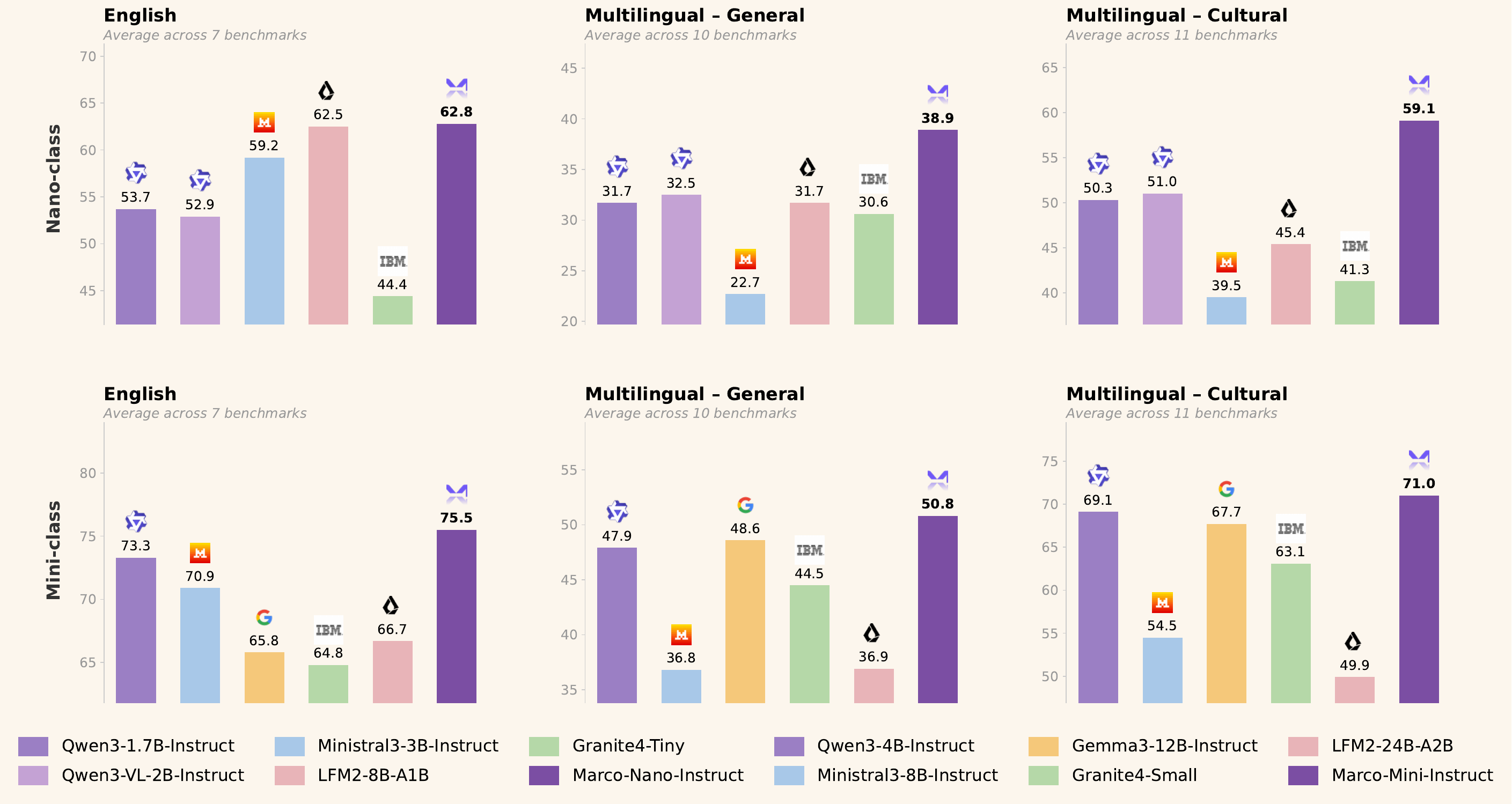}
    \caption{Performance comparison of \marcomoe Instruct models against open-weight Instruct models across English, General Multilingual, and Cultural \& Regional Multilingual benchmarks. Models are grouped into \textbf{Nano-class} (top, $\leq$ 3B activated parameters) and \textbf{Mini-class} (bottom, $\leq$ 12B activated parameters). Despite activating only \textbf{0.6B} and \textbf{0.86B} parameters respectively, \marcosmallins and \marcomediumins consistently match or outperform models with $3$--$14\times$ more activated parameters.}
    \label{fig:instruct_performance}
\end{figure*}
\section{Introduction}
The rapid advancement of Large Language Models (LLMs) has transformed natural language understanding, yet achieving a balance between extensive linguistic coverage and high per-language proficiency remains a critical hurdle, often referred to as the "curse of multilinguality"~\citep{conneau-etal-2020-unsupervised}. This challenge stems from the fact that expanding a model’s language coverage within a fixed parameter budget often degrades performance on individual languages due to capacity bottlenecks and cross-lingual interference. While leading models such as Qwen3~\citep{yang2025qwen3technicalreport} and Gemma3~\citep{gemmateam2025gemma3technicalreport} attempt to bypass these constraints through massive-scale pre-training on up to 36T tokens or sophisticated strong-to-weak distillation~\citep{comanici2025gemini25pushingfrontier}, such strategies remain computationally prohibitive for most pre-training scenarios. Meanwhile, existing small-scale multilingual models, such as Tiny-Aya~\citep{salamanca2026tinyayabridgingscale}, typically use dense architectures that struggle to reconcile broad linguistic breadth with deep task proficiency, as they lack the inherent architectural flexibility to scale effectively across diverse language families without compromising performance.

Mixture-of-Experts (MoE) architectures~\citep{shazeer2017outrageouslylargeneuralnetworks} offer a promising solution to this capacity bottleneck by leveraging conditional computation, which enhances model capacity without increasing the number of activated parameters. However, the computational overhead of training large-scale MoE models from scratch has spurred the emergence of MoE Upcycling~\citep{komatsuzaki2023sparse}, a paradigm that initializes sparse models using pre-trained dense checkpoints. Yet current approaches predominantly rely on coarse-grained expert replication, in which entire feed-forward networks (FFNs) of a dense transformer are replicated. This monolithic duplication deviates from existing wisdom that uses fine-grained experts~\citep{dai-etal-2024-deepseekmoe}, thereby impeding the emergence of specialized representations necessary for nuanced multilingual reasoning.

To bridge this gap, we introduce \marcomoe, a family of compact, highly sparse multilingual MoE models designed to break these traditional trade-offs in multilinguality. Our work distinguishes itself from previous efforts in three fundamental ways:
\begin{compactenum}
    \item \textbf{First Sparse Multilingual Upcycling}: To the best of our knowledge, this is the first work to leverage the MoE upcycling paradigm specifically to optimize multilingual performance in compact model sizes. By repurposing pre-trained dense representations into a fine-grained MoE framework, we significantly reduce computational overhead while enhancing model capacity.
    \item \textbf{Fine-Grained Expert Specialization}: Unlike conventional coarse-grained upcycling methods that replicate entire FFN blocks, we employ a sub-matrix splitting technique to initialize a large number of fine-grained experts. This approach, combined with Drop-Upcycling~\citep{nakamura2025dropupcycling}, catalyzes specialized expert convergence and minimizes the redundancy bottlenecks typical of monolithic replication.
    \item \textbf{Full Transparency and Openness}: In contrast to many models that keep their pre-training data and recipes proprietary, \marcomoe is fully transparent. We disclose the entirety of our pre-training datasets, our data synthesis methodologies, and our rigorous four-stage pre-training curriculum spanning 5.1T tokens. By open-sourcing our models, data, and training recipes, \marcomoe sets a new standard for accessible, high-performance multilingual LLM development.
\end{compactenum}

\begin{figure}[t]
    \setlength{\abovecaptionskip}{-0.005cm}
    \centering
    \subfigure{
        \label{fig:performance_vs_flops}
        \includegraphics[width=0.48\textwidth]{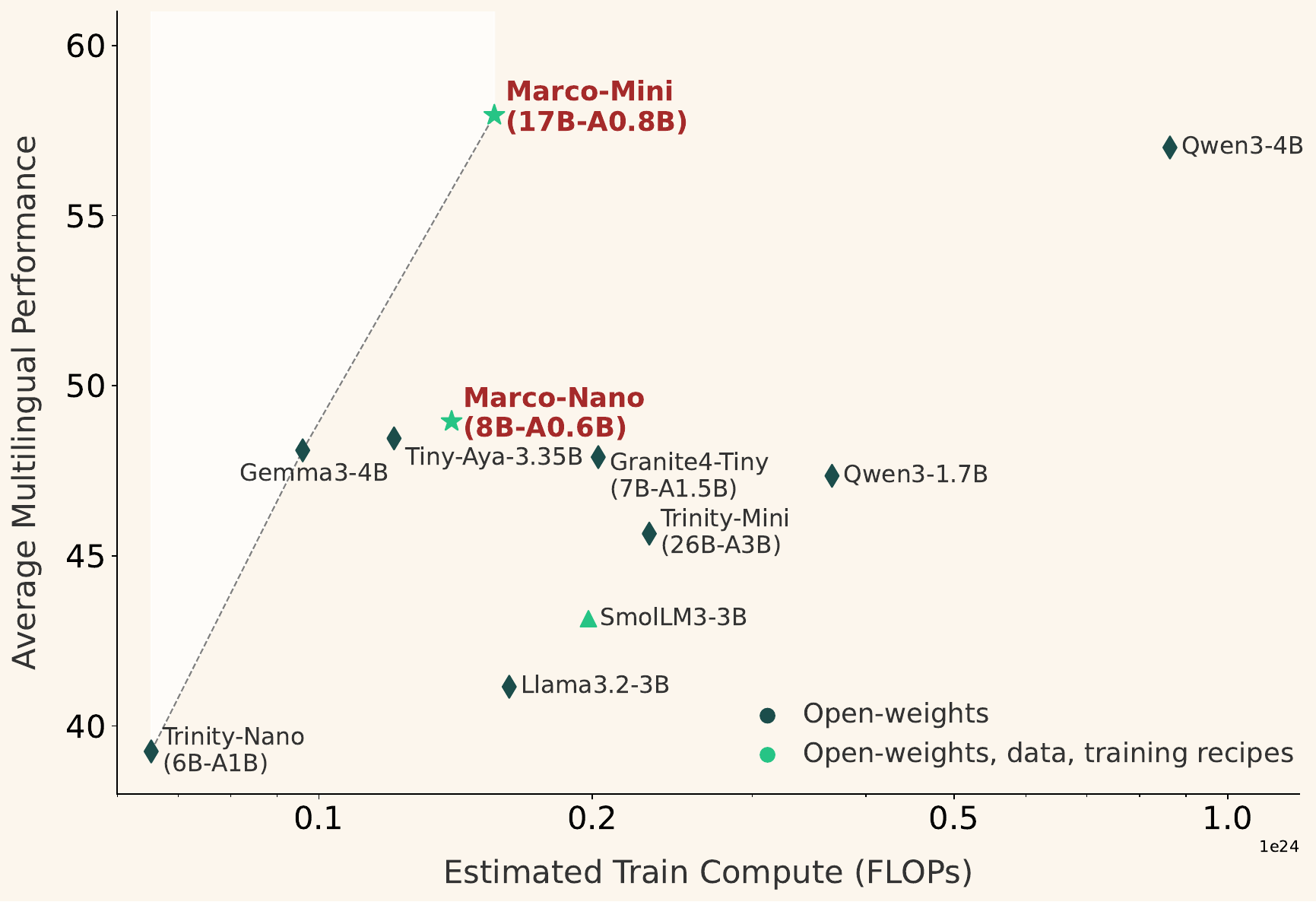}
    }
    \subfigure{
        \label{fig:multilingual_vs_English}
        \includegraphics[width=0.48\textwidth]{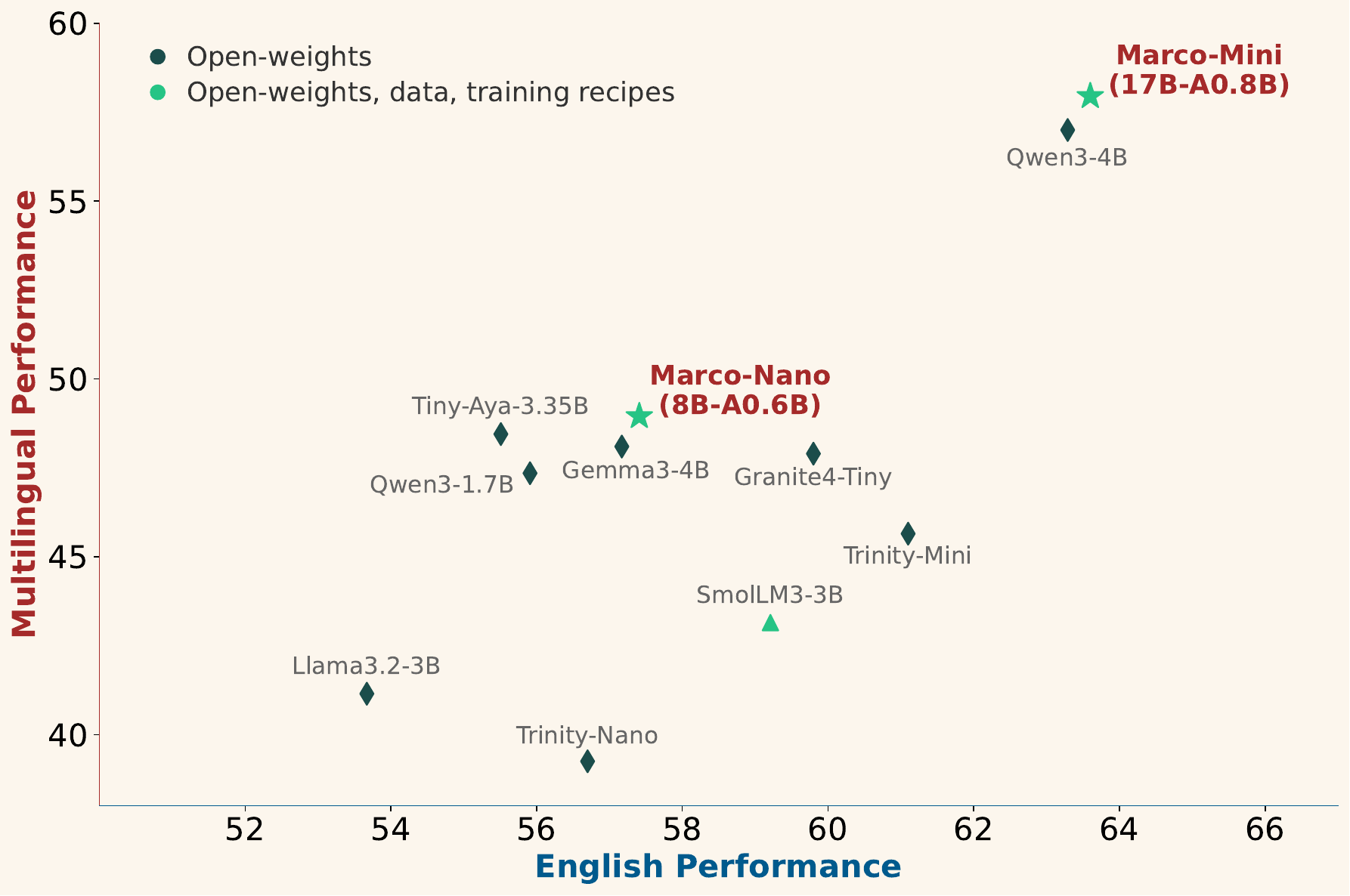}
    }
    \caption{Comparison of performance and efficiency. Our \marcomoe base models demonstrate a superior performance-to-compute ratio in (a) and set the state-of-the-art for simultaneous proficiency in both English and multilingual capabilities in (b).}
    \label{fig:main_figure}
\end{figure}
\begin{figure*}[t]
    \setlength{\belowcaptionskip}{-0.3cm}
    \centering
    \includegraphics[width=\textwidth]{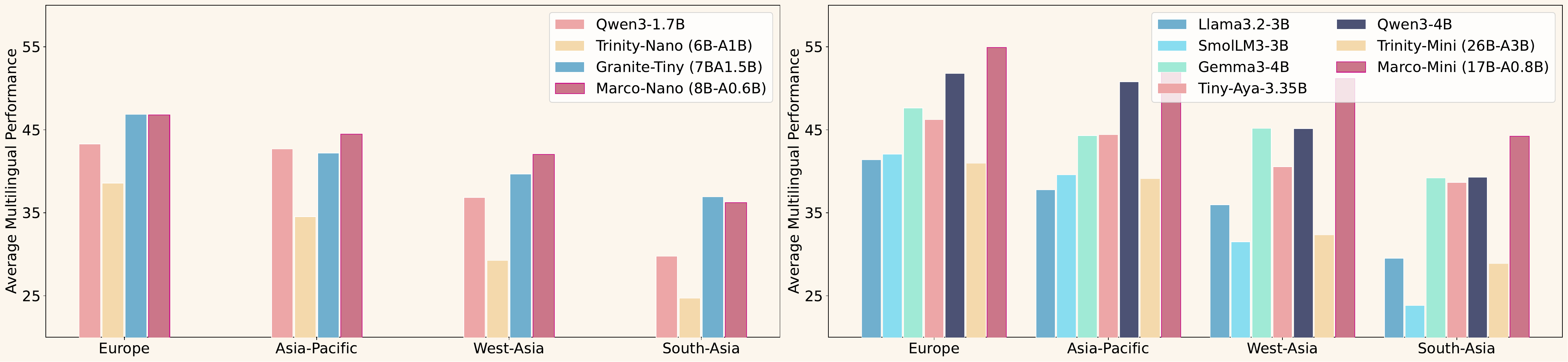}
    \caption{Multilingual benchmark performance of base models across geographic regions.}
    \label{fig:result_by_region}
\end{figure*}

Our empirical results demonstrate that \marcomedium (0.86B activated parameters) and \marcosmall (0.6B activated parameters) achieve a superior performance-to-compute ratio (Figure~\ref{fig:main_figure}). Notably, our models set a new state-of-the-art for simultaneous proficiency in both English and multilingual capabilities, particularly excelling in long-tail and low-resource languages where the performance gap in specific geographic regions (\eg West and South Asia) over dense counterparts widens significantly (Figure~\ref{fig:result_by_region}).

We further adapt the \marcomoe base models through a two-stage post-training pipeline to develop the Instruct variants. The pipeline starts with Supervised Fine-Tuning (SFT) on high-quality instruction-response pairs, followed by On-Policy Distillation (OPD)~\citep{lu2025onpolicydistillation,yang2025qwen3technicalreport} to efficiently transfer knowledge from high-capacity teacher models. 
As shown in Figure~\ref{fig:instruct_performance}, \marcosmallins establishes a highly efficient frontier within the compact model class, surpassing LFM2-8B-A1B-Instruct in English proficiency despite activating 2.5$\times$ fewer parameters. Furthermore, it consistently outperforms larger dense and sparse baselines across general and regional multilingual benchmarks. In the larger Mini-class, \marcomediumins achieves the highest overall scores in both general and cultural multilingual tasks with only 0.86B activated parameters, representing a fraction of the computational footprint required by competing models in the 4B to 32B parameter range.

\section{Model Architecture}

\begin{table}
\setlength{\tabcolsep}{20pt}
\footnotesize
\centering
\resizebox{0.8\linewidth}{!}{
\begin{tabular}{l|c|c}
\toprule
    \bf Model & \bf \marcosmall & \bf \marcomedium \\ 
    \midrule
    Num Layers & 28 & 28 \\
    Model Dimension & 1024 & 1024 \\ 
    FFN Intermediate Dimension & 3072 & 3072 \\
    \midrule
    Q-heads & 16 & 16 \\
    KV-heads & 8 & 8 \\
    Head Dimension & 128 & 128 \\
    \midrule
    Expert Dimension & 384 & 768 \\
    Total Experts & 232 & 256 \\
    Number of Activated Experts & 8 & 8 \\
    \midrule
    Tie Embedding & True & True \\
    \bottomrule
\end{tabular}}
\caption{Summary of the \marcomoe LLM architectures.}
\label{tab:marco_architecture}
\end{table}

\subsection{Architecture Overview}
\marcomoe LLM leverages a decoder-only Transformer architecture \citep{vaswani2023attentionneed}, substituting conventional Feed-Forward Network (FFN) layers with sparse Mixture-of-Experts (MoE) layers \citep{shazeer2017outrageouslylargeneuralnetworks}. This substitution facilitates enhanced model capacity and accuracy while significantly reducing the number of activated parameters. To optimize performance, we implement a granular MoE architecture following \citet{dai-etal-2024-deepseekmoe}. Other architectural refinements, including Grouped-Query Attention (GQA) \citep{ainslie-etal-2023-gqa}, RMSNorm \citep{zhang2019rootmeansquarelayer}, SwiGLU activation \citep{dauphin2017languagemodelinggatedconvolutional}, and Rotary Positional Embeddings (RoPE) \citep{su2023roformerenhancedtransformerrotary}, are aligned with the Qwen framework \citep{yang2025qwen3technicalreport}. Detailed configurations are summarized in Table~\ref{tab:marco_architecture}. Specifically, the \marcosmall variant comprises 8B total parameters with 0.6B activated (7.5\% active ratio), while the \marcomedium Base contains 17.3B total parameters with 0.86B activated (5\% active ratio). Both of them are upcycled from the Qwen3-0.6B-Base model~\citep{yang2025qwen3technicalreport} using the method detailed in~\S\ref{sec:upcycle}.

\begin{figure*}[t]
    \centering
    \includegraphics[width=\textwidth]{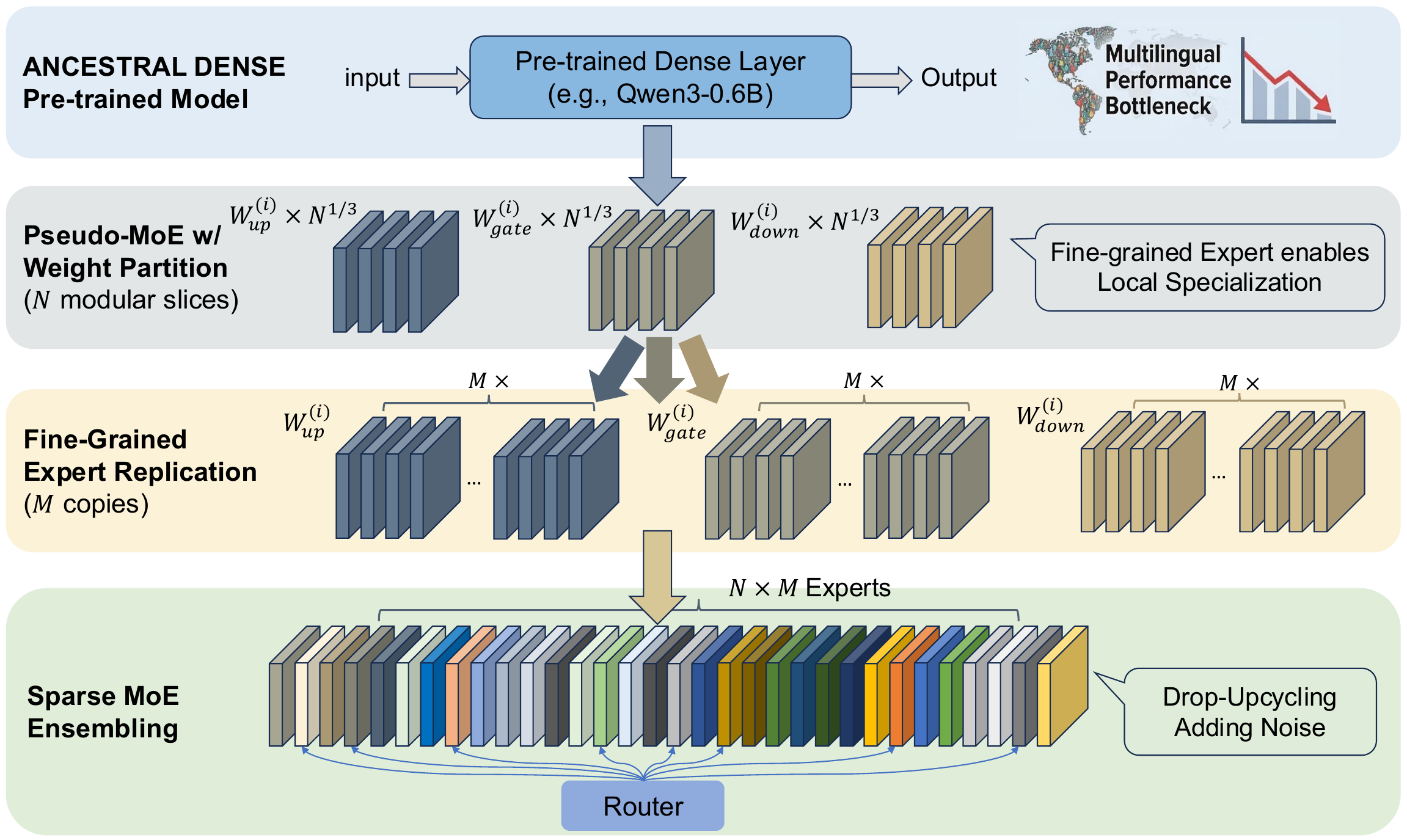}
    \caption{The overview of our method on upcycling dense models to MoE models with fine-grained experts.}
    \label{fig:moe_fine_grained_drop_upcycle}
\end{figure*}

\subsection{Upcycle From Dense Model}\label{sec:upcycle}
Upcycling~\citep{komatsuzaki2023sparse} serves as an efficient paradigm for initializing MoE models by repurposing pre-trained dense models. By leveraging the rich representations already encoded within a dense model, this approach substantially reduces computational overhead while achieving performance comparable to that of MoE models trained from scratch. However, naive upcycling, the practice of simply replicating the FFN layers of a dense model to initialize experts, inherently discourages expert diversification. Because identical initializations provide no initial gradient variance between experts, they struggle to specialize during the subsequent training phase. This lack of heterogeneity prevents the MoE architecture from reaching its full capacity, often leading to sub-optimal convergence rates over extended training periods.

To mitigate these issues and promote expert specialization, recent methods introduce random noise into the FFN replicas~\citep{yang2024qwen2technicalreport}, ensuring each expert begins from a distinct point in the weight space. In this work, we adopt Drop-Upcycling~\citep{nakamura2025dropupcycling}, a strategy centered on the selective re-initialization of expert parameters. Specifically, indices are randomly sampled along the intermediate dimensions of the dense model’s FFNs, and the weights along the corresponding column or row axes are dropped. These parameters are then re-initialized using a Gaussian distribution, with the mean and variance ($\mu$ and $\sigma^2$) derived directly from the statistics of those dropped weights. In this work, we adopt this upcycling strategy, adapting it to fit within the fine-grained MoE architecture, as shown in Figure~\ref{fig:moe_fine_grained_drop_upcycle}.

Given the transition to a fine-grained MoE architecture, a structural discrepancy arises between the expert dimension $H_{expert}$ and the intermediate FFN dimension $H_{dense}$ of the ancestral dense model. To facilitate seamless parameter inheritance, we first reconfigure the dense model into a pseudo-MoE framework where all experts remain active. In a standard dense Transformer FFN, the computation is defined as: $Y=(\operatorname{Act}(XW_{gate})XW_{up})W_{down}$, where the weight matrices $W_{up} \in \mathbb{R}^{D \times H_{dense}}$, $W_{gate} \in \mathbb{R}^{D \times H_{dense}}$, and $W_{down} \in \mathbb{R}^{H_{dense} \times D}$ represent the up-projection, gate-projection, and down-projection layers, respectively. To initialize the MoE layers, we partition these weights into $N = H_{dense} / H_{expert}$ slices along the intermediate dimension:
\begin{gather}
    W_{up} \rightarrow [W_{up}^{(1)}, W_{up}^{(2)}, ..., W_{up}^{(N)}]\quad W_{gate} \rightarrow [W_{gate}^{(1)}, W_{gate}^{(2)}, ..., W_{gate}^{(N)}]\quad W_{down} \rightarrow [W_{down}^{(1)}, W_{down}^{(2)}, ..., W_{down}^{(N)}] \nonumber
\end{gather}
Following this partitioning, the $i$-th expert $E_i$ is initialized using the corresponding weight pair $(W_{up}^{(i)}, W_{gate}^{(i)}, W_{down}^{(i)})$.

\begin{figure*}[t]
    \setlength{\abovecaptionskip}{-0.005cm}
    \centering
    \includegraphics[width=0.8\textwidth]{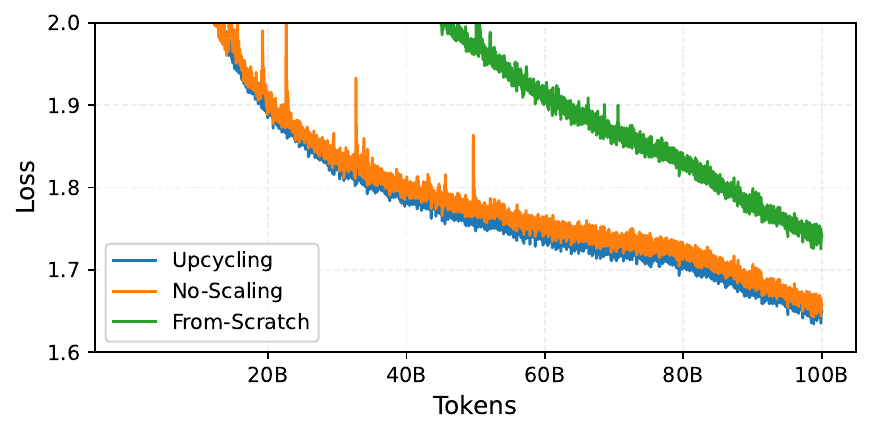}
    \caption{Ablation on weight scaling and upcycling. We train MoE models with 0.8B active, 17B total parameters on 100B tokens. Upcycling significantly speeds up pre-training, leading to faster convergence. Weight scaling on expert parameters helps reduce loss spikes while achieving similar final training loss.}
    \label{fig:upcycling_vs_no_weight_scale_vs_from_scratch}
\end{figure*}

However, a direct weight initialization for each expert via naive slicing introduces a significant scaling disparity. In the dense model, the FFN computes an aggregate output by summing all $N$ constituent slices: $Y = \sum_{i=1}^{N} E_i(x)$. Conversely, a MoE model typically employs a softmax-based gating mechanism where the routing probabilities $p_i$ are constrained such that $\sum p_i = 1$. The resulting output is a weighted combination: $Y_{MoE} = \sum_{i=1}^{N} p_i E_i(x)$. This creates a functional mismatch. While the dense model implicitly assigns a unit coefficient to each slice, the MoE model distributes weight according to $p_i$. Under a uniform routing assumption where $p_i \approx \frac{1}{N}$, the MoE output magnitude is effectively reduced by a factor of $N$ relative to the original dense signal~\citep{he2025upcyclinglargelanguagemodels}. To ensure the Pseudo-MoE is mathematically identical to the source dense model, the MoE output must be scaled by the total number of experts ($N$):
\begin{gather}
    Y_{MoE} = N \times \sum_{i=1}^{N} p_i \times E_i(x) \nonumber
\end{gather}
In practice, rather than altering the model's forward-pass logic, we apply a scaling factor $\lambda=N^{1/3}$ directly to the sliced weight matrices. This approach preserves the numerical integrity of the original dense model while maintaining a consistent architectural implementation. Figure~\ref{fig:upcycling_vs_no_weight_scale_vs_from_scratch} demonstrates that this weight scaling makes training more stable by reducing loss spikes. Finally, we apply upcycling to the pseudo-MoE to expand it to a real MoE model with zero-initialized router weights, together with the Drop-Upcycling technique to promote expert diversification and specialization.

\section{Pre-Training}
\subsection{Pre-training Data}
\marcosmall and \marcomedium are pre-trained on a large corpus of high-quality curated and synthetically-generated data derived from both open-sourced and our curated datasets spanning 29 languages: English, Chinese, Arabic, German, Spanish, French, Korean, Japanese, Portuguese, Turkish, Indonesian, Italian, Dutch, Polish, Russian, Vietnamese, Thai, Bengali, Czech, Hebrew, Ukrainian, Malay, Urdu, Kazakh, Greek, Romanian, Hungarian, Nepali, Azerbaijani.

\subsubsection{High-Quality English Data}
To enhance the general language understanding of \marcomoe, we leverage Nemotron-CC-v2 \citep{nvidia2025nemotronhfamilyaccurateefficient} as the primary source of high-quality English data, specifically utilizing the \textbf{High} and \textbf{High-Synthetic} partitions. This is supplemented by an internal web-crawled English corpus, curated by applying the Fineweb-EDU \citep{lozhkov2024fineweb-edu} classifier and retaining only the top 10\% of documents. Furthermore, to bolster targeted capabilities, we incorporate the \textbf{Diverse question-answer (QA)} pairs from Nemotron-CC-v2. These pairs encompass various formats, including yes/no, open-ended, and multiple-choice questions, centered on factual information within the text.

\subsubsection{Reasoning and Instruction Data}
Our dataset incorporates a diverse array of open-source sources spanning the STEM, Coding, and instruction-following domains. Specifically, we leverage \href{https://huggingface.co/datasets/nvidia/}{Nemotron-Pretraining-SFT-v1}~\citep{nvidia2025nvidianemotronnano2}, \href{https://huggingface.co/datasets/nvidia/Nemotron-Pretraining-Specialized-v1}{Nemotron-Pretraining-Specialized-v1}~\citep{nvidia_nemotron_nano_v3_2025}, \href{https://huggingface.co/datasets/nvidia/Nemotron-CC-Math-v1}{Nemotron-CC-Math-v1}~\citep{mahabadi2025nemotronccmath133billiontokenscalehigh}, \href{https://huggingface.co/datasets/HuggingFaceTB/finemath}{FineMath}~\citep{allal2025smollm2smolgoesbig}, \href{https://huggingface.co/datasets/LLM360/MegaMath}{MegaMath}~\citep{zhou2025megamath}, \href{https://huggingface.co/datasets/open-thoughts/OpenThoughts3-1.2M}{OpenThoughts3-1.2M}~\citep{guha2025openthoughtsdatarecipesreasoning}, and \href{https://huggingface.co/datasets/allenai/dolmino-mix-1124}{FLAN}~\citep{wei2022finetuned, olmo20252olmo2furious}. Collectively, these provide a robust foundation of reasoning-intensive synthetic problems, high-quality domain-specific documents, and diverse instruction data.

\subsubsection{High-Quality Multilingual Data}
\marcomoe is pre-trained on a large amount of high-quality multilingual data sourced from the web and synthesized by using various strategies.

\subsubsubsection{Web-Crawled Data}
We aggregate multilingual web-crawled data for 28 target languages, sourcing from open-source repositories. Our primary resource is Fineweb-2 \citep{penedo2025fineweb}, which provides corpora for over a thousand languages. To ensure data quality, we prioritize the FineWeb2-HQ variant \citep{messmer2025multilingdatacomp}, which utilizes a model-based classifier to filter out low-quality documents for 20 languages. We use this dataset when a high-quality version is available for a language. For the remaining languages that lack high-quality subsets, we employ a rephrasing strategy to mitigate the inherent noise in web-crawled content. Specifically, we employ Qwen3-30B-A3B-Instruct~\citep{yang2025qwen3technicalreport} to generate synthetic data using four prompting templates derived from \citet{su-etal-2025-nemotron} and \citet{maini-etal-2024-rephrasing}: 
\begin{compactenum}
    \item \textbf{Distillation}: Reformulating the text into a concise and clear passage. 
    \item \textbf{Knowledge Extraction}: Isolating informative facts while discarding uninformative content. 
    \item \textbf{Structured Listing}: Organizing key information into a systematic list. 
    \item \textbf{Encyclopedic Paraphrasing}: Generating high-quality, Wikipedia-style prose. 
\end{compactenum}
We specifically curate the training data for Chinese by sourcing from the \href{https://huggingface.co/datasets/opencsg/Fineweb-Edu-Chinese-V2.1}{Fineweb-Edu-Chinese-V2.1} dataset~\citep{yu2025opencsgchinesecorpusseries}. To ensure data integrity, we restrict our selection to the high-quality partition, exclusively utilizing samples with a quality score exceeding 3.

\subsubsubsection{Synthetic Data}
\boldtitle{Multilingual QA Data.}
\citet{nvidia2025nvidianemotronnano2} established that incorporating high-quality multilingual question-answering (QA) data significantly enhances model performance across a variety of downstream tasks. As part of the Nemotron-CC-v2 release, translated versions of the English \textbf{Diverse question-answer (QA)} dataset are provided for 13 languages. To extend coverage to the remaining languages, we leverage the Qwen3-30B-A3B-Instruct model to translate the English Diverse QA data. We employ a curated translation prompt (see Appendix Prompt~\ref{prompt:translation_prompt}) designed to preserve the integrity of non-translatable elements, such as LaTeX syntax and code snippets.

To maximize translation fidelity, we implement a granular processing pipeline wherein documents are segmented into individual lines for discrete translation, followed by a reconstruction phase to restore document coherence. Furthermore, language identification tools are utilized to exclude off-target translations. For low-resource languages where Qwen3-30B-A3B-Instruct exhibits performance degradation, such as Urdu and Kazakh, we apply heuristic filtering mechanisms to remove outputs characterized by repetitive 10-grams.

\begin{table}
\setlength{\tabcolsep}{3pt}
\footnotesize
\centering
\resizebox{\linewidth}{!}{
\begin{tabular}{l|c|ccccccccc}
\toprule
    \bf Model & Avg. & nld\_Latn & ron\_Latn & vie\_Latn & ell\_Grek & zsm\_Latn & tha\_Thai & urd\_Arab & ben\_Beng & kaz\_Cyrl \\ 
    \midrule
    Qwen3-4B-Multilingual & 82.2 & 83.7 & 83.2 & 80.6 & 82.6 & 81.2 & 78.9 & 75.6 & 74.2 & 72.9 \\
    \quad + Multilingual Diverse QA & \bf 83.5 & \bf 85.3 & \bf 85.6 & \bf 83.6 & \bf 84.6 & \bf 83.0 & \bf 80.9 & \bf 78.8 & \bf 77.9 & \bf 77.1 \\
    \bottomrule
\end{tabular}}
\caption{The effects of including multilingual diverse QA data into the pre-training mixture. Performance of selected languages on the BELEBELE benchmark is presented.}
\label{tab:multilingual_qa_ablation}
\end{table}

To evaluate the efficacy of the curated dataset, we perform an ablation study employing an annealing phase of 100B tokens, following the methodology of \citet{olmo20252olmo2furious}. Specifically, we continuously pre-train a Qwen3-4B base model checkpoint for an additional 100B tokens. In the experimental group, the data mixture consists of 50\% translated diverse QA data and 50\% standard web-crawled data. To ensure a fair comparison, we evaluate this against a baseline variant pre-trained for the same duration solely on the default web mixture. Performance is measured via the BELEBELE benchmark \citep{bandarkar-etal-2024-belebele}, with results detailed in Table~\ref{tab:multilingual_qa_ablation}. The integration of synthesized multilingual QA pairs yields consistent improvements, with an average gain of 1.4\% across 29 languages. Notably, these performance gains are most pronounced in low-resource languages, including Urdu, Bengali, and Kazakh.

\begin{table}
\setlength{\tabcolsep}{3pt}
\footnotesize
\centering
\resizebox{\linewidth}{!}{
\begin{tabular}{l|c|ccccccccc}
    \toprule
    \multicolumn{11}{c}{\bf\emph{GlobalMMLU}} \\
    \midrule
    \bf Model & Avg. & zho\_Hans & deu\_Latn & rus\_Cyrl & nld\_Latn & ces\_Latn & ukr\_Cyrl & tur\_Latn & heb\_Hebr & ben\_Beng \\ 
    \midrule
    Qwen3-4B-Multilingual & 64.2 & 65.9 & 65.6 & 65.8 & 66.0 & 63.7 & 62.9 & 60.9 & 57.5 & 55.7 \\
    \midrule
    \quad + Multilingual STEM & \bf 65.6 & \bf 66.9 & \bf 67.1 & \bf 67.0 & \bf 67.3 & \bf 65.5 & \bf 64.8 & \bf 63.3 & \bf 60.3 & \bf 59.1 \\
    \bottomrule
\end{tabular}}
\setlength{\tabcolsep}{3pt}
\resizebox{\linewidth}{!}{
\begin{tabular}{l|c|cccccccccc}
    \toprule
    \multicolumn{11}{c}{\bf\emph{MMMLU}} \\
    \midrule
    \bf Model & Avg. & arb\_Arab & ben\_Beng & deu\_Latn & ind\_Latn & ita\_Latn & jpn\_Jpan & kor\_Hang & por\_Latn & zho\_Hans \\ 
    \midrule
    Qwen3-4B-Multilingual & 60.7 & 57.4 & 50.8 & 62.7 & 64.1 & \bf 63.5 & 58.7 & 57.2 & 64.1 & 63.2  \\
    \quad + Multilingual STEM & \bf 62.2 & \bf 60.5 & \bf 54.9 & \bf 65.4 & \bf 64.5 & 62.4 & \bf 60.6 & \bf 60.6 & \bf 65.0 & \bf 64.2 \\
    \bottomrule
\end{tabular}}
\setlength{\tabcolsep}{3pt}
\resizebox{\linewidth}{!}{
\begin{tabular}{l|c|ccccccccc}
    \toprule
    \multicolumn{11}{c}{\bf\emph{MGSM}} \\
    \midrule
    \bf Model & Avg. & eng\_Latn & spa\_Latn & fra\_Latn & deu\_Latn & rus\_Cyrl & zho\_Hans & jpn\_Jpan & tha\_Thai & ben\_Beng \\ 
    \midrule
    Qwen3-4B-Multilingual & 73.7 & 82.0 & 80.8 & 72.4 & 74.0 & 84.4 & 71.2 & 60.0 & 76.0 & 62.4  \\
    \quad + Multilingual STEM & \bf 81.6 & \bf 88.4 & \bf 88.0 & \bf 83.2 & \bf 84.0 & \bf 87.6 & \bf 79.2 & \bf 76.0 & \bf 80.4 & \bf 67.6 \\
    \bottomrule
\end{tabular}}
\caption{The effects of including multilingual STEM data into the pre-training mixture. Performance of selected languages on the GlobalMMLU, MMMLU, and MGSM benchmarks is presented.}
\label{tab:multilingual_stem_ablation}
\end{table}

\boldtitle{Multilingual STEM Data.}
STEM corpora are characterized by high information density and have been shown to enhance both the general and task-specific capabilities of LLMs. Nevertheless, such resources remain scarce across most non-English languages. To address this disparity, we employ a translation-based data synthesis pipeline to project English STEM knowledge into various target languages. Specifically, for general-domain STEM content, we leverage the \href{https://huggingface.co/datasets/nvidia/}{Nemotron-Pretraining-SFT-v1} dataset as our primary source, applying our synthesis framework to generate high-fidelity multilingual data across 28 languages. Furthermore, we curate a specialized multilingual mathematical dataset by translating the OpenMathInstruct-2 corpus \citep{toshniwal2024openmath2}.

To assess the efficacy of the curated multilingual STEM dataset, we conduct an ablation study utilizing a 100B-token annealing phase with a 50\% mixture of curated data. Performance is measured using the GlobalMMLU \citep{singh-etal-2025-global}, \href{https://huggingface.co/datasets/openai/MMMLU}{MMMLU}, and MGSM \citep{shi2023language} benchmarks. As demonstrated in Table \ref{tab:multilingual_stem_ablation}, the integration of multilingual STEM synthetic data significantly enhances performance across relevant benchmarks, yielding an average absolute gain of up to 8\%. Similarly, these benefits become increasingly pronounced as the resource level of the target language decreases.

\boldtitle{Cultural and Regional Data.}
An effective multilingual LLM must transcend basic cross-lingual processing; it must interpret the nuanced socio-cultural contexts and conceptual frameworks inherent to each language. However, such culturally salient knowledge is often fragmented across vast web corpora, and its volume is disproportionately small compared to language-agnostic or general-domain data. To mitigate this data scarcity, we employ two distinct strategies to curate high-quality datasets rich in regional and cultural features.

We first identify web-crawled documents (\eg Fineweb-2) characterized by high cultural density. Specifically, we utilize an LLM-based evaluator (see Appendix Prompt~\ref{prompt:regional_annotate}) to assess multilingual documents across three dimensions: quality, geographic origin, and domain. We retain only those documents that achieve a quality score exceeding 7 and whose predicted country tag aligns with the primary language. To maximize token utility, these curated regional documents are further augmented through rephrasing and QA generation strategies~\citep{kimiteam2026kimik2openagentic} for specific low-resource languages. The final dataset is denoted as \textbf{Fineweb2-Culture}.

Complementing our web-mining efforts, we develop a pipeline for generating diverse synthetic data focused on culture-specific topics. This process begins with the gathering of coarse-grained cultural themes. We then leverage Gemini-2.5-Pro \citep{comanici2025gemini25pushingfrontier} to perform hierarchical expansion, decomposing these themes into fine-grained sub-topics. Finally, these sub-topics serve as the basis for generating a vast array of diverse multiple-choice questions (MCQs), utilizing a prompt adapted from~\citet{bercovich2025llamanemotronefficientreasoningmodels} and \citet{NemotronPostTrainingDatasetV1} (see Appendix Prompt~\ref{prompt:cultural_mcq_prompt}). We denote the final dataset as \textbf{Synthetic-Regional-MCQs}.

\begin{table}
\setlength{\tabcolsep}{15pt}
\footnotesize
\centering
\resizebox{0.85\linewidth}{!}{
\begin{tabular}{l|cccc}
\toprule
    \bf Model & INCLUDE & Indo-MMLU & Turkish-MMLU \\
    \midrule
    Qwen3-4B-Multilingual & 62.3 & 60.1 & 60.1 \\
    \quad + Cultural \& Regional Data & \bf 63.5 & \bf 61.2 & \bf 64.0 \\
    \bottomrule
\end{tabular}}
\caption{The effects of including cultural and regional data into the pre-training mixture. Performance of selected languages on the INCLUDE, Indo-MMLU, and Turkish-MMLU benchmarks is presented.}
\label{tab:cultural_ablation}
\end{table}
To evaluate the efficacy of the curated data, we conduct an ablation study utilizing an annealing phase with a 10B-token budget. Performance was assessed across three cultural benchmarks: INCLUDE~\citep{romanou2025include}, Indo-MMLU~\citep{koto-etal-2023-large}, and Turkish-MMLU~\citep{yuksel-etal-2024-turkishmmlu}. As illustrated in Table~\ref{tab:cultural_ablation}, the integration of our synthetic cultural data consistently enhances model performance on tasks requiring localized cultural knowledge.

\begin{figure}
    \centering
    \subfigure[Data mixture of Phase 1.]{
        \label{fig:phase1_data_mixture}
        \includegraphics[width=\textwidth]{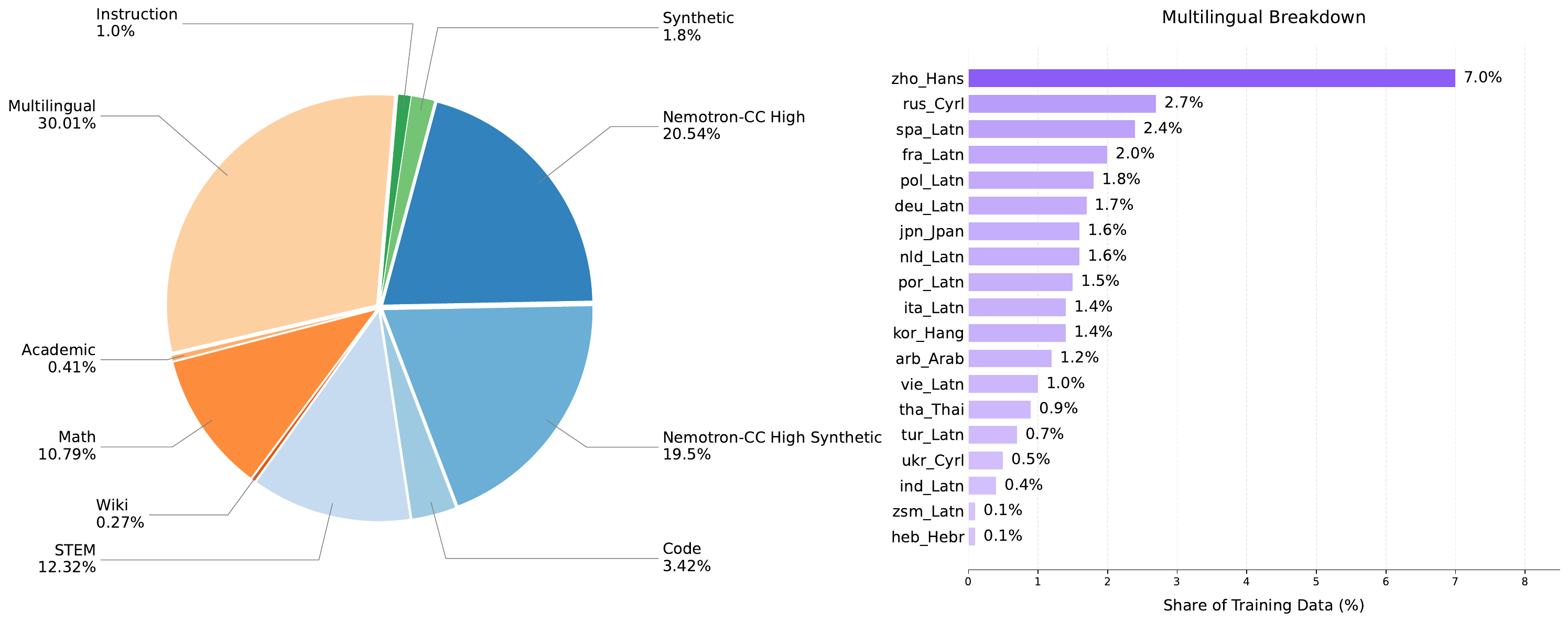}
    }
    \subfigure[Data mixture of Phase 2.]{
        \label{fig:phase2_data_mixture}
        \includegraphics[width=\textwidth]{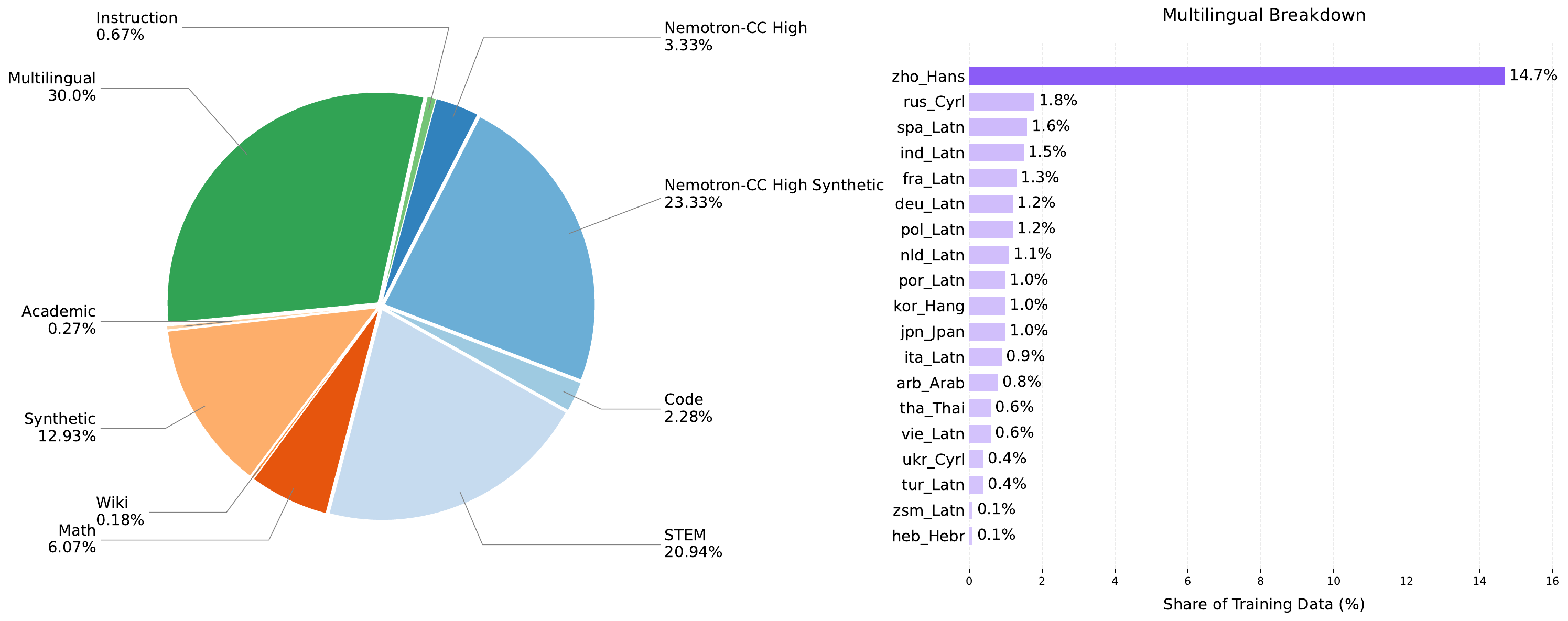}
    }
    \subfigure[Data mixture of Phase 3.]{
        \label{fig:phase3_data_mixture}
        \includegraphics[width=\textwidth]{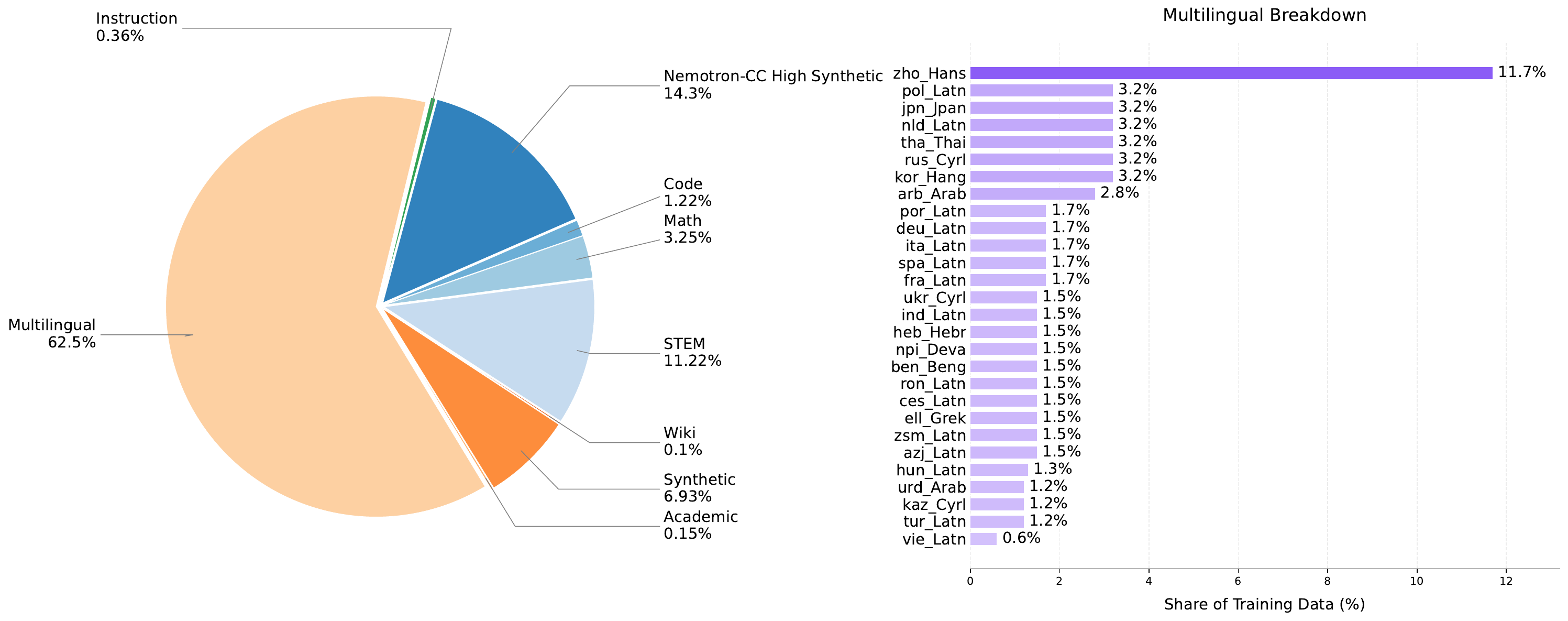}
    }
\end{figure}
\begin{figure}
    \centering
    \subfigure[Data mixture of Phase 4.]{
        \label{fig:phase4_data_mixture}
        \includegraphics[width=\textwidth]{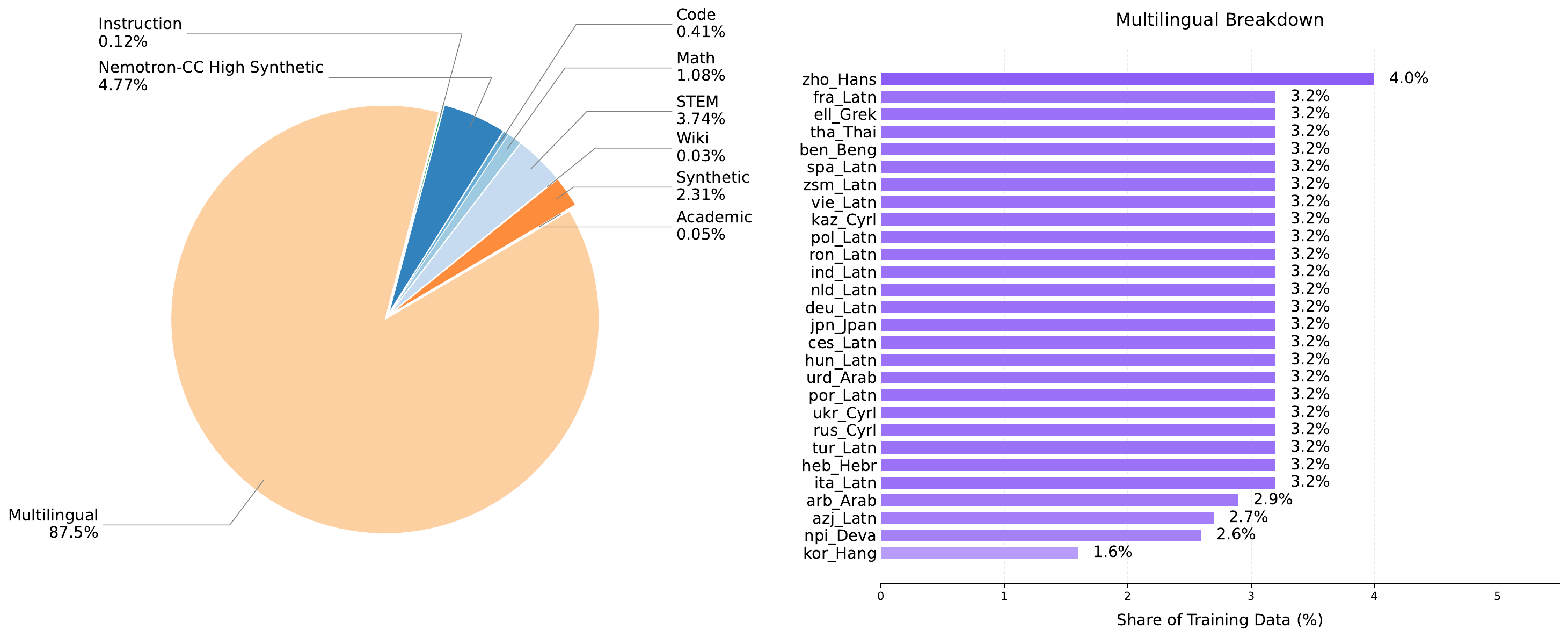}
    }
    \caption{Data mixtures for each pre-training phase.}
    \label{fig:data_mixture_pt}
\end{figure}
\begin{table}
\setlength{\tabcolsep}{5pt}
\footnotesize
\centering
\resizebox{\linewidth}{!}{
\begin{tabular}{l|cccc}
\toprule
    Dataset & Phase1-Mix & Phase2-Mix & Phase3-Mix & Phase4-Mix \\
    \midrule
    \href{https://huggingface.co/datasets/nvidia/Nemotron-CC-v2/tree/main/High-Quality}{Nemotron-CC-v2/High-Quality} & 20.54\% & 6.5\% & - & - \\
    \href{https://huggingface.co/datasets/nvidia/Nemotron-CC-v2/tree/main/Diverse-QA}{Nemotron-CC-v2/Diverse-QA} & 18.5\% & 18.33\% & 14.13\% & 6.5\% \\
    \href{https://huggingface.co/datasets/nvidia/Nemotron-CC-v2/tree/main/High-Quality-Synthetic}{Nemotron-CC-v2/High-Quality-Synthetic} & 0.94\% & 5.17\% & - & - \\
    \href{https://huggingface.co/datasets/nvidia/Nemotron-Pretraining-SFT-v1/tree/main/Nemotron-SFT-General}{Nemotron-Pretraining-SFT-v1/Nemotron-SFT-General} & 6.0\% & 2.02\% & 1.18\% & 0.30\% \\
    \href{https://huggingface.co/datasets/nvidia/Nemotron-Pretraining-Specialized-v1/tree/main/Nemotron-Pretraining-STEM-SFT}{Nemotron-Pretraining-Specialized-v1/Nemotron-Pretraining-STEM-SFT} & 2.61\% & 7.70\% & 4.50\% & 1.15\% \\
    \href{https://huggingface.co/datasets/nvidia/Nemotron-Pretraining-Specialized-v1/tree/main/Nemotron-Pretraining-RQA}{Nemotron-Pretraining-Specialized-v1/Nemotron-Pretraining-RQA} & 2.44\% & 10.37\% & 6.06\% & 1.56\% \\
    \href{https://huggingface.co/datasets/nvidia/Nemotron-Pretraining-Specialized-v1/tree/main/Nemotron-Pretraining-Wiki-Rewrite}{Nemotron-Pretraining-Specialized-v1/Nemotron-Pretraining-Wiki-Rewrite} & 0.27\% & 0.17\% & 0.1\% & 0.026\% \\
    \href{https://huggingface.co/datasets/nvidia/Nemotron-Pretraining-Specialized-v1/tree/main/Nemotron-Pretraining-InfiniByte-Reasoning}{Nemotron-Pretraining-Specialized-v1/Nemotron-Pretraining-InfiniByte-Reasoning} & 1.0\% & 0.64\% & 0.37\% & 0.096\% \\
    \href{https://huggingface.co/datasets/nvidia/Nemotron-Pretraining-Specialized-v1/tree/main/Nemotron-Pretraining-Math-Textbooks}{Nemotron-Pretraining-Specialized-v1/Nemotron-Pretraining-Math-Textbooks} & 0.87\% & 0.70\% & 0.41\% & 0.10\% \\
    \href{https://huggingface.co/datasets/nvidia/Nemotron-CC-v2.1/tree/main/High-Quality-DQA}{Nemotron-CC-v2.1/High-Quality-DQA} & 0.27\% & 0.17\% & 0.1\% & 0.026\% \\
    \href{https://huggingface.co/datasets/allenai/dolmino-mix-1124/tree/main/data/pes2o}{peS2o} & 0.41\% & 0.26\% & 0.15\% & 0.039\% \\
    \href{https://huggingface.co/datasets/nvidia/Nemotron-Pretraining-SFT-v1/tree/main/Nemotron-SFT-MATH}{Nemotron-Pretraining-SFT-v1/Nemotron-SFT-MATH} & 6.93\% & 4.45\% & 2.60\% & 0.67\% \\
    \href{https://huggingface.co/datasets/nvidia/Nemotron-CC-Math-v1/tree/main/4plus}{Nemotron-CC-Math-v1/4plus} & 1.30\% & 0.69\% & 0.40\% & 0.10\% \\
    \href{https://huggingface.co/datasets/nvidia/OpenMathInstruct-2}{OpenMathInstruct-2} & 0.50\% & 2.25\% & 1.31\% & 0.37\% \\
    \href{https://huggingface.co/datasets/HuggingFaceTB/finemath/tree/main/finemath-4plus}{Finemath/finemath-4plus} & 0.33\% & - & - & - \\
    \href{https://huggingface.co/datasets/HuggingFaceTB/finemath/tree/main/infiwebmath-4plus}{Finemath/infiwebmath-4plus} & 0.30\% & - & - & - \\
    \href{https://huggingface.co/datasets/LLM360/MegaMath/tree/main/megamath-web-pro}{Megamath/megamath-web-pro} & 0.45\% & - & - & - \\
    \href{https://huggingface.co/datasets/LLM360/MegaMath/tree/main/megamath-qa/qwen-2.5}{Megamath/megamath-qa/qwen2.5} & 0.11\% & - & - & - \\
    \href{https://huggingface.co/datasets/nvidia/Nemotron-Pretraining-SFT-v1/tree/main/Nemotron-SFT-Code}{Nemotron-Pretraining-SFT-v1/Nemotron-SFT-Code} & 1.90\% & 1.22\% & 0.71\% & 0.18\% \\
    \href{https://huggingface.co/datasets/LLM360/MegaMath/tree/main/megamath-text-code-block}{Megamath/megamath-text-code-block} & 1.43\% & 0.92\% & 0.54\% & 0.14\% \\
    \href{https://huggingface.co/datasets/nvidia/Nemotron-Pretraining-Specialized-v1/tree/main/Nemotron-Pretraining-Scientific-Coding}{Nemotron-Pretraining-Specialized-v1/Nemotron-Pretraining-Scientific-Coding} & 0.04\% & 0.026\% & 0.015\% & 0.004\% \\
    \href{https://huggingface.co/datasets/allenai/dolmino-mix-1124/tree/main/data/stackexchange}{Stackexchange} & 0.04\% & 0.028\% & 0.016\% & 0.004\% \\
    \href{https://huggingface.co/datasets/open-thoughts/OpenThoughts3-1.2M}{OpenThoughts3-1.2M} & 1.80\% & 2.69\% & 1.57\% & 0.40\% \\
    \href{https://huggingface.co/datasets/allenai/dolmino-mix-1124/tree/main/data/flan}{FLAN} & 1.0\% & 0.64\% & 0.37\% & 0.096\% \\
    \href{https://huggingface.co/datasets/opencsg/Fineweb-Edu-Chinese-V2.1}{Fineweb-Edu-Chinese-V2.1} & 6.69\% & 14.74\% & 10.0\% & 0.89\% \\
    \href{https://huggingface.co/datasets/HuggingFaceFW/fineweb-2}{Fineweb-2} \& \href{https://huggingface.co/datasets/epfml/FineWeb2-HQ}{Fineweb-2-HQ} & 13.99\% & 9.15\% & 30.0\% & - \\
    \href{https://huggingface.co/datasets/nvidia/Nemotron-CC-v2/tree/main/Translated-Diverse-QA}{Nemotron-CC-v2/Translated-Diverse-QA} & 9.33\% & 6.11\% & 22.5\% & 14.3\% \\
    \href{https://huggingface.co/datasets/allenai/dolma3_dolmino_mix-100B-1125}{Dolma3-dolmino-mix-100B-1125} & - & 5.06\% & 2.96\% & 0.79\% \\
    Fineweb2-Culture & - & - & - & 25.0\% \\
    Translated-Nemotron-SFT-General & - & - & - & 25.0\% \\
    Translated-OpenMathInstruct-2 & - & - & - & 12.5\% \\
    Synthetic-Regional-MCQs & - & - & - & 9.82\% \\
    \bottomrule
\end{tabular}}
\caption{Pre-training datasets and mixture ratios for the four pre-training stages.}
\label{tab:data_mixtures}
\end{table}

\subsection{Pre-training Recipe}
\subsubsection{Four-Stage Data Mixture}
We adopt a four-stage data mixture strategy to pre-train \marcomoe models, processing a total of 5.1T tokens. The data mixtures used in each stage are shown in Figure~\ref{fig:data_mixture_pt} and Table~\ref{tab:data_mixtures}.
\begin{compactenum}
    \item \textbf{Stage 1 (0 -- 2.4T tokens)}: The foundational training stage comprises a high-quality blend of English, reasoning, and instruction corpora, alongside multilingual web and QA data. The multilingual component covers 19 languages (Chinese, Arabic, German, Spanish, French, Korean, Japanese, Portuguese, Turkish, Indonesian, Italian, Dutch, Polish, Russian, Vietnamese, Thai, Hebrew, Ukrainian, and Malay).
    \item \textbf{Stage 2 (2.4T -- 4.1T tokens)}: To optimize the curriculum, we adjust the data mixture by upsampling reasoning corpora and downsampling English web data. We simultaneously upsample Chinese data to improve targeted linguistic performance. 
    \item \textbf{Stage 3 (4.1T -- 4.6T tokens)}: The sampling weights for English, reasoning, and instruction domains are reduced to allocate token bandwidth for nine newly introduced languages: Bengali, Czech, Urdu, Kazakh, Greek, Romanian, Hungarian, Nepali, and Azerbaijani. We concurrently upsample medium-resource languages to ensure balanced cross-lingual efficacy.
    \item \textbf{Stage 4 (4.6T -- 5.1T tokens)}: We further decay the sampling weights of the English, reasoning, and instruction corpora, integrating a diverse array of curated multilingual synthetic data to explicitly target and boost specific reasoning and linguistic capabilities across all languages.
\end{compactenum}

\begin{figure*}[t]
    \centering
    \includegraphics[width=\textwidth]{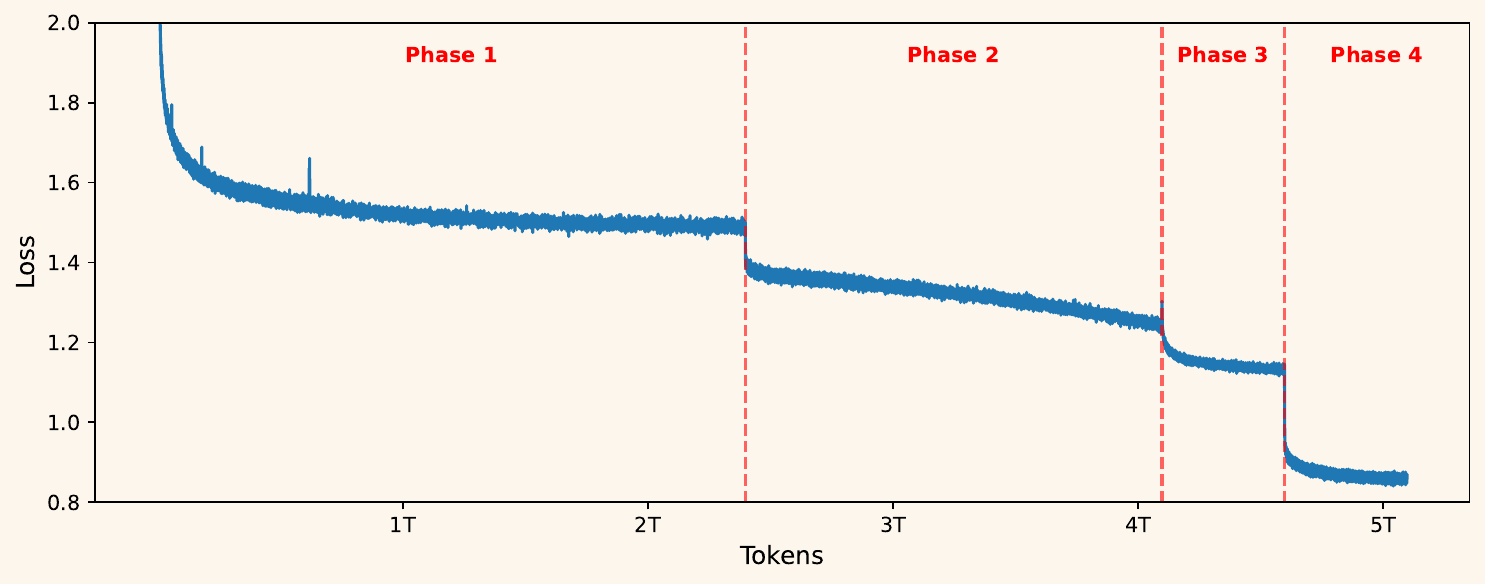}
    \caption{The training loss graph for \marcomedium. We indicate the boundaries between training phases where data mixtures are switched.}
    \label{fig:pretraining_loss}
\end{figure*}

\begin{table}
\setlength{\tabcolsep}{10pt}
\footnotesize
\centering
\resizebox{0.8\linewidth}{!}{
\begin{tabular}{l|cccc}
\toprule
    \bf \marcosmall & \bf Stage-1 & \bf Stage-2 & \bf Stage-3 & \bf Stage-4 \\ 
    \midrule
    Learning Rate Schedule & Constant & Linear Decay & Linear Decay & Linear Decay \\
    LR Warmup steps & 890 & 0 & 0 & 0 \\
    Peak LR & $4.9505\times10^{-4}$ & $4.9505\times10^{-4}$ & $1\times10^{-5}$ & $6\times10^{-6}$ \\
    Final LR & $4.9505\times10^{-4}$ & $1\times10^{5}$ & $6\times10^{-6}$ & 0 \\
    Batch size (\# instances) & 1,152 & 1,584 & 2,304 & 2,304 \\
    Sequence length & 8,192 & 8,192 & 8,192 & 8,192 \\
    Total training tokens & 2.4T & 1.7T & 500B & 500B \\
    \bottomrule
    \toprule
    \bf \marcomedium & \bf Stage-1 & \bf Stage-2 & \bf Stage-3 & \bf Stage-4 \\ 
    \midrule
    Learning Rate Schedule & Constant & Linear Decay & Linear Decay & Linear Decay \\
    LR Warmup steps & 890 & 0 & 0 & 0 \\
    Peak LR & $4.6854\times10^{-4}$ & $4.6854\times10^{-4}$ & $1\times10^{-5}$ & $6\times10^{-6}$ \\
    Final LR & $4.6854\times10^{-4}$ & $1\times10^{5}$ & $6\times10^{-6}$ & 0 \\
    Batch size (\# instances) & 1,152 & 1,584 & 2,304 & 2,304 \\
    Sequence length & 8,192 & 8,192 & 8,192 & 8,192 \\
    Total training tokens & 2.4T & 1.7T & 500B & 500B \\
    \bottomrule
\end{tabular}}
\caption{Pre-training hyperparameters for each stage of \marcosmall and \marcomedium.}
\label{tab:hyperparameters}
\end{table}
\subsubsection{Hyperparameters}
We pre-train the \marcomoe models on a total of 5.1T tokens using the Megatron-LM framework~\citep{shoeybi2020megatronlmtrainingmultibillionparameter}, employing the Warmup-Stable-Decay learning rate (LR) schedule~\citep{hu2024minicpm}. Specifically, the LR is linearly warmed up over 8.4B tokens to its peak value, which is subsequently maintained throughout Stage 1. Peak learning rates are determined based on our training FLOPs, guided by the scaling laws proposed by~\citet{tian2025greaterleveragescalinglaws}, yielding $4.9505\times 10^{-4}$ for \marcosmall and $4.6854\times 10^{-4}$ for \marcomedium. During Stage 2, the LR undergoes a linear decay to $1.0\times10^{-5}$ over 1.7T tokens. In Stages 3 and 4, the LR is further linearly decayed to zero over an additional 1T tokens. Initial pre-training utilizes a sequence length of 8,192 and a batch size of 1,152, resulting in approximately 9.4M tokens per batch. This batch size is later expanded to 1,584 (roughly 13M tokens per batch) to accommodate a hardware scaling from 128 to 176 GPUs in Stage 2. In Stages 3 and 4, we further increase the batch size to 2,304 (roughly 19M tokens per batch). Following~\citet{muennighoff2025olmoe}, we stabilize MoE training by incorporating a Load Balancing Loss~\citep{shazeer2017outrageouslylargeneuralnetworks} with a coefficient of 0.01 and a Router Z-loss~\citep{zoph2022stmoedesigningstabletransferable} with a coefficient of 0.001. Optimization is performed using AdamW~\citep{loshchilov2018decoupled} with a weight decay of 0.1, $\beta_1 = 0.9$, and $\beta_2 = 0.95$. Table~\ref{tab:hyperparameters} presents the detailed hyperparameters across the four training stages. Figure~\ref{fig:pretraining_loss} illustrates the detailed pre-training loss trajectory of \marcomedium.
\subsection{Base Model Evaluation}

\subsubsection{Experimental Settings}
\boldtitle{Evaluation Benchmarks.}
Unless otherwise specified, all model evaluations are conducted using the Light-Eval framework~\citep{lighteval} to ensure a standardized and fair comparison. The evaluation suite is divided into two primary dimensions:
\begin{compactenum}
    \item English Proficiency: We assess general language understanding and reasoning via a broad array of benchmarks, including MMLU~\citep{hendrycks2021measuring}, MMLU-Redux~\citep{gema-etal-2025-done}, MMLU-Pro~\citep{wang2024mmlupro}, BBH~\citep{suzgun-etal-2023-challenging}, AGIEval~\citep{zhong-etal-2024-agieval}, ARC-Easy/Challenge~\citep{clark2018thinksolvedquestionanswering}, HellaSwag~\citep{zellers-etal-2019-hellaswag}, WinoGrande~\citep{sakaguchi2019winogrande}, and several commonsense reasoning tasks such as BoolQ~\citep{clark-etal-2019-boolq}, CommonsenseQA~\citep{talmor-etal-2019-commonsenseqa}, OpenBookQA~\citep{banerjee-etal-2019-careful}, PIQA~\citep{bisk2019piqareasoningphysicalcommonsense}, SIQA~\citep{sap-etal-2019-social}, and GSM8K~\citep{cobbe2021trainingverifierssolvemath}.
    \item Multilingual Capability: To measure performance across diverse languages, we evaluate the model on both general multilingual benchmarks and datasets featuring cultural and regional knowledge assessment. General datasets include GlobalMMLU~\citep{singh-etal-2025-global}, MMLU-ProX-Lite~\citep{xuan-etal-2025-mmlu}, BELEBELE~\citep{bandarkar-etal-2024-belebele}, multilingual HellaSwag \& ARC~\citep{lai-etal-2023-okapi}, \href{https://huggingface.co/datasets/openai/MMMLU}{MMMLU}, MGSM~\citep{shi2023language}, and translation tasks with FLORES-200~\citep{nllbteam2022languageleftbehindscaling} and WMT24$++$~\citep{freitag-etal-2024-llms}. For cultural and regional benchmarks with native-sourced content, we evaluate the models on INCLUDE~\citep{romanou2025include}, Global-PIQA~\citep{chang2025globalpiqaevaluatingphysical}, CMMLU~\citep{li-etal-2024-cmmlu},  C-Eval~\citep{huang2023ceval}, ArabicMMLU~\citep{koto-etal-2024-arabicmmlu}, TurkishMMLU~\citep{yuksel-etal-2024-turkishmmlu}, GreekMMLU~\citep{zhang2026greekmmlunativesourcedmultitaskbenchmark}, KazakhMMLU~\citep{togmanov-etal-2025-kazmmlu}, IndoMMLU~\citep{koto-etal-2023-large}, IndoCareer~\citep{koto-2025-cracking}, and IndoCulture~\citep{koto-etal-2024-indoculture}.
\end{compactenum}

\boldtitle{Baselines.}
We compare \marcosmall and \marcomedium against a diverse set of open-source baselines, including both open-weight and open-weight \& open-data models, selected for their explicit multilingual focus as well as their comparable model scale and training compute. The comparison set includes Qwen3 1.7B and 4B~\citep{yang2025qwen3technicalreport}, Granite4-Tiny~\citep{granite2025}, Llama3.2-3B~\citep{grattafiori2024llama3herdmodels}, SmolLM3-3B~\citep{bakouch2025smollm3}, Gemma3-4B~\citep{gemmateam2025gemma3technicalreport}, Tiny-Aya-3.35B~\citep{salamanca2026tinyayabridgingscale}, and Trinity Nano and Mini~\citep{singh2026arceetrinitylargetechnical}.

\subsubsection{Main Results}
We evaluate the proposed \marcosmall and \marcomedium across three axes: English capability, multilingual generalization, and multilingual cultural/regional knowledge. The results in Table~\ref{tab:main_results} show a clear pattern: the \marcomoe family is strongest where broad cross-lingual transfer matters most, and does so with an unusually favorable compute–performance tradeoff. \marcosmall establishes a highly competitive efficiency frontier for compact multilingual models, while \marcomedium emerges as the strongest overall model in the comparison.

\boldtitle{English results.}
On English benchmarks, \marcomedium achieves the best average performance overall (63.7), surpassing all baselines, including Qwen3-4B Base (63.3). This is a particularly strong result because it is obtained with substantially lower training compute than Qwen3-4B ($1.56×10^{23}$ \versus $8.64×10^{23}$ FLOPs), and way less activated and total parameters than Trinity-Mini (0.86B \versus 3.85B). The gains are not driven by a single benchmark: \marcomedium is consistently strong across knowledge, commonsense, and reasoning tasks, achieving the top result on ARC-Challenge (56.3) and CommonsenseQA (61.5), while remaining near the top on knowledge-intensive and math-oriented benchmarks including MMLU, MMLU-Redux, MMLU-Pro, and GSM8K, where in particular, Qwen3-4B remains stronger due to its excessive exposure to knowledge and reasoning-heavy pre-training data. These results establish \marcomedium as a top-tier general-purpose English model, not merely a multilingual model with acceptable English transfer.
\begin{table}[H]
\setlength{\tabcolsep}{2.5pt}
\footnotesize
\centering
\resizebox{\linewidth}{!}{
\begin{tabular}{lc|cccc|ccccccc}
\toprule
    \multirow{3}{*}[0ex]{\bf Benchmark $_{(\text{Metric})}$} & \multirow{3}{*}[0ex]{\bf\# Shots} & \bf Qwen3 & \bf Trinity & \bf Granite4 & \bf Marco & \bf Llama3.2 & \bf SmolLM3 & \bf Gemma3 & \bf Tiny-Aya & \bf Qwen3 & \bf Trinity & \bf Marco \\
    & & \bf 1.7B & \bf Nano & \bf Tiny & \bf Nano & \bf 3B & \bf 3B & \bf 4B & \bf 3.35B & \bf 4B & \bf Mini & \bf Mini \\
    & & \scriptsize \textsc{Base} & \scriptsize \textsc{Base} & \scriptsize \textsc{Base} & \scriptsize \textsc{Base} & \scriptsize \textsc{Base} & \scriptsize \textsc{Base} & \scriptsize \textsc{Base} & \scriptsize \textsc{Base} & \scriptsize \textsc{Base} & \scriptsize \textsc{Base} & \scriptsize \textsc{Base} \\
    \midrule
    \# Activated Params & - & 1.7B & 1.09B & 1.47B & 0.6B & 3B & 3B & 4B & 3.35B & 4B & 3.85B & 0.86B \\
    \# Total Params & - & 1.7B & 6B & 7B & 8B & 3B & 3B & 4B & 3.35B & 4B & 26.5B & 17.3B \\
    \# Train FLOPs ($\times10^{23}$) & - & 3.67 & 0.654 & 2.03 & 1.40 & 1.62 & 1.98 & 0.96 & 1.21 & 8.64 & 2.31 & 1.56 \\
    \midrule
    \multicolumn{11}{l}{\bf English} \\
    MMLU $_{\text{(Acc)}}$ & 5-shot & \underline{65.1} & 64.7 & \bf 69.1 & 64.7 & 57.6 & 62.6 & 61.1 & 58.6 & \bf 75.2 & 71.4 & \underline{72.8} \\
    MMLU-Redux $_{\text{(Acc)}}$ & 0-shot & 61.2 & 60.1 & \bf 65.8 & \underline{62.9} & 56.9 & 58.4 & 57.7 & 51.7 & \bf 71.3 & 68.2 & \underline{68.8} \\
    MMLU-Pro $_{\text{(Acc)}}$ & 5-shot & \underline{33.2} & 32.0 & 32.1 & \bf 35.9 & 26.0 & 35.1 & 28.8 & 26.9 & \bf 45.9 & 41.3 & \underline{45.3} \\
    AGIEval $_{\text{(Acc)}}$ & 0-shot & 35.9 & 31.4 & \underline{36.1} & \bf 38.4 & 31.2 & 34.5 & 32.6 & 29.0 & \bf 44.0 & 39.7 & \underline{41.9} \\
    BBH $_{\text{(EM)}}$ & 3-shot & \underline{54.5} & 49.3 & \bf 59.9 & 53.5 & 47.1 & 60.0 & 52.2 & 46.8 & \bf 72.3 & 57.6 & \underline{65.1} \\
    ARC-Easy $_{\text{(Acc)}}$ & 0-shot & 69.3 & \underline{77.9} & \bf 78.5 & 75.3 & 71.8 & 78.5 & \bf 82.6 & 76.5 & 75.0 & 80.6 & \underline{82.4} \\
    ARC-Challenge $_{\text{(Acc)}}$ & 0-shot & 42.8 & \bf 53.5 & \underline{52.3} & 49.4 & 46.0 & 52.6 & 54.1 & 47.4 & 49.9 & \bf 57.8 & \underline{56.3} \\
    HellaSwag $_{\text{(Acc)}}$ & 0-shot & 66.6 & \underline{77.4} & \bf 77.9 & 69.2 & 75.6 & 76.1 & 76.7 & 71.0 & 74.4 & \bf 82.8 & \underline{77.4} \\
    WinoGrande $_{\text{(Acc)}}$ & 0-shot & \underline{57.1} & 57.1 & \bf 58.6 & 53.4 & 58.6 & 58.9 & \bf 61.4 & 56.6 & 59.6 & \underline{60.8} & 57.7 \\
    BoolQ $_{\text{(Acc)}}$ & 0-shot & \bf 74.6 & \underline{71.5} & 63.5 & 71.2 & 75.2 & \bf 79.3 & \underline{76.6} & 74.6 & 74.2 & 72.5 & 74.2 \\
    CommonsenseQA $_{\text{(Acc)}}$ & 0-shot & 49.5 & 54.1 & \bf 55.9 & \underline{55.7} & 60.4 & 55.4 & \underline{61.1} & 60.4 & 52.9 & 57.7 & \bf 61.5 \\
    OpenBookQA $_{\text{(Acc)}}$ & 0-shot & 36.4 & \underline{42.0} & \bf 43.6 & 39.4 & 42.2 & 40.4 & \underline{42.6} & 40.4 & 42.6 & \bf 44.8 & \underline{44.6} \\
    PIQA $_{\text{(Acc)}}$ & 0-shot & 75.5 & 69.6 & \bf 80.6 & \underline{76.5} & 78.2 & 79.1 & \underline{80.3} & 76.9 & 77.4 & 71.7 & \bf 81.1 \\
    SIQA $_{\text{(Acc)}}$ & 0-shot & 47.8 & \underline{52.7} & \bf 53.0 & 46.0 & 51.0 & 49.8 & 50.4 & 49.9 & \bf 53.0 & \underline{52.5} & 49.4 \\
    GSM8K $_{\text{(EM)}}$ & 5-shot & 69.1 & 57.8 & \bf 70.7 & \underline{69.7} & 27.3 & 67.4 & 39.3 & 58.0 & \bf 81.7 & 57.5 & \underline{76.4} \\
    \bf Average & - & 55.9 & 56.7 & \bf 59.8 & \underline{57.5} & 53.7 & 59.2 & 57.2 & 55.5 & \underline{63.3} & 61.1 & \bf 63.7 \\
    
    \midrule
    \multicolumn{11}{l}{\bf Multilingual -- General} \\
    GlobalMMLU $_{\text{(Acc)}}$ & 5-shot & 49.6 & 43.6 & \bf 54.8 & \underline{52.2} & 43.2 & 46.7 & 50.8 & 50.0 & \underline{61.6} & 52.6 & \bf 64.2 \\
    MMMLU $_{\text{(Acc)}}$ & 0-shot & 48.6 & 41.2 & \underline{52.3} & \bf 52.6 & 44.0 & 47.3 & 47.4 & 44.5 & \underline{59.3} & 50.9 & \bf 62.0 \\
    MMLU-ProX-Lite $_{\text{(Acc)}}$ & 5-shot & 27.2 & 20.3 & \bf 30.1 & \underline{28.9} & 22.4 & 28.3 & 24.3 & 24.3 & \underline{38.5} & 32.2 & \bf 39.2 \\ 
    BELEBELE $_{\text{(Acc)}}$ & 0-shot & \underline{67.5} & 54.5 & 61.2 & \bf 73.8 & 60.1 & 54.3 & 65.7 & 65.4 & \bf 81.5 & 67.6 & \underline{79.8} \\
    mHellaSwag $_{\text{(Acc\_norm)}}$ & 0-shot & 43.9 & 42.5 & \bf 53.2 & \underline{48.8} & 49.0 & 49.6 & \underline{55.2} & 53.5 & 53.2 & 51.5 & \bf 58.6 \\
    mARC-Challenge $_{\text{(Acc\_norm)}}$ & 0-shot & 34.7 & 30.9 & \bf 39.9 & \underline{36.9} & 34.2 & 36.1 & 41.5 & 37.2 & \underline{42.5} & 37.5 & \bf 45.4 \\
    FLORES-200 (En-Xx) $_{\text{(BLEU)}}$ & 5-shot & 18.6 & 15.1 & \bf 25.4 & \underline{24.7} & 23.5 & 19.7 & \underline{32.1} & 30.2 & 25.4 & 13.7 & \bf 32.3 \\
    FLORES-200 (Xx-En) $_{\text{(BLEU)}}$ & 5-shot & 31.5 & 31.1 & \bf 36.7 & \underline{33.6} & 34.6 & 30.3 & \underline{39.7} & 37.3 & 36.8 & 24.1 & \bf 40.1 \\
    WMT24$++$ (En-Xx) $_{\text{(BLEU)}}$ & 5-shot & 18.3 & 15.0 & \bf 21.9 & \underline{20.7} & 16.4 & 17.8 & \underline{27.7} & 26.1 & 23.9 & 7.5 & \bf 28.1 \\
    WMT24$++$ (Xx-En) $_{\text{(BLEU)}}$ & 5-shot & \underline{28.3} & 28.0 & \bf 30.7 & 28.1 & 28.9 & 27.4 & \underline{34.0} & 32.7 & 32.9 & 10.6 & \bf 34.4 \\
    MGSM $_{\text{(EM)}}$ & 8-shot & \underline{58.8} & 40.6 & 56.7 & \bf 65.3 & 22.4 & 50.8 & 36.6 & 38.4 & \bf 76.0 & 57.2 & \underline{75.6} \\
    \bf Average & - & 38.8 & 33.0 & \underline{42.1} & \bf 42.3 & 34.4 & 37.1 & 41.4 & 39.9 & \underline{48.3} & 36.9 & \bf 50.9 \\
    
    \midrule
    \multicolumn{11}{l}{\bf Multilingual -- Cultural \& Regional} \\
    INCLUDE $_{\text{(Acc)}}$ & 5-shot & 51.2 & 43.9 & \underline{52.1} & \bf 53.2 & 45.5 & 46.2 & 52.6 & 53.9 & \underline{61.4} & 51.9 & \bf 61.7 \\
    Global-PIQA $_{\text{(Acc\_norm)}}$ & 0-shot & 60.3 & 52.3 & \underline{64.0} & \bf 64.3 & 62.2 & 60.9 & \underline{69.4} & 67.9 & 65.4 & 57.2 & \bf 72.3 \\
    CMMLU $_{\text{(Acc)}}$ & 5-shot & \bf 66.1 & 49.6 & 53.5 & \underline{55.5} & 44.1 & 50.1 & 50.2 & 58.8 & \bf 76.2 & 58.6 & \underline{68.0} \\
    C-Eval $_{\text{(Acc)}}$ & 5-shot & \bf 65.1 & 47.6 & 50.9 & \underline{56.0} & 43.1 & 47.9 & 48.5 & 57.6 & \bf 76.6 & 57.1 & \underline{66.0} \\
    ArabicMMLU $_{\text{(Acc)}}$ & 3-shot & \underline{57.6} & 44.0 & \bf 60.5 & 55.8 & 48.9 & 60.6 & 61.6 & 63.2 & \underline{67.0} & 57.1 & \bf 67.1 \\
    TurkishMMLU $_{\text{(Acc)}}$ & 5-shot & \underline{47.9} & 29.6 & 41.8 & \bf 48.9 & 36.7 & 28.4 & 43.7 & 45.2 & \underline{60.6} & 43.0 & \bf 62.7 \\
    GreekMMLU $_{\text{(Acc)}}$ & 5-shot & 58.1 & 52.2 & \underline{62.3} & \bf 64.1 & 56.4 & 64.0 & 63.4 & 66.3 & \underline{69.4} & 59.7 & \bf 70.3 \\
    KazakhMMLU $_{\text{(Acc)}}$ & 5-shot & 52.1 & 43.1 & \underline{52.6} & \bf 53.1 & 44.7 & 47.4 & 52.1 & 47.1 & \underline{62.3} & 49.6 & \bf 62.6 \\
    IndoMMLU $_{\text{(Acc)}}$ & 0-shot & \bf 51.0 & 41.5 & \underline{49.0} & \bf 51.0 & 47.0 & 43.7 & 48.5 & 52.0 & \bf 60.1 & 51.0 & \underline{59.9} \\
    IndoCareer $_{\text{(Acc)}}$ & 3-shot & \bf 53.9 & 46.7 & \underline{53.0} & 52.1 & 48.6 & 47.7 & 53.4 & 56.6 & \bf 61.5 & 55.2 & \bf 61.5 \\
    IndoCulture $_{\text{(Acc)}}$ & 0-shot & \underline{51.6} & 49.8 & 51.3 & \bf 57.4 & 50.1 & 44.5 & 59.1 & 58.5 & \underline{61.1} & 57.6 & \bf 62.3 \\
    \bf Average & - & \bf 55.9 & 45.5 & 53.7 & \underline{55.6} & 47.9 & 49.2 & 54.8 & 57.0 & \bf 65.6 & 54.4 & \underline{65.0} \\
    \bottomrule
\end{tabular}}
\caption{Comparison among \marcosmall \& \marcomedium and other open-source base models of comparable sizes and training costs. The best and second best results are marked in \textbf{bold} and \underline{underline}. }
\label{tab:main_results}
\end{table}

\marcosmall is also strong relative to its active size. With only 0.6B activated parameters, it achieves an English average of 57.5, outperforming several larger baselines, including Llama3.2-3B, Gemma3-4B, and Tiny-Aya-3.35B. It is especially competitive on AGIEval, CommonsenseQA, and MMLU-Pro, indicating that the proposed design preserves substantial reasoning and commonsense capacity even in a highly compact regime. This is a meaningful result: \marcosmall is efficient without collapsing on core English capability.

\boldtitle{Multilingual general results.}
The multilingual general benchmarks show the clearest advantage of the \marcomoe family. \marcomedium achieves the best average score (50.9), outperforming all baselines, including Qwen3-4B-Base (48.3). It ranks first on a broad set of tasks covering multilingual knowledge, reasoning, and translation, including GlobalMMLU (64.2), MMMLU (62.0), mHellaSwag (58.6), mARC-Challenge (45.4), FLORES-200 and WMT24$++$ in both directions. On MGSM, it achieves 75.6, effectively matching the strongest baseline. This consistent performance indicates that \marcomedium is a well-balanced multilingual model rather than one optimized for a narrow subset of tasks. \marcosmall is also highly competitive, achieving an average of 43.7. Notably, it obtains the best scores on BELEBELLE and MGSM among all compared models, while remaining strong on the rest. Given its small activated footprint, these results highlight particularly strong cross-lingual reasoning efficiency.

\boldtitle{Multilingual cultural and regional results.}
The cultural and regional benchmarks provide the most demanding test of whether multilingual performance extends beyond broad language transfer into localized knowledge and culturally grounded reasoning. On this suite, \marcomedium achieves an average score of 65.0, ranking second overall and very close to Qwen3-4B-Base (65.7), while outperforming all remaining baselines by a clear margin. The model is particularly strong on Global-PIQA (72.3), ArabicMMLU (67.1), TurkishMMLU (62.7), GreekMMLU (70.3), KazakhMMLU (62.6), and IndoCulture (62.3), where it achieves the best score or is effectively tied for best. This broad coverage across multiple linguistic and cultural contexts suggests that \marcomedium captures more than generic multilingual transfer; it also learns substantial region-specific knowledge and localized reasoning ability.

\marcosmall also demonstrates strong performance for its size, achieving an average of 55.6, which is competitive with Qwen3-1.7B-Base (55.9) and stronger than several larger baselines, including Llama3.2-3B-Base (47.9), SmolLM3-3B-Base (49.2), and Gemma3-4B-Base (54.8). It performs well on INCLUDE (53.2), Global-PIQA (64.3), TurkishMMLU (48.9), GreekMMLU (64.1), KazakhMMLU (53.1), and IndoCulture (57.4). These results indicate that \marcosmall maintains robust regional and cultural generalization despite its compact active architecture, making it a strong option for efficient multilingual deployment across diverse language communities.

There are, however, some clear weaknesses: CMMLU and C-Eval remain two difficult benchmarks for the Marco models, where Qwen3-4B-Base holds a substantial advantage. This suggests that highly localized, exam-style Chinese knowledge is one of the few areas where the strongest competing baseline remains ahead, indicating that \marcomedium may still underperform specialized competitors on certain culturally concentrated or exam-style evaluations. Nevertheless, the broader pattern remains favorable: \marcomedium is one of the strongest culturally grounded multilingual models in the comparison, and \marcosmall is unusually capable for its size.

\boldtitle{Summary.}
\marcomedium is the strongest overall model in the comparison. It achieves the best average on English and multilingual general evaluation, and near-best performance on multilingual cultural/regional benchmarks. This establishes it as a highly capable multilingual foundation model rather than a narrowly optimized system. Besides, \marcosmall defines a compelling compact regime. With only 0.6B activated parameters, it remains competitive on English, strong on multilingual benchmarks, and particularly effective on multilingual reasoning. This places it on a favorable efficiency frontier for multilingual deployment.

More importantly, the results show that the Marco design scales in the right direction: increasing model capacity from Nano to Mini yields large and consistent gains, especially in multilingual transfer, without sacrificing compute efficiency. The remaining gaps are concentrated in a small number of specialized benchmarks rather than reflecting a broader weakness. Overall, the experiments demonstrate that the \marcomoe family delivers state-of-the-art multilingual performance with strong English capability and excellent compute efficiency.

\subsubsection{Analysis}

\begin{figure*}[t]
    \centering
    \includegraphics[width=\textwidth]{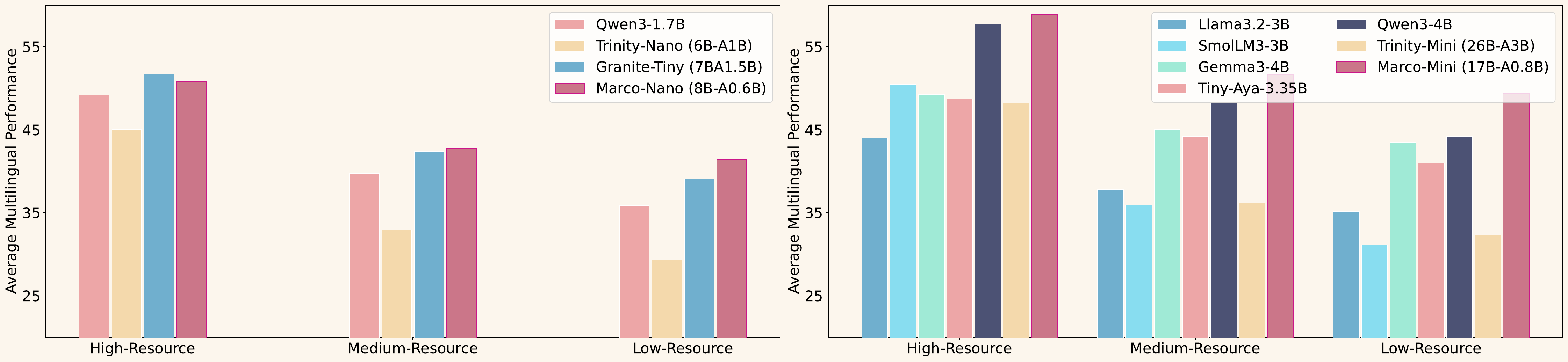}
    \caption{Comparison between \marcomoe models and baselines across different language resource levels.}
    \label{fig:result_by_resource_level}
\end{figure*}
\boldtitle{\marcomoe models excel in long-tail languages.}
Figure~\ref{fig:result_by_resource_level} demonstrates the superior multilingual capabilities of the \marcomoe models, particularly in long-tail languages. In the approximately 1B-parameter category, \marcosmall consistently outperforms counterparts such as Granite4-Tiny and Qwen3-1.7B, which have more activated parameters, and the performance gap widens significantly as language resource availability decreases. A similar trend is observed at the 3B-4B scale, where \marcomedium achieves the highest overall performance across all resource tiers, with a substantial lead in low-resource languages over established models.

\begin{table}
\setlength{\tabcolsep}{15pt}
\footnotesize
\centering
\resizebox{0.8\linewidth}{!}{
\begin{tabular}{l|cccc}
\toprule
     & \bf Stage-1 & \bf Stage-2 & \bf Stage-3 & \bf Stage-4 \\
    \midrule
    English & 59.4 & 63.5 & \bf 63.7 & \bf 63.7 \\
    General Multilingual & 40.6 & 46.3 & 50.3 & \bf 52.1 \\
    Cultural \& Regional & 54.0 & 59.8 & 63.2 & \bf 65.0 \\
    \bottomrule
\end{tabular}}
\caption{Average performance on different benchmark categories after each pre-training stage. Per-benchmark results are shown in Appendix Table~\ref{tab:phase_results_detail}.}
\label{tab:phase_results}
\end{table}
\boldtitle{Multi-stage pre-training gradually improves downstream performance.}
As shown in Table~\ref{tab:phase_results}, the four-stage pre-training curriculum yields consistent improvements across all evaluation dimensions. From Stage-1 to Stage-2, performance increases substantially on every benchmark category, highlighting the importance of the learning-rate decay phase for enhancing overall model capability. From Stage-3 onward, both general multilingual and cultural \& regional multilingual performance continue to improve, while English performance largely saturates, suggesting that the data-mixture switch introduced in Stage-3 effectively strengthens multilingual capability without degrading English performance. Overall, these results confirm the effectiveness of the multi-stage curriculum: each stage provides complementary gains, and the later stages are particularly important for improving multilingual coverage and culturally grounded knowledge.

\begin{figure*}[t]
    \centering
    \includegraphics[width=0.8\textwidth]{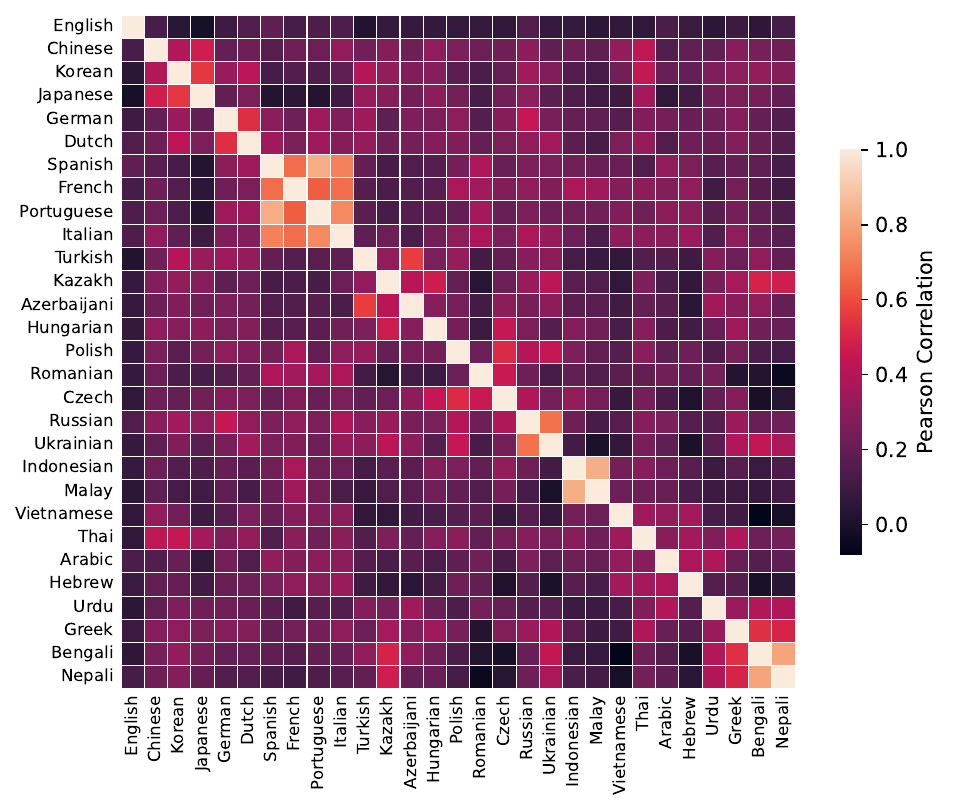}
    \caption{Pearson correlation heatmap between language-specific expert activation patterns. Structured experts are shared among related languages (\eg Romance, East Asian, and Austronesian), while low-correlation pairs highlight specialized expert utilization.}
    \label{fig:language_correlation_by_expert_activation_heatmap}
\end{figure*}
\boldtitle{MoE expert activation patterns resemble language similarity.}
To investigate the degree to which MoE routing dynamics align with established linguistic taxonomies, we formalize the concept of a \emph{Language-Expert Signature}. For each language $L$, we randomly sample 100 documents with $N_L$ tokens from the FineWeb-2 corpus and perform an inference pass to extract the routing statistics. We define the specialization score for an expert $E_i$ at layer $l$ as the proportion of tokens from $L$ that get routed to a particular expert $E_i$:
$$\text{Specialization}(E_{i,l}, L) = \frac{N_{E_{i,l}, L}}{N_L},$$
where $N_{E_{i,l}, L}$ denotes the frequency with which expert $E_i$ at layer $l$ is activated for language $L$. This procedure yields a language-specific activation matrix $\mathbf{M}_L \in \mathbb{R}^{D \times E}$, where $D$ and $E$ represent the total number of layers and experts per layer, respectively. To quantify the functional similarity between any two languages $L_i$ and $L_j$, we flatten them into vectors to represent the expert activation patterns for each language across all layers and expert indices. We then compute the Pearson Correlation Coefficient between these vectors to measure their alignment.

Figure~\ref{fig:language_correlation_by_expert_activation_heatmap} visualizes the pairwise correlation between languages based on their expert activation signatures. There exists a high-correlation cluster for Romance languages (Spanish, French, Portuguese, and Italian), suggesting that the router leverages shared morphological and syntactic structures by assigning them to common expert pools. Similar clustering is observed in the Slavic (Russian and Ukrainian), Austronesian (Indonesian and Malay), and Indic (Bengali and Nepali) groups, underscoring the model's ability to exploit cross-lingual transfer.

Conversely, languages written in unique scripts or isolated grammars, such as Thai, Vietnamese, Arabic, and Hebrew, exhibit low correlations with other languages. This indicates that the router has effectively partitioned specialized expert subsets to handle unique linguistic features, thereby minimizing language interference. Interestingly, English remains relatively isolated, suggesting that the model may dedicate a distinct, high-capacity expert pool to its primary training language to maintain representational precision. In Appendix Figure~\ref{fig:hiererchical_language_clustering_via_expert_activation}, we demonstrate that this pairwise correlation can be used for language clustering, with the results largely mirroring established linguistic family structures.

\begin{table}[H]
\setlength{\tabcolsep}{10pt}
\footnotesize
\centering
\resizebox{0.8\linewidth}{!}{
\begin{tabular}{lc|cccc}
\toprule
    \multirow{2}{*}[0ex]{\bf Benchmark $_{(\text{Metric})}$} & \multirow{2}{*}[0ex]{\bf\# Shots} & \bf Gemma3 & \bf Tiny-Aya & \bf Qwen3 & \bf Marco Mini\\
    & & \bf 4B Base & \bf 3.35B Base & \bf 4B Base &\bf Global Base \\
    \midrule
    \# Activated Params & - & 4B & 3.35B & 4B & 0.86B \\
    \# Total Params & - & 4B & 3.35B & 4B & 17.3B \\
    \# Train FLOPs ($\times10^{23}$) & - & 0.96 & 1.21 & 8.64 & 1.584 \\
    \midrule
    \multicolumn{6}{l}{\bf English} \\
    MMLU $_{\text{(Acc)}}$ & 5-shot & 61.1 & 58.6 & \bf 75.2 & 72.9 \\
    MMLU-Redux $_{\text{(Acc)}}$ & 0-shot & 57.7 & 51.7 & \bf 71.3 & 68.9 \\
    MMLU-Pro $_{\text{(Acc)}}$ & 5-shot & 28.8 & 26.9 & \bf 45.9 & 44.5 \\
    AGIEval $_{\text{(Acc)}}$ & 0-shot & 32.6 & 29.0 & \bf 44.0 & 41.0 \\
    BBH $_{\text{(EM)}}$ & 3-shot & 52.2 & 46.8 & \bf 72.3 & 65.0 \\
    ARC-Easy $_{\text{(Acc)}}$ & 0-shot & \bf 82.6 & 76.5 & 75.0 & 82.4 \\
    ARC-Challenge $_{\text{(Acc)}}$ & 0-shot & 54.1 & 47.4 & 49.9 & \bf 57.0 \\
    HellaSwag $_{\text{(Acc)}}$ & 0-shot & 76.7 & 71.0 & 74.4 & \bf 77.2 \\
    WinoGrande $_{\text{(Acc)}}$ & 0-shot & \bf 61.4 & 56.6 & 59.6 & 58.3 \\
    BoolQ $_{\text{(Acc)}}$ & 0-shot & 76.6 & 74.6 & 74.2 & 75.6 \\
    CommonsenseQA $_{\text{(Acc)}}$ & 0-shot & 61.1 & 60.4 & 52.9 & \bf 61.2 \\
    OpenBookQA $_{\text{(Acc)}}$ & 0-shot & 42.6 & 40.4 & 42.6 & \bf 45.0 \\
    PIQA $_{\text{(Acc)}}$ & 0-shot & 80.3 & 76.9 & 77.4 & \bf 80.7 \\
    SIQA $_{\text{(Acc)}}$ & 0-shot & 50.4 & 49.9 & \bf 53.0 & 48.4 \\
    GSM8K $_{\text{(EM)}}$ & 5-shot & 39.3 & 58.0 & \bf 81.7 & 76.4 \\
    \bf Average & - & 57.2 & 55.5 & 63.3 & \bf 63.6 \\
    
    \midrule
    \multicolumn{6}{l}{\bf Multilingual -- General} \\
    GlobalMMLU $_{\text{(Acc)}}$ & 5-shot & 49.1 & 48.4 & 57.8 & \bf 60.9 \\
    MMMLU $_{\text{(Acc)}}$ & 0-shot & 45.0 & 42.8 & 54.8 & \bf 58.2 \\
    MMLU-ProX-Lite $_{\text{(Acc)}}$ & 5-shot & 23.3 & 23.5 & 35.6 & \bf 36.2 \\
    BELEBELE $_{\text{(Acc)}}$ & 0-shot & 62.3 & 62.5 & 74.0 & \bf 76.0 \\
    mHellaSwag $_{\text{(Acc\_norm)}}$ & 0-shot & 51.9 & 50.3 & 48.5 & \bf 54.4 \\
    mARC-Challenge $_{\text{(Acc\_norm)}}$ & 0-shot & 39.3 & 35.7 & 39.3 & \bf 41.2 \\
    FLORES-200 (En-Xx) $_{\text{(BLEU)}}$ & 5-shot & 27.9 & 25.6 & 25.8 & \bf 29.5 \\
    FLORES-200 (Xx-En) $_{\text{(BLEU)}}$ & 5-shot & 39.2 & 37.2 & 33.4 & \bf 40.2 \\
    WMT24$++$ (En-Xx) $_{\text{(BLEU)}}$ & 5-shot & \bf 26.0 & 24.4 & 19.6 & \bf 26.0 \\
    WMT24$++$ (Xx-En) $_{\text{(BLEU)}}$ & 5-shot & 34.4 & 32.9 & 31.2 & \bf 34.5 \\
    MGSM $_{\text{(EM)}}$ & 8-shot & 35.7 & 36.6 & 69.1 & \bf 71.7 \\
    \bf Average & - & 39.5 & 37.3 & 44.5 & \bf 48.1 \\
    
    \midrule
    \multicolumn{6}{l}{\bf Multilingual -- Cultural \& Regional} \\
    INCLUDE $_{\text{(Acc)}}$ & 5-shot & 52.3 & 53.5 & 60.0 & \bf 61.1 \\
    Global-PIQA $_{\text{(Acc\_norm)}}$ & 0-shot & 67.8 & 66.7 & 61.8 & \bf 70.2 \\
    CMMLU $_{\text{(Acc)}}$ & 5-shot & 50.2 & 58.8 & \bf 76.2 & 67.9 \\
    C-Eval $_{\text{(Acc)}}$ & 5-shot & 48.5 & 57.6 & \bf 76.6 & 66.2 \\
    ArabicMMLU $_{\text{(Acc)}}$ & 3-shot & 61.6 & 63.2 & \bf 67.0 & 66.6 \\
    TurkishMMLU $_{\text{(Acc)}}$ & 5-shot & 43.7 & 45.2 & 60.6 & \bf 63.1 \\
    GreekMMLU $_{\text{(Acc)}}$ & 5-shot & 63.4 & 66.3 & 69.4 & \bf 70.4 \\
    KazakhMMLU $_{\text{(Acc)}}$ & 5-shot & 52.1 & 47.1 & \bf 62.3 & 61.8 \\
    IndoMMLU $_{\text{(Acc)}}$ & 0-shot & 48.5 & 52.0 & \bf 60.1 & 59.5 \\
    IndoCareer $_{\text{(Acc)}}$ & 3-shot & 53.4 & 56.6 & 61.5 & \bf 61.8 \\
    IndoCulture $_{\text{(Acc)}}$ & 0-shot & 59.1 & 58.5 & 61.1 & \bf 62.5 \\
    \bf Average & - & 54.6 & 56.9 & \bf 65.1 & 64.7 \\
    \bottomrule
\end{tabular}}
\caption{Comparing \marcomediumglobal on different benchmark categories with baselines on 64 languages.}
\label{tab:main_results_64langs}
\end{table}
\subsubsection{Scaling up to 64 Languages}
To evaluate the scalability of the Marco-MoE framework, we expand the model's linguistic coverage from 29 core languages to 64 languages with 35 newly introduced languages: Danish, Swedish, Norwegian, Catalan, Galician, Welsh, Irish, Basque, Croatian, Latvian, Lithuanian, Slovak, Slovenian, Estonian, Finnish, Serbian, Bulgarian, Persian, Maltese, Hindi, Marathi, Gujarati, Punjabi, Tamil, Telugu, Tagalog, Javanese, Khmer, Lao, Burmese, Amharic, Swahili, Yoruba, Igbo, Zulu. Following the multi-stage curriculum, we branch from the Stage-2 checkpoint and recalibrate the data mixtures in Stages 3 and 4 to integrate pre-training corpora for the 35 newly introduced languages. This variant, designated as \marcomediumglobal, was trained on an additional 1.4T tokens and consistently demonstrates superior performance over comparable baselines as shown in Table~\ref{tab:main_results_64langs}. Notably, \marcomediumglobal preserves robust English proficiency, achieving 63.6 compared to the 63.7 average of the original \marcomedium, while simultaneously increasing the performance advantage in multilingual tasks from 2.6\% to 3.6\% relative to Qwen3-4B-Base.

\section{Post-Training}
We employ a two-stage post-training pipeline, comprising supervised fine-tuning (SFT) and on-policy distillation~\citep{lu2025onpolicydistillation} from high-capacity teacher models, to develop \emph{general} \textsc{Instruct} models. This pipeline produces models for low-latency responses to standard user queries by bypassing explicit, computationally intensive reasoning processes.

\subsection{Supervised Fine-Tuning}

\subsubsection{Data}

\boldtitle{General Instruction Data.}
Our SFT process leverages the Dolci-Instruct dataset~\citep{olmo2025olmo3}, originally employed to train the \textsc{Olmo}3-Instruct models. To further enhance the instruction-following capabilities of \marcomoe, we augment this corpus with additional data sourced from \citet{yang2026nemotroncascade2posttrainingllms}.

\boldtitle{Knowledge-Intensive Data.}
We observe that scientific data is underrepresented within the Dolci-Instruct dataset. To address this gap, we augment the corpus by sourcing scientific prompts from~\citet{yang2026nemotroncascade2posttrainingllms} and distilling the corresponding responses from \href{https://storage.googleapis.com/deepmind-media/Model-Cards/Gemini-3-Flash-Model-Card.pdf}{Gemini3-Flash}.

\boldtitle{Translation Data.}
For translation data, our data curation pipeline begins by sourcing parallel corpora from the web-mined NLLB translation dataset~\citep{nllbteam2022languageleftbehindscaling}. To ensure high data fidelity, we implement a multi-stage filtering strategy. First, we apply heuristic text filters adapted from Gopher~\citep{rae2022scalinglanguagemodelsmethods} and FineWeb~\citep{penedo2024finewebdatasetsdecantingweb} to the English segments to exclude low-quality pairs. To further enhance semantic alignment, we leverage the Qwen3-Embedding-8B model~\citep{zhang2025qwen3embeddingadvancingtext} to compute relevance scores for each translation pair. Ultimately, we retain only the top 10,000 highest-scoring pairs per language.

\boldtitle{Multilingual \& Cultural Data.}
To generate culturally-grounded data, we leverage Wikidata~\citep{wikidata} as our primary knowledge source within a multi-stage generation pipeline. Initially, we identify a taxonomy of topics pertinent to regional and cultural contexts. For each target language, we then filter entities within these topics, selecting only those whose "\emph{country}" property aligns with the corresponding linguistic region. Utilizing these retrieved entities and their associated metadata, we employ Gemini3-Flash to synthesize coherent textual descriptions. Building on these descriptions, we adapt a back-translation approach~\citep{li2024selfalignment} to prompt the model to derive corresponding instructions. Recognizing that descriptions derived solely from structured properties can be overly formulaic or information-sparse, we further task Gemini3-Flash with augmenting these instructions by recalling relevant background knowledge, thereby producing more natural and comprehensive responses. This process yields a high-quality, diverse dataset of instruction-response pairs tailored to the cultural nuances of each language. In addition, we augment these pairs by transforming them into multiple-choice question-answering (QA) formats with concise step-by-step chain-of-thoughts to further enhance training data diversity.

\subsubsection{Implementation Details}
We fine-tune \marcomoe utilizing the SLIME framework \citep{slime_github}. The learning rate is initialized at $1\times 10^{-5}$ and follows a cosine decay schedule, reaching a minimum of $1\times 10^{-6}$. The model is trained for a single epoch with a global batch size of 512, totaling approximately 4,000 training steps. We set the maximum context length to 8,192 tokens. Optimization is performed using AdamW with a weight decay of 0.1, $\beta_1 = 0.9$, and $\beta_2 = 0.95$. The whole SFT process takes around 24 hours using 64 GPUs.
\subsection{On-Policy Distillation from Stronger Models}
On-policy learning involves training a model on samples generated by its own current policy, fundamentally enabling it to correct its own errors. While this approach is a cornerstone of reinforcement learning (RL), it is often constrained by sparse feedback; RL typically yields a limited amount of information per training episode regardless of sequence length~\citep{schulman2025lora}, resulting in significant sample inefficiency and high computational overhead. In contrast, off-policy learning seeks to \textit{distill} the behavior of a superior teacher model. By leveraging supervision for every token produced by the teacher, off-policy methods provide substantially denser learning signals. However, this paradigm is susceptible to distribution shift: the student model is trained on trajectories native to the teacher rather than those it will encounter during its own inference. This can lead to compounding errors, where early deviations from the teacher's distribution cause the model to diverge into unfamiliar states. To bridge these two paradigms, On-Policy Distillation (OPD)~\citep{lu2025onpolicydistillation,yang2025qwen3technicalreport} utilizes trajectories sampled from the student model while incorporating the dense per-token supervision characteristic of off-policy distillation. Consequently, OPD allows the model to learn from its own generations while maintaining the high signal density required for efficient optimization.

\begin{table}
\setlength{\tabcolsep}{10pt}
\footnotesize
\centering
\resizebox{0.85\linewidth}{!}{
\begin{tabular}{l|c|c}
\toprule
    Category & Datasets & Mixture Ratio \\
    \midrule
    \multirow{2}{*}[0ex]{Instruction Following} & \href{https://huggingface.co/datasets/nvidia/Nemotron-RL-instruction\_following}{Nemotron-RL-instruction-following} & \multirow{2}{*}[0ex]{25\%} \\
    & \href{https://huggingface.co/datasets/nvidia/Nemotron-RL-instruction\_following-structured\_outputs}{Nemotron-RL-instruction-following-structured-outputs} & \\
    \midrule
    \multirow{2}{*}[0ex]{Konwledge \& Reasoning} & \href{https://huggingface.co/datasets/nvidia/Nemotron-RL-ReasoningGym-v1}{Nemotron-RL-ReasoningGym-v1} & \multirow{2}{*}[0ex]{25\%} \\
    & \href{https://huggingface.co/datasets/nvidia/Nemotron-RL-knowledge-mcqa}{Nemotron-RL-knowledge-mcqa} & \\
    \midrule
    Alignment & \href{https://huggingface.co/datasets/nvidia/Nemotron-Cascade-RL-RLHF}{Nemotron-Cascade-RL-RLHF} & 10\% \\
    \midrule
    \multirow{2}{*}[0ex]{Math} & \href{https://huggingface.co/datasets/BytedTsinghua-SIA/DAPO-Math-17k}{DAPO-Math-17k} & \multirow{2}{*}[0ex]{10\%} \\
    & \href{https://huggingface.co/datasets/Skywork/Skywork-OR1-RL-Data}{Skywork-OR1-RL-Data} & \\
    \midrule
    \multirow{3}{*}[0ex]{Multilingual} & Translation & \multirow{3}{*}[0ex]{30\%} \\
    & Multilingual \& Cultural Data & \\
    & \href{https://huggingface.co/datasets/nvidia/Nemotron-SFT-Multilingual-v1}{Nemotron-SFT-Multilingual-v1} & \\
    \bottomrule
\end{tabular}}
\caption{Datasets and mixture ratios for the on-policy distillation stage.}
\label{tab:opd_data_mixtures}
\end{table}
\subsubsection{Data}

We employ a diverse set of data sources to post-train the SFT checkpoint using OPD, with the exact data mixtures shown in Table~\ref{tab:opd_data_mixtures}.

\boldtitle{Instruction Following.}
We sample instruction-following prompts from the \href{https://huggingface.co/datasets/nvidia/Nemotron-RL-instruction\_following}{Nemotron-RL-instruction-following} dataset. To further enhance the ability of \marcomoe in following specific output formats and generating structured data, we further augment this set with prompts from \href{https://huggingface.co/datasets/nvidia/Nemotron-RL-instruction\_following-structured\_outputs}{Nemotron-RL-instruction-following-structured-outputs}.

\boldtitle{Knowledge \& Reasoning.}
We gather a diverse collection of datasets focused on reasoning and knowledge grounding. Specifically, we sample reasoning tasks from \href{https://huggingface.co/datasets/nvidia/Nemotron-RL-ReasoningGym-v1}{Nemotron-RL-ReasoningGym-v1}, which encompasses a wide array of domains, including algebra, geometry, and graph theory. Furthermore, we incorporate \href{https://huggingface.co/datasets/nvidia/Nemotron-RL-knowledge-mcqa}{Nemotron-RL-knowledge-mcqa}, a dataset comprising knowledge-intensive multiple-choice question-answering (MCQA) samples across various scientific disciplines.

\boldtitle{Alignment.}
To facilitate human preference alignment and enhance conversational proficiency, we utilize alignment prompts from the \href{https://huggingface.co/datasets/nvidia/Nemotron-Cascade-RL-RLHF}{Nemotron-Cascade-RL-RLHF} dataset to enhance \marcomoe.

\boldtitle{Math.}
While \marcomoe is not explicitly optimized for mathematical reasoning, we incorporate the DAPO-Math-17k~\citep{yu2025dapoopensourcellmreinforcement} and Skywork-OR1-RL-Data~\citep{he2025skyworkopenreasoner1} datasets to enhance the model’s general problem-solving proficiency.

\boldtitle{Multilingual.}
The multilingual corpus incorporates a subsample of our curated translation and cultural data. We further augment this with the \href{https://huggingface.co/datasets/nvidia/Nemotron-SFT-Multilingual-v1}{Nemotron-SFT-Multilingual-v1} dataset, which provides high-quality translations of mathematical and scientific problems.

\subsubsection{Implementation Details}
We implement OPD within the SLIME framework and adopt a cascaded distillation strategy that progressively transfers knowledge from increasingly stronger teacher models. Specifically, for both \marcosmallins and \marcomediumins models, we first use \href{https://huggingface.co/Qwen/Qwen3-30B-A3B-Instruct-2507}{Qwen3-30B-A3B-Instruct-2507} as the initial teacher. After the student model converges, we switch to the stronger \href{https://huggingface.co/Qwen/Qwen3-Next-80B-A3B-Instruct}{Qwen3-Next-80B-A3B-Instruct} and continue distillation. Training uses a constant learning rate of $1\times10^{-6}$ and a global batch size of 512. To improve training efficiency, we sample only two responses per prompt, resulting in 1,024 sequences per training step. The maximum sequence lengths are set to 2,048 tokens for prompts and 8,192 tokens for responses.
The complete OPD procedure for \marcosmallins involves approximately 1,900 steps with Qwen3-30B-A3B-Instruct-2507 followed by 1,000 steps with Qwen3-Next-80B-A3B-Instruct; \marcomediumins undergoes 1,900 steps for each phase. The total training process requires roughly 110 hours on a cluster of 64 GPUs.
\subsection{Instruct Model Evaluation}

\subsubsection{Experimental Settings}
\boldtitle{Evaluation Benchmarks.}
Following base mode evaluation settings, all models are conducted using the Light-Eval framework~\citep{lighteval} to ensure fair comparisons. The evaluation suite is divided into two primary dimensions:
\begin{compactenum}
    \item English Proficiency: We assess general language understanding and reasoning via diverse benchmarks, including MMLU~\citep{hendrycks2021measuring}, MMLU-Redux~\citep{gema-etal-2025-done}, MMLU-Pro~\citep{wang2024mmlupro}, AGIEval~\citep{zhong-etal-2024-agieval}, GPQA-Diamond~\citep{rein2023gpqagraduatelevelgoogleproofqa}, GSM8K~\citep{cobbe2021trainingverifierssolvemath}, and MATH~\citep{hendrycks2021measuringmathematicalproblemsolving}.
    \item Multilingual Capability: To measure performance across diverse languages, we evaluate the model on both general multilingual benchmarks and datasets assessing cultural and regional knowledge. General datasets include GlobalMMLU~\citep{singh-etal-2025-global}, MMLU-ProX-Lite~\citep{xuan-etal-2025-mmlu},  \href{https://huggingface.co/datasets/openai/MMMLU}{MMMLU}, MGPQA~\citep{huang2025benchmax}, MGSM~\citep{shi2023language}, PolyMath~\citep{wang2025polymath} and translation tasks with FLORES-200~\citep{nllbteam2022languageleftbehindscaling} and WMT24$++$~\citep{freitag-etal-2024-llms}. For cultural and regional benchmarks, we evaluate the models on INCLUDE~\citep{romanou2025include}, Global-PIQA~\citep{chang2025globalpiqaevaluatingphysical}, CMMLU~\citep{li-etal-2024-cmmlu},  C-Eval~\citep{huang2023ceval}, ArabicMMLU~\citep{koto-etal-2024-arabicmmlu}, TurkishMMLU~\citep{yuksel-etal-2024-turkishmmlu}, GreekMMLU~\citep{zhang2026greekmmlunativesourcedmultitaskbenchmark}, KazakhMMLU~\citep{togmanov-etal-2025-kazmmlu}, IndoMMLU~\citep{koto-etal-2023-large}, IndoCareer~\citep{koto-2025-cracking}, and IndoCulture~\citep{koto-etal-2024-indoculture}.
\end{compactenum}

\boldtitle{Baselines.}
We compare \marcosmallins and \marcomediumins against various open-source \textsc{Instruct} models. The comparison set includes dense models: Qwen3-1.7B-Instruct (\ie non-thinking), Qwen3-4B-Instruct-2507~\citep{yang2025qwen3technicalreport}, Qwen3-VL-2B-Instruct~\citep{bai2025qwen3vltechnicalreport}, Ministral-3-3B-Instruct-2512 \& Ministral-3-8B-Instruct-2512~\citep{liu2026ministral3}, and Gemma3-12B-Instruct~\citep{gemmateam2025gemma3technicalreport}; MoE models: Granite4-Tiny \& Granite4-Small~\citep{granite2025}, and LFM2-8B-A1B \& LFM2-24B-A2B~\citep{amini2025lfm2technicalreport}.
All models are evaluated using a sampling temperature of $1.0$ and a top-p value of $1.0$. For most benchmarks, we generate eight responses per test instance with a maximum response length of 8,192 and report the mean accuracy. However, for GlobalMMLU and MMMLU, we only sample a single response per query to optimize computational efficiency, given the substantially larger scale of these datasets.

\begin{table}
\setlength{\tabcolsep}{2pt}
\footnotesize
\centering
\resizebox{\linewidth}{!}{
\begin{tabular}{l|cccccc|cccccc}
\toprule
    \multirow{3}{*}[0ex]{\bf Benchmark $_{(\text{Metric})}$} & \bf Qwen3 & \bf Qwen3-VL & \bf Ministral3 & \bf LFM2 & \bf Granite4 & \bf Marco & \bf Qwen3 & \bf Ministral3 & \bf Gemma3 & \bf Granite4 & \bf LFM2 & \bf Marco \\
    & \bf 1.7B & \bf 2B & \bf 3B & \bf 8B-A1B & \bf Tiny & \bf Nano & \bf 4B & \bf 8B & \bf 12B & \bf Small & \bf 24B-A2B & \bf Mini \\
    & \scriptsize Instruct & \scriptsize Instruct & \scriptsize Instruct & \scriptsize Instruct & \scriptsize Instruct & \scriptsize Instruct & \scriptsize Instruct & \scriptsize Instruct & \scriptsize Instruct & \scriptsize Instruct & \scriptsize Instruct & \scriptsize Instruct \\
    \midrule
    \# Activated Params & 1.7B & 2B & 3.84B & 1.5B & 1.47B & 0.6B & 4B & 8.8B & 12B & 9B & 2B & 0.86B \\
    \# Total Params & 1.7B & 2B & 3.84B & 8.3B & 7B & 8B & 4B & 8.8B & 12B & 32B & 24B & 17.3B \\
    \midrule
    \multicolumn{10}{l}{\bf English} \\
    MMLU $_{\text{(Acc)}}$ & 62.4 & 62.1 & 69.8 & \underline{72.1} & 50.8 & \bf 73.2 & \underline{80.8} & 79.8 & 76.2 & 76.7 & 74.9 & \bf 83.4 \\
    MMLU-Redux $_{\text{(Acc)}}$ & 62.4 & 62.2 & 69.6 & \underline{71.9} & 51.2 & \bf 73.3 & \underline{80.9} & 79.9 & 76.2 & 76.7 & 74.9 & \bf 83.5 \\
    MMLU-Pro $_{\text{(Acc)}}$ & 35.2 & 38.3 & \underline{49.5} & \underline{49.5} & 25.3 & \bf 54.5 & \underline{66.9} & 63.9 & 55.8 & 57.1 & 57.6 & \bf 70.7 \\
    AGIEval $_{\text{(Acc)}}$ & 39.6 & 33.0 & 44.7 & \underline{45.2} & 30.7 & \bf 49.8 & 51.7 & \underline{52.4} & 43.6 & 44.7 & 49.0 & \bf 55.4 \\
    GPQA-Diamond $_{\text{(Acc)}}$ & 27.5 & 21.0 & \underline{31.6} & \bf 31.9 & 28.3 & 22.2 & \bf 50.8 & 44.8 & 35.2 & 38.6 & 39.7 & \underline{50.3} \\
    GSM8K $_{\text{(EM)}}$ & 77.9 & 79.7 & 79.0 & \underline{84.6} & 71.1 & \bf 86.7 & 88.6 & 89.5 & \underline{89.7} & 83.9 & 87.2 & \bf 93.1 \\
    MATH $_{\text{(EM)}}$ & 70.6 & 73.7 & 70.2 & \bf 82.6 & 53.4 & \underline{79.6} & \bf 93.4 & 86.2 & 83.8 & 75.7 & 83.9 & \underline{91.8} \\
    \bf Average & 53.7 & 52.9 & 59.2 & \underline{62.5} & 44.4 & \bf 62.8 & \underline{73.3} & 70.9 & 65.8 & 64.8 & 66.7 & \bf 75.5 \\

    \midrule
    \multicolumn{10}{l}{\bf Multilingual -- General} \\
    GlobalMMLU $_{\text{(Acc)}}$ & 46.3 & 45.9 & 38.4 & \underline{49.0} & 43.0 & \bf 58.7 & \underline{70.2} & 55.4 & 69.2 & 67.4 & 57.0 & \bf 73.3 \\
    MMMLU $_{\text{(Acc)}}$ & 49.0 & 49.0 & 39.4 & \underline{56.5} & 44.1 & \bf 59.9 & \underline{71.3} & 56.4 & 69.4 & 68.1 & 62.3 & \bf 73.7 \\
    MMLU-ProX-Lite $_{\text{(Acc)}}$ & 28.6 & 30.3 & 26.7 & \underline{33.8} & 22.1 & \bf 43.2 & \underline{58.3} & 43.3 & 51.3 & 51.6 & 43.3 & \bf 61.2 \\ 
    MGPQA $_{\text{(Acc)}}$ & 25.3 & 22.3 & 18.8 & \bf 27.2 & \underline{25.9} & 21.6 & \underline{41.0} & 30.5 & 32.8 & 35.0 & 32.7 & \bf 41.8 \\ 
    FLORES-200 (En-Xx) $_{\text{(BLEU)}}$ & 12.7 & 15.3 & 8.3 & 14.9 & \bf 22.5 & \underline{22.3} & 22.1 & 17.5 & \bf 35.6 & 31.9 & 19.2 & \underline{30.6} \\
    FLORES-200 (Xx-En) $_{\text{(BLEU)}}$ & 28.2 & 28.6 & 18.9 & 20.1 & \underline{30.4} & \bf 31.1 & 33.5 & 31.0 & \bf 40.3 & 32.2 & 22.7 & \underline{36.8} \\
    WMT24$++$ (En-Xx) $_{\text{(BLEU)}}$ & 13.2 & 14.6 & 4.4 & 14.6 & \bf 18.9 & \underline{18.7} & 20.9 & 14.4 & \bf 32.1 & 26.6 & 16.0 & \underline{26.8} \\
    WMT24$++$ (Xx-En) $_{\text{(BLEU)}}$ & \underline{26.4} & 26.2 & 8.3 & 17.9 & 25.1 & \bf 27.3 & 29.9 & 24.2 & \bf 35.5 & 27.5 & 18.8 & \underline{31.3} \\
    MGSM $_{\text{(EM)}}$ & 63.6 & \underline{67.6} & 47.0 & 56.5 & 55.3 & \bf 76.5 & \underline{84.4} & 68.7 & 84.0 & 75.7 & 67.8 & \bf 87.4 \\
    PolyMath $_{\text{(EM)}}$ & 23.4 & 25.5 & 16.3 & \underline{26.5} & 18.7 & \bf 29.6 & \bf 47.2 & 26.4 & 35.5 & 28.9 & 29.3 & \underline{44.7} \\
    \bf Average & 31.7 & \underline{32.5} & 22.7 & 31.7 & 30.6 & \bf 38.9 & 47.9 & 36.8 & \underline{48.6} & 44.5 & 36.9 & \bf 50.8 \\

    \midrule
    \multicolumn{10}{l}{\bf Multilingual -- Cultural \& Regional} \\
    INCLUDE $_{\text{(Acc)}}$ & \underline{44.9} & 44.4 & 35.4 & 43.5 & 38.6 & \bf 54.3 & 63.8 & 50.7 & \underline{65.0} & 60.3 & 49.1 & \bf 65.6 \\
    Global-PIQA $_{\text{(Acc)}}$ & 62.0 & \underline{65.8} & 50.6 & 60.8 & 63.3 & \bf 70.7 & 79.6 & 61.3 & \underline{82.2} & 80.2 & 69.0 & \bf 84.2 \\
    CMMLU $_{\text{(Acc)}}$ & \underline{60.4} & \bf 63.3 & 48.9 & 52.7 & 39.2 & 60.0 & \bf 78.6 & 67.4 & 60.8 & 59.6 & 56.7 & \underline{75.3} \\
    C-Eval $_{\text{(Acc)}}$ & 58.7 & \bf 63.2 & 50.6 & 50.8 & 39.4 & \underline{60.8} & \bf 80.4 & 68.0 & 59.7 & 59.4 & 56.7 & \underline{75.4} \\
    ArabicMMLU $_{\text{(Acc)}}$ & \underline{48.8} & 46.9 & 22.7 & \bf 56.5 & 43.4 & \bf 56.5 & 66.0 & 41.4 & \bf 70.1 & 66.3 & 61.3 & \underline{67.8} \\
    TurkishMMLU $_{\text{(Acc)}}$ & \underline{42.7} & 39.6 & 38.6 & 26.3 & 31.6 & \bf 59.9 & \underline{71.6} & 48.2 & 64.4 & 57.9 & 33.4 & \bf 74.7 \\
    GreekMMLU $_{\text{(Acc)}}$ & \underline{48.7} & 48.0 & 38.4 & 40.0 & 44.8 & \bf 61.6 & 68.6 & 49.5 & \bf 77.7 & 71.7 & 44.7 & \underline{72.5} \\
    KazakhMMLU $_{\text{(Acc)}}$ & 46.0 & \underline{47.1} & 41.4 & 39.6 & 39.6 & \bf 56.3 & 66.6 & 59.1 & \underline{66.8} & 63.5 & 47.6 & \bf 68.8 \\
    IndoMMLU $_{\text{(Acc)}}$ & 48.8 & \underline{49.3} & 35.2 & 41.1 & 37.2 & \bf 56.3 & 64.4 & 52.4 & \underline{65.3} & 59.6 & 42.7 & \bf 65.7 \\
    IndoCareer $_{\text{(Acc)}}$ & \underline{46.1} & 45.7 & 36.0 & 41.7 & 34.7 & \bf 54.9 & 62.2 & 53.4 & \underline{63.2} & 56.3 & 43.7 & \bf 64.4 \\
    IndoCulture $_{\text{(Acc)}}$ & 45.8 & \underline{47.7} & 37.2 & 45.9 & 42.8 & \bf 59.1 & 58.7 & 47.8 & \bf 69.6 & 59.3 & 44.2 & \underline{67.1} \\
    \bf Average & 50.3 & \underline{51.0} & 39.5 & 45.4 & 41.3 & \bf 59.1 & \underline{69.1} & 54.5 & 67.7 & 63.1 & 49.9 & \bf 71.0 \\

    \bottomrule
\end{tabular}}
\caption{Comparison among \marcosmallins \& \marcomediumins and other open-source Instruct models of comparable sizes. The best and second best results are marked in \textbf{bold} and \underline{underline}. $\text{Avg}@8$ accuracies are reported, except for GlobalMMLU and MMMLU where $\text{Acc}@1$ is reported.}
\label{tab:instruct_main_results}
\end{table}

\subsubsection{Main Results}
We demonstrate the results of \textsc{Instruct} models in Table~\ref{tab:instruct_main_results}.

\boldtitle{English Results.}
On English benchmarks, \marcomediumins achieves an average score of 75.5, effectively surpassing Qwen3-4B-Instruct (73.3) and significantly outperforming Ministral3-8B-Instruct (70.9) and Gemma3-12B-Instruct (65.8), despite utilizing a fraction of the activated parameters. This efficiency is even more pronounced when compared to other sparse models; \marcomediumins significantly outperforms LFM2-24B-A2B (66.7) and Granite-4-Small (64.8), even though these baselines activate 2B and 9B parameters, respectively. Similarly, \marcosmallins defines a highly efficient frontier, achieving an average score of 62.8. It notably surpasses LFM2-8B-A1B (62.5) while requiring only 40\% of its activated parameter budget (0.6B \versus 1.5B). 

\boldtitle{Multilingual General Results.}
The advantages of the \marcomoe models are most pronounced in general multilingual capabilities, where \marcomediumins ranks first among all evaluated models with an average score of 50.8. It not only surpasses Qwen3-4B-Instruct but also demonstrates nearly double the efficiency of comparable MoE competitors. For instance, \marcomediumins outperforms LFM2-24B-A2B by an absolute point of 12.4, despite having 38\% fewer total parameters and 57\% fewer activated parameters. \marcosmallins also shows remarkable cross-lingual transfer, achieving 38.9 and outperforming not only baselines with comparable size but also the significantly larger LFM2-24B-A2B variant.

\boldtitle{Multilingual Cultural and Regional Results.}
Regarding localized knowledge and culturally grounded reasoning, \marcomediumins emerges as the top-performing model with an average of 71.0, outperforming Qwen3-4B-Instruct (69.1) and the substantially larger Gemma3-12B-Instruct (67.7). The performance gap between \marcomoe and other MoE baselines widens substantially in this category, with \marcomediumins outperforming LFM2-24B-A2B-Instruct (49.9) by 21.1 points. Even the compact \marcosmallins (59.1) demonstrates superior cultural proficiency relative to both LFM2-8B-A1B and LFM2-24B-A2B models. Nevertheless, \marcomoe models still lag behind Qwen3 models in highly localized Chinese knowledge reasoning, as evidenced by lower relative performance on the CMMLU and C-Eval benchmarks.

\begin{table}
\setlength{\tabcolsep}{10pt}
\footnotesize
\centering
\resizebox{0.8\linewidth}{!}{
\begin{tabular}{l|ccc|ccc}
\toprule
    & \multicolumn{3}{c|}{\bf \marcosmallins} & \multicolumn{3}{c}{\bf \marcomediumins} \\
    \cmidrule(lr){2-4} \cmidrule(lr){5-7}
    & \bf Stage-1 & \bf Stage-2 & $\bf \Delta$ & \bf Stage-1 & \bf Stage-2 & $\bf \Delta$ \\
    \midrule
    English & 60.5 & 62.8 & \bf +2.3 & 73.7 & 75.5 & \bf +1.8 \\
    General Multilingual & 38.0 & 38.9 & \bf +0.9 & 49.3 & 50.8 & \bf +1.5 \\
    Cultural \& Regional & 58.6 & 59.1 & \bf +0.5 & 69.9 & 71.0 & \bf +1.1 \\
    \bottomrule
\end{tabular}}
\caption{Performance progression across the two-stage cascaded OPD. The $\Delta$ indicates the absolute improvement from Stage-1 (utilizing the 30B-A3B teacher) to Stage-2 (transitioning to the 80B-A3B teacher) across three benchmark categories. Per-benchmark results are shown in Appendix Table~\ref{tab:per_benchmark_cascaded_opd_results}.}
\label{tab:cascaded_opd_results}
\end{table}
\subsubsection{The Effects of Cascaded Distillation}
Table~\ref{tab:cascaded_opd_results} demonstrate the efficacy of our cascaded distillation strategy, which progressively transfers knowledge from increasingly capable teacher models. Both \marcosmallins and \marcomediumins models achieve consistent gains across all evaluation axes upon transitioning from the 30B-A3B to the 80B-A3B teacher.
This progressive improvement validates that the student models have not reached their capacity limits after the first stage. Instead, OPD from a stronger teacher provides higher-quality reward signals that allow the model to correct more sophisticated errors.
\section{Related Work}
\boldtitle{MoE Model Upcycling.}
The computational overhead of training large-scale Mixture-of-Experts (MoE) models from scratch has led to the emergence of MoE Upcycling~\citep{komatsuzaki2023sparse}, a paradigm that initializes sparse models from pre-trained dense checkpoints. Prominent methods, such as Sparse Upcycling (\eg Mixtral 8x7B~\citep{jiang2024mixtralexperts}), typically utilize coarse-grained expert construction. In these frameworks, the feed-forward networks (FFNs) of a dense transformer are replicated in their entirety to form the expert pool, followed by continued pre-training. Because all experts are identical at initialization, this approach may impede the emergence of specialized representations. While recent strategies attempt to break this symmetry by injecting random noise during replication~\citep{yang2024qwen2technicalreport, nakamura2025dropupcycling}, the fundamental units of the model remain monolithic.

While computationally convenient, coarse-grained upcycling stands in tension with recent architectural trends suggesting that a large number of fine-grained experts is more effective than a few coarse-grained experts under an equivalent parameter budget~\citep{dai-etal-2024-deepseekmoe}. Our method shifts toward fine-grained expert upcycling. Rather than duplicating entire FFN blocks, we decompose dense weights into modular sub-components via sub-matrix splitting before expanding them into an expert ensemble. Our method mitigates the redundancy bottlenecks inherent in monolithic block replication and facilitates expert specialization.

\boldtitle{Multilingual LLMs}
The development of multilingual Large Language Models (LLMs) is fundamentally constrained by the "curse of multilinguality"~\citep{conneau-etal-2020-unsupervised}, wherein increasing language coverage within a fixed-parameter budget leads to performance degradation due to capacity bottlenecks and cross-lingual interference. While cross-lingual transfer can benefit low-resource languages~\citep{joshi-etal-2020-state, nigatu-etal-2024-zenos}, it is most effective between typologically related languages~\citep{he-etal-2025-scaling}, making it difficult for small-scale models to maintain high performance across diverse families~\citep{ustun-etal-2024-aya}.

Recent state-of-the-art compact models have adopted varying strategies to mitigate these constraints: Qwen3-4B~\citep{yang2025qwen3technicalreport} and Gemma3-4B~\citep{gemmateam2025gemma3technicalreport} utilize massive pre-training (up to 36T tokens) and sophisticated strong-to-weak distillation and claim to support 119 and 140+ languages, respectively. In contrast, models such as SmolLM3~\citep{bakouch2025smollm3} prioritize per-language depth over breadth, focusing exclusively on a narrow subset of high-resource languages. Other efforts, including SEA-LION~\citep{tjhi-etal-2023-sea}, Sailor2~\citep{dou2025sailor2sailingsoutheastasia}, EuroLLM~\citep{martins2024eurollmmultilinguallanguagemodels, ramos2026eurollm22btechnicalreport}, and Franken-Adapter~\citep{jiang2025frankenadaptercrosslingualadaptationllms}, address capacity limitations through regional specialization. While extreme cases like Apertus~\citep{apertus2025apertusdemocratizingopencompliant} expand coverage to over 1,000 languages, they often encounter substantial performance trade-offs on standard benchmarks.

Despite these advances, most existing small-scale multilingual models rely on dense architectures, which inherently struggle to balance breadth and proficiency. Our work introduces the first compact, highly-sparse Multilingual MoE models with a fully open pre-training recipe. By leveraging the conditional computation of MoE, our method effectively breaks the curse of multilinguality, scaling support to 64 languages without the performance degradation typical of dense counterparts. This architectural shift facilitates specialized expert activation patterns that preserve the integrity of individual languages while maximizing cross-lingual transfer across related languages.

\boldtitle{Post-Training and On-Policy Distillation.}
Standard post-training typically commences with Supervised Fine-Tuning (SFT) utilizing high-quality instruction-response pairs to establish foundational instruction-following capabilities~\citep{wei2022finetunedlanguagemodelszeroshot,chung2022scalinginstructionfinetunedlanguagemodels}. However, SFT frequently suffers from exposure bias and significant distribution shift~\citep{gudibande2023falsepromiseimitatingproprietary}. Because the model is trained exclusively on trajectories native to the teacher rather than those encountered during its own inference, it often lacks the robustness required to recover from compounding errors during auto-regressive generation.

To address these distribution mismatches, on-policy learning, the basic element of reinforcement learning (RL), has emerged as an effective paradigm. In this framework, the model is trained on samples generated by its own current policy, providing a principled method for interactive alignment by exposing the student to its actual state distribution. This allows the model to learn directly from its own sampled trajectories rather than relying on external, static data.

While RL techniques like PPO~\citep{schulman2017proximalpolicyoptimizationalgorithms} or GRPO~\citep{shao2024deepseekmathpushinglimitsmathematical} offer powerful optimization, they are often constrained by sparse reward signals and significant computational overhead. In contrast, On-Policy Distillation (OPD)~\citep{lu2025onpolicydistillation,yang2025qwen3technicalreport} provides substantially higher signal density that enables efficient optimization comparable to RL performance at a reduced compute cost.
In our work, we implement an OPD pipeline, adopting a cascaded distillation strategy. By progressively transferring knowledge from increasingly capable teachers, we ensure the iterative refinement of multilingual and reasoning capabilities while maintaining superior sample efficiency.

\section{Conclusion and Discussion}
In this work, we introduced \marcomoe, a family of fully open-source, multilingual, and highly sparse MoE models that effectively mitigate the capacity bottlenecks inherent in dense architectures. By employing a fine-grained upcycling strategy, we successfully converted pre-trained dense models into highly sparse MoE layers that activate only around 5\% of parameters per token. Our base models, \marcosmall and \marcomedium, demonstrate state-of-the-art performance across English and multilingual benchmarks, establishing a superior performance-to-compute ratio over similarly sized dense and MoE baselines. Leveraging these foundations, we developed \marcomoe-\textsc{Instruct} variants through a lightweight two-stage post-training pipeline. Empirical results indicate that \marcosmallins and \marcomediumins consistently outperform competing architectures possessing $3$--$14\times$ more activated parameters. 
Furthermore, our analysis elucidates that \marcomoe naturally learns structured expert activation signatures that mirror established linguistic family structures, facilitating the scalable expansion to 64 languages while minimizing the cross-lingual interference typical of dense counterparts. By disclosing our complete four-stage training curriculum, comprehensive datasets, and model weights, we aim to provide a transparent and high-performance foundation for future multilingual research.

While we have demonstrated the efficient scaling of \marcomoe to 64 languages, this coverage encompasses only a small fraction of global linguistic diversity, particularly compared to massively multilingual models that target over 1,000 languages. Our reliance on translation-based and synthetic data pipelines to boost performance in low-resource settings, while empirically validated, underscores the persistent scarcity of authentic, high-quality multilingual reasoning data in naturalistic environments. Future research will focus on broadening \marcomoe’s linguistic scope while investigating more sophisticated routing mechanisms to enhance expert specialization for extremely long-tail languages. Furthermore, because our current framework necessitates retraining the full model to integrate new languages, developing methods for modular, incremental expansion remains a compelling direction for efficient language scaling.

\bibliography{anthology,custom}

\clearpage
\appendix
\section{Per-Benchmark Results across Pre-training Phases}
\begin{table}[H]
\setlength{\tabcolsep}{15pt}
\footnotesize
\centering
\resizebox{0.8\linewidth}{!}{
\begin{tabular}{l|cccc}
\toprule
    \bf Benchmark $_{(\text{Metric})}$ & \bf Stage-1 & \bf Stage-2 & \bf Stage-3 & \bf Stage-4 \\
    \midrule
    \multicolumn{5}{l}{\bf English} \\
    MMLU $_{\text{(Acc)}}$ & 66.1 & 72.8 & \bf 73.5 & 72.8 \\
    MMLU-Redux $_{\text{(Acc)}}$ & 63.0 & \bf 69.4 & 68.9 & 68.8 \\
    MMLU-Pro $_{\text{(Acc)}}$ & 35.6 & 42.7 & 44.8 & \bf 45.3 \\
    AGIEval $_{\text{(Acc)}}$ & 37.5 & 40.6 & \bf 41.9 & \bf 41.9 \\
    BBH $_{\text{(EM)}}$ & 55.6 & 63.7 & 62.0 & \bf 65.1 \\
    ARC-Easy $_{\text{(Acc)}}$ & 80.1 & 82.5 & \bf 82.7 & 82.4 \\
    ARC-Challenge $_{\text{(Acc)}}$ & 53.1 & \bf 57.4 & \bf 57.4 & 56.3 \\
    HellaSwag $_{\text{(Acc)}}$ & 73.0 & \bf 77.6 & 77.5 & 77.4 \\
    WinoGrande $_{\text{(Acc)}}$ & 55.5 & \bf 59.0 & 58.0 & 57.7 \\
    BoolQ $_{\text{(Acc)}}$ & 74.5 & 74.9 & \bf 76.1 & 74.2 \\
    CommonsenseQA $_{\text{(Acc)}}$ & 59.1 & 61.3 & \bf 61.5 & \bf 61.5 \\
    OpenBookQA $_{\text{(Acc)}}$ & 44.2 & 44.6 & \bf 44.8 & 44.6 \\
    PIQA $_{\text{(Acc)}}$ & 79.2 & 79.9 & 80.7 & \bf 81.1 \\
    SIQA $_{\text{(Acc)}}$ & 47.9 & 50.6 & 49.0 & \bf 49.4 \\
    GSM8K $_{\text{(EM)}}$ & 66.3 & 75.0 & 75.9 & \bf 76.4 \\
    \bf Average & 59.4 & 63.5 & \bf 63.7 & \bf 63.7 \\
    
    \midrule
    \multicolumn{5}{l}{\bf Multilingual -- General} \\
    GlobalMMLU $_{\text{(Acc)}}$ & 52.4 & 59.5 & 62.6 & \bf 64.2 \\
    MMMLU $_{\text{(Acc)}}$ & 51.5 & 59.3 & 60.5 & \bf 62.0 \\
    MMLU-ProX-Lite $_{\text{(Acc)}}$ & 29.2 & 34.2 & 38.2 & \bf 39.2 \\
    BELEBELE $_{\text{(Acc)}}$ & 63.4 & 70.5 & \bf 80.1 & 79.8 \\
    mHellaSwag $_{\text{(Acc\_norm)}}$ & 51.4 & 55.2 & 58.2 & \bf 58.6 \\
    mARC-Challenge $_{\text{(Acc\_norm)}}$ & 39.1 & 42.3 & 44.8 & \bf 45.4 \\
    FLORES-200 (En-Xx) $_{\text{(BLEU)}}$ & 22.3 & 25.6 & 31.7 & \bf 32.3 \\
    FLORES-200 (Xx-En) $_{\text{(BLEU)}}$ & 30.2 & 34.5 & 40.0 & \bf 40.1 \\
    WMT24$++$ (En-Xx) $_{\text{(BLEU)}}$ & 19.8 & 23.8 & 27.7 & \bf 28.1 \\
    WMT24$++$ (Xx-En) $_{\text{(BLEU)}}$ & 26.5 & 30.9 & \bf 34.4 & \bf 34.4 \\
    MGSM $_{\text{(EM)}}$ & 61.0 & 73.1 & 75.1 & \bf 75.6 \\
    \bf Average & 40.6 & 46.3 & 50.3 & \bf 52.1 \\
    
    \midrule
    \multicolumn{5}{l}{\bf Multilingual -- Cultural \& Regional} \\
    INCLUDE $_{\text{(Acc)}}$ & 51.5 & 56.7 & 61.0 & \bf 61.7 \\
    Global-PIQA $_{\text{(Acc\_norm)}}$ & 64.9 & 66.6 & 70.6 & \bf 72.3 \\
    CMMLU $_{\text{(Acc)}}$ & 58.6 & 63.8 & 64.2 & \bf 68.0 \\
    C-Eval $_{\text{(Acc)}}$ & 57.1 & 62.9 & 63.5 & \bf 66.0 \\
    ArabicMMLU $_{\text{(Acc)}}$ & 60.3 & 65.8 & 66.6 & \bf 67.1 \\
    TurkishMMLU $_{\text{(Acc)}}$ & 50.1 & 60.1 & 61.2 & \bf 62.7 \\
    GreekMMLU $_{\text{(Acc)}}$ & 41.8 & 51.6 & 68.1 & \bf 70.3 \\
    KazakhMMLU $_{\text{(Acc)}}$ & 47.9 & 54.3 & 60.5 & \bf 62.6 \\
    IndoMMLU $_{\text{(Acc)}}$ & 51.1 & 56.3 & 59.1 & \bf 59.9 \\
    IndoCareer $_{\text{(Acc)}}$ & 54.8 & 60.1 & 60.3 & \bf 61.5 \\
    IndoCulture $_{\text{(Acc)}}$ & 55.6 & 60.0 & 60.5 & \bf 62.3 \\
    \bf Average & 54.0 & 59.8 & 63.2 & \bf 65.0 \\
    \bottomrule
\end{tabular}}
\caption{Per-benchmark model performance across the four pre-training stages. }
\label{tab:phase_results_detail}
\end{table}

\clearpage
\section{Per-Benchmark Results across the Two-Stage Cascaded OPD}
\begin{table}[H]
\setlength{\tabcolsep}{10pt}
\footnotesize
\centering
\resizebox{0.8\linewidth}{!}{
\begin{tabular}{l|ccc|ccc}
\toprule
    & \multicolumn{3}{c|}{\bf \marcosmallins} & \multicolumn{3}{c}{\bf \marcomediumins} \\
    \cmidrule(lr){2-4} \cmidrule(lr){5-7}
    & \bf Stage-1 & \bf Stage-2 & $\bf \Delta$ & \bf Stage-1 & \bf Stage-2 & $\bf \Delta$ \\
    \midrule
    \multicolumn{7}{l}{\bf English} \\
    MMLU $_{\text{(Acc)}}$ & 71.5 & 73.2 & \bf +1.7 & 82.1 & 83.4 & \bf +1.3 \\
    MMLU-Redux $_{\text{(Acc)}}$ & 71.7 & 73.3 & \bf +1.6 & 82.0 & 83.5 & \bf +1.5 \\
    MMLU-Pro $_{\text{(Acc)}}$ & 51.4 & 54.5 & \bf +3.1 & 67.6 & 70.7 & \bf +3.1 \\
    AGIEval $_{\text{(Acc)}}$ & 48.4 & 49.8 & \bf +1.4 & 54.6 & 55.4 & \bf +0.8 \\
    GPQA-Diamond $_{\text{(Acc)}}$ & 19.4 & 22.2 & \bf +2.8 & 47.2 & 50.3 & \bf +3.1 \\
    GSM8K $_{\text{(EM)}}$ & 85.4 & 86.7 & \bf +1.3 & 92.1 & 93.1 & \bf +1.0 \\
    MATH $_{\text{(EM)}}$ & 75.7 & 79.6 & \bf +3.9 & 90.2 & 91.8 & \bf +1.6 \\
    \bf Average & 60.5 & 62.8 & \bf +2.3 & 73.7 & 75.5 & \bf +1.8 \\

    \midrule
    \multicolumn{7}{l}{\bf Multilingual -- General} \\
    GlobalMMLU $_{\text{(Acc)}}$ & 58.6 & 58.7 & \bf +0.1 & 71.8 & 73.3 & \bf +1.5 \\
    MMMLU $_{\text{(Acc)}}$ & 59.6 & 59.9 & \bf +0.3 & 72.4 & 73.7 & \bf +1.3 \\
    MMLU-ProX-Lite $_{\text{(Acc)}}$ & 41.9 & 43.2 & \bf +1.3 & 57.2 & 61.2 & \bf +4.0 \\ 
    MGPQA $_{\text{(Acc)}}$ & 20.5 & 21.6 & \bf +1.1 & 39.8 & 41.8 & \bf +2.0 \\ 
    FLORES-200 (En-Xx) $_{\text{(BLEU)}}$ & 21.5 & 22.3 & \bf +0.8 & 29.9 & 30.6 & \bf +0.7 \\
    FLORES-200 (Xx-En) $_{\text{(BLEU)}}$ & 31.5 & 31.1 & $\bf\color{red}{-}$\bf\color{red}{0.4} & 36.9 & 36.8 & $\bf\color{red}{-}$\bf\color{red}{0.1} \\
    WMT24$++$ (En-Xx) $_{\text{(BLEU)}}$ & 17.8 & 18.7 & \bf +0.9 & 26.3 & 26.8 & \bf +0.5 \\
    WMT24$++$ (Xx-En) $_{\text{(BLEU)}}$ & 27.2 & 27.3 & \bf +0.1 & 31.9 & 31.3 & $\bf\color{red}{-}$\bf\color{red}{0.6} \\
    MGSM $_{\text{(EM)}}$ & 74.4 & 76.5 & \bf +2.1 & 85.8 & 87.4 & \bf +1.6 \\
    PolyMath $_{\text{(EM)}}$ & 27.3 & 29.6 & \bf +2.3 & 41.4 & 44.7 & \bf +3.3 \\
    \bf Average & 38.0 & 38.9 & \bf +0.9 & 49.3 & 50.8 & \bf +1.5 \\

    \midrule
    \multicolumn{7}{l}{\bf Multilingual -- Cultural \& Regional} \\
    INCLUDE $_{\text{(Acc)}}$ & 54.1 & 54.3 & \bf +0.2 & 64.5 & 65.6 & \bf +1.1 \\
    Global-PIQA $_{\text{(Acc)}}$ & 69.3 & 70.7 & \bf +1.4 & 83.7 & 84.2 & \bf +0.5 \\
    CMMLU $_{\text{(Acc)}}$ & 60.1 & 60.0 & $\bf\color{red}{-}$\bf\color{red}{0.1} & 74.4 & 75.3 & \bf +0.9 \\
    C-Eval $_{\text{(Acc)}}$ & 60.2 & 60.8 & \bf +0.6 & 74.1 & 75.4 & \bf +1.3 \\
    ArabicMMLU $_{\text{(Acc)}}$ & 56.8 & 56.5 & $\bf\color{red}{-}$\bf\color{red}{0.3} & 66.7 & 67.8 & \bf +1.1 \\
    TurkishMMLU $_{\text{(Acc)}}$ & 58.7 & 59.9 & \bf +1.2 & 73.3 & 74.7 & \bf +1.4 \\
    GreekMMLU $_{\text{(Acc)}}$ & 61.6 & 61.6 & 0.0 & 71.0 & 72.5 & \bf +1.5 \\
    KazakhMMLU $_{\text{(Acc)}}$ & 56.0 & 56.3 & \bf +0.3 & 67.4 & 68.8 & \bf +1.4 \\
    IndoMMLU $_{\text{(Acc)}}$ & 56.4 & 56.3 & $\bf\color{red}{-}$\bf\color{red}{0.1} & 64.7 & 65.7 & \bf +1.0 \\
    IndoCareer $_{\text{(Acc)}}$ & 53.8 & 54.9 & \bf +1.1 & 63.3 & 64.4 & \bf +1.1 \\
    IndoCulture $_{\text{(Acc)}}$ & 57.8 & 59.1 & \bf +1.3 & 66.0 & 67.1 & \bf +1.1 \\
    \bf Average & 58.6 & 59.1 & \bf +0.5 & 69.9 & 71.0 & \bf +1.1 \\

    \bottomrule
\end{tabular}}
\caption{Per-benchmark performance progression across the two-stage cascaded OPD. The $\Delta$ indicates the absolute improvement from Stage-1 (utilizing the 30B-A3B teacher) to Stage-2 (transitioning to the 80B-A3B teacher) across each benchmark.}
\label{tab:per_benchmark_cascaded_opd_results}
\end{table}

\clearpage
\section{Language Clustering through Expert Activation Patterns}
\begin{figure*}[t]
    \setlength{\abovecaptionskip}{-0.005cm}
    \centering
    \includegraphics[width=\textwidth]{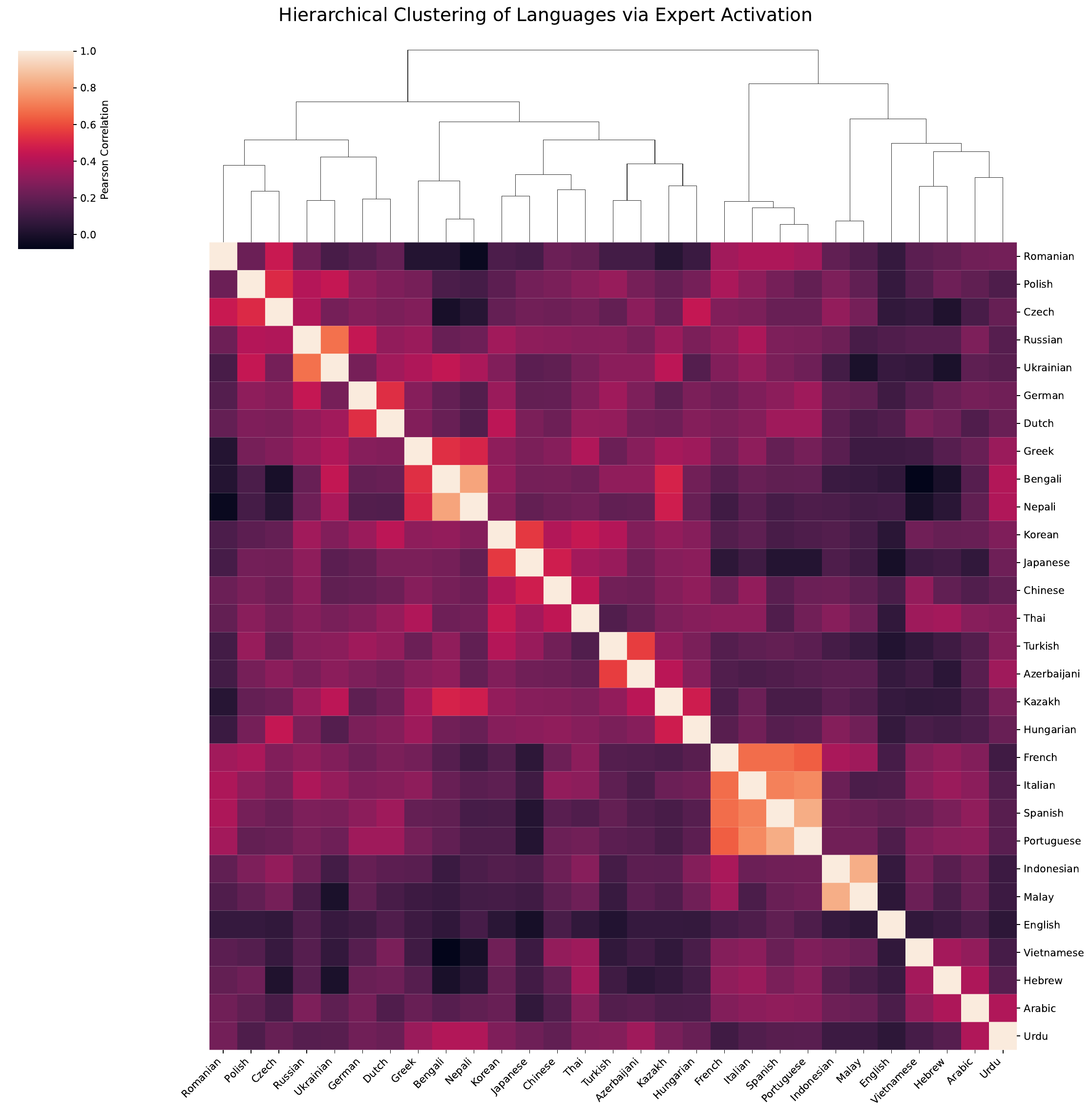}
    \caption{Hierarchical clustering of expert activation patterns, demonstrating that the model's neural routing trajectories tightly mirror established linguistic family structures. The emergence of distinct blocks for Romance, Slavic, Germanic, and East Asian groups illustrates how the MoE architecture naturally discovers and exploits cross-lingual commonalities.}
    \label{fig:hiererchical_language_clustering_via_expert_activation}
\end{figure*}
To further corroborate that expert routing in \marcomoe encodes linguistically meaningful structure, we perform agglomerative hierarchical clustering based on the Language-Expert Signatures shown in Figure~\ref{fig:language_correlation_by_expert_activation_heatmap}. Specifically, we treat each language's flattened activation vector $\mathbf{M}_L$ as a point in expert-activation space, adopt $1 - \rho(\mathbf{M}_{L_i}, \mathbf{M}_{L_j})$ (with $\rho$ the Pearson correlation coefficient) as the pairwise distance between language $L_i$ and $L_j$, and apply average-linkage clustering to construct the dendrogram shown in Figure~\ref{fig:hiererchical_language_clustering_via_expert_activation}. The resulting hierarchy tightly mirrors established linguistic taxonomies: Romance, Slavic, Germanic, and East Asian languages each coalesce into compact subtrees at small merge distances, while at deeper levels phylogenetically related branches further converge into broader super-clusters that echo the Indo-European macro-family. Languages with isolated typology or unique scripts, in contrast, attach to the tree only at large distances, reflecting the router's allocation of dedicated expert subsets to accommodate their distinctive morphology. This unsupervised rediscovery of language families offers mechanistic evidence that \marcomoe has learned to exploit cross-lingual regularities in a manner that mirrors human-curated linguistic phylogenies, promoting positive transfer among related languages while isolating typologically distant ones from mutual interference.

\clearpage
\section{Synthetic Data Generation Prompts}
\begin{tcolorbox}[
    colback=gray!5, 
    colframe=black, 
    width=\textwidth, 
    arc=2mm, 
    boxrule=0.5pt, 
    title=\textbf{Translation Prompt},
    fonttitle=\sffamily\bfseries,
    colbacktitle=gray!20,
    coltitle=black,
    enhanced,
    label={prompt:translation_prompt}
]
\small
Translate the following text into $[\text{Language}]$, ensuring that all LaTeX symbols, code snippets, and any related formatting structures remain unchanged and intact. Only translate the non-code and non-symbol portions of the text, preserving all technical formatting exactly as presented. Pay careful attention to maintaining the original layout and ensuring the translated text integrates seamlessly with non-translated elements. \\ \\ $[\text{Document}]$
\end{tcolorbox}
\begin{tcolorbox}[
    colback=gray!5, 
    colframe=black, 
    width=\textwidth, 
    arc=2mm, 
    boxrule=0.5pt, 
    title=\textbf{Regional Web Document Annotation Prompt},
    fonttitle=\sffamily\bfseries,
    colbacktitle=gray!20,
    coltitle=black,
    enhanced,
    label={prompt:regional_annotate}
]
\small
\textbf{System Prompt:} \\
You are given a [Document] from the web. \\
Your goal is to: \\
\textbf{[Document]:} \{\} \\

\textbf{Task 1: Training Value Score (1--10)} \\
Analyze the content and assign a score based on quality and trustworthiness for LLM training:
\begin{compactenum}
    \item \textbf{10 (Pristine):} Factually impeccable, encyclopedic (e.g., peer-reviewed papers).
    \item \textbf{9 (Excellent):} Expert-level, well-researched (e.g., reputable news).
    \item \textbf{8 (Very Good):} Reliable, solid blog posts or documentation.
    \item \textbf{7 (Good):} Standard "how-to" guides, clear information.
    \item \textbf{6 (Acceptable):} Average, coherent but superficial (e.g., forum posts).
    \item \textbf{5 (Borderline):} Mixed quality, requires heavy cleaning.
    \item \textbf{4 (Low Quality):} Problematic, high fluff, questionable facts.
    \item \textbf{3 (Very Low Quality):} SEO spam, keyword stuffing, machine-generated.
    \item \textbf{2 (Harmful/Junk):} Hate speech, dangerous theories, malicious code.
    \item \textbf{1 (Toxic):} Unintelligible, entirely devoid of value.
\end{compactenum}

\textbf{Task 2: Subject Classification} \\
Subject Classification: Assign a subject/category to the document based on its content. Choose from the following categories: \\
- Accounting
- Agriculture
- Anthropology
- Architecture and Design
- Arts \& Humanities
- Biology
- Business administration
- Business ethics
- Business
- Chemistry
- Computer Science
- Culturology
- Earth science
- Economics
- Education
- Engineering
- Environmental studies and forestry
- Family and consumer science
- Finance
- Geography
- Health
- History
- Human physical performance and recreation
- Industrial and labor relations
- International trade
- Journalism
- media studies
- and communication
- Language
- Law
- Library and museum studies
- Literature
- Logic
- Management
- Marketing
- Math
- Medicine
- Military Sciences
- Multiple exams
- Performing arts
- Philosophy
- Physics
- Political sciences
- Psychology
- Public Administration
- Public Policy
- Qualimetry
- Religious studies
- Risk management and insurance
- Social work
- Sociology
- STEM
- Transportation
- Visual Arts
- Driving License
- Marine License
- Medical License
- Professional Certifications
- Others \\
\textbf{Task 3: Country/Region Classification} \\
Decide if the content is culturally specific (e.g., Japan) or "No specific country."

\textbf{Output Format:} \\
1. Text Quality Score (1-10): \\
2. Subject Category: \\
3. Country/Region: \\
\textit{Only generate the final result without additional descriptions.}
\end{tcolorbox}
\begin{tcolorbox}[
    colback=gray!5, 
    colframe=black, 
    width=\textwidth, 
    arc=2mm, 
    boxrule=0.5pt, 
    title=\textbf{Regional MCQs Generation Prompt},
    fonttitle=\sffamily\bfseries,
    colbacktitle=gray!20,
    coltitle=black,
    enhanced,
    label={prompt:cultural_mcq_prompt}
]
\small
Write a question on [country], more specifically on [sub\_topic], with up to four choices.

Desired difficulty level: [difficulty].
The required language: [language]

- For questions marked "intermediate", the question should require applying basic concepts independently or in simple combinations. The relationships between concepts should be straightforward and directly tied to the problem. Solutions should be accessible with a foundational understanding and methodical effort. \\
- For questions marked "complex", the question should require identifying and combining multiple clear but non-obvious connections between concepts. The relationships may span different areas within the topic, requiring structured reasoning and careful application of techniques. Focus on recognizing patterns and synthesizing information rather than straightforward computation. \\
- For questions marked "hard", the question should involve intricate or abstract relationships between concepts. Solving them should demand advanced reasoning skills, mastery of topic-specific techniques, and the ability to deal with ambiguities or edge cases. Solutions often require navigating multiple layers of logic or constraints. \\
- For questions marked "brutal", the question should push problem-solving to the limit, requiring mastery of nuanced relationships across diverse concepts. It should demand precise reasoning, significant abstraction, and fluency in advanced techniques. Problems may include nested or multi-step dependencies, where overlooking a detail can derail the solution. \\
- For questions marked "borderline unsolvable", the question should challenge even experts, involving exceptionally intricate or unconventional problem structures. It should combine deep abstraction, subtle constraints, and highly interconnected concepts. While solutions exist, they often require extensive expertise and innovative approaches. Avoid speculative or unsolvable topics (e.g., "Does God exist?" or "How can time travel be achieved?"). \\

Ensure that one of the four choices is correct. Your question must include a valid answer within the given choices; you should not specify which one is correct.

Your response should not contain any headers, explanations, or your answer to the question.

Question: [Question] \\
(A): [Option A] \\
(B): [Option B] \\
(C): [Option C] \\
(D): [Option D]
\end{tcolorbox}
\begin{tcolorbox}[
    colback=gray!5, 
    colframe=black, 
    width=\textwidth, 
    arc=2mm, 
    boxrule=0.5pt, 
    title=\textbf{Regional MCQ Example},
    fonttitle=\sffamily\bfseries,
    colbacktitle=gray!20,
    coltitle=black,
    enhanced,
    breakable,
    label={prompt:cultural_mcq_example}
]
\small
\begin{CJK}{UTF8}{gkai} 
\textbf{Question:} 在《素问·痹论》“痛者，寒气多也，有寒故痛也”的理论基础上，若患者表现为关节剧烈疼痛、如锥刺刀割，且痛处固定不移，遇寒则痛甚，得热则稍减，局部皮色不红、触之不温，甚至出现明显的“筋脉拘急”导致关节屈伸不利。从寒邪“收引凝滞”的病理特性及其对“营卫气血运行受阻”的深度逻辑推导，下列关于“痛痹（寒痹）”发病机制的描述最准确且全面的是：

\begin{description}
    \item[A:] 寒邪凝滞导致血流减慢，其深层逻辑在于寒邪易于诱发“血瘀化热”的病理转型。因此，痛痹的治疗重点应在于清热化瘀，以缓解因热扰筋脉所致的局部红肿与活动受限。
    \item[B:] 寒为阴邪，极易损伤人体阳气，导致卫外之气涣散，使得风邪乘虚深袭经络。其病机关键在于风寒相搏引起的“气行游走”，导致关节局部出现一过性的拘挛与酸楚感。
    \item[C:] 寒邪性主收引，入经则使经脉绌急，入络则使血行凝滞；其剧烈疼痛的根源不仅在于“凝则不通”，更在于寒邪遏阻阳气，使营卫之气在受阻部位产生激烈的“郁遏交争”与脉络极度挛缩。
    \item[D:] 寒邪侵袭导致阳气受损，其病理重点在于肾阳虚衰而不能温煦经脉，导致水饮内停并流注于骨节。临床辨证核心应聚焦于“水饮溢于经络”所致的关节肿胀与沉重感。
\end{description}

\textbf{Answer:} 根据《素问·痹论》的理论及中医病理逻辑，对“痛痹（寒痹）”的发病机制进行如下推理：

\begin{compactenum}
    \item \textbf{寒邪的特性分析}：
    \begin{compactenum}
        \item \textbf{收引 (Constriction)}：寒邪侵袭，导致经络、筋脉缩急。正如文中所述“筋脉拘急”，这是因为寒性收缩，使脉道、筋膜挛缩，从而出现关节屈伸不利。
        \item \textbf{凝滞 (Stagnation)}：寒则气血凝涩不通，“不通则痛”。寒邪入于脉络，使血流减慢甚至瘀阻，形成固定不移的剧烈疼痛。
    \end{compactenum}
    \item \textbf{对营卫气血的影响}：
    \begin{compactenum}
        \item \textbf{营卫受阻}：营行脉中，卫行脉外。寒邪侵入后，卫阳被遏，营血凝滞。
        \item \textbf{郁遏交争}：剧烈疼痛（如锥刺刀割）的深层逻辑在于，寒邪作为阴邪遏制了局部阳气的运行，阳气试图冲破寒邪的阻碍，在受阻部位产生剧烈的搏击、交争（即“郁遏交争”），加之脉络极度挛缩，导致疼痛感远重于其他类型的痹症。
    \end{compactenum}
    \item \textbf{选项辨析}：
    \begin{compactenum}
        \item \textbf{A项}：误导在于“血瘀化热”和“清热化瘀”。虽然长期瘀滞可能化热，但题目明确描述为“皮色不红、触之不温、遇寒痛甚”，属于典型的寒证，治疗应温经散寒，而非清热。
        \item \textbf{B项}：误导在于“风邪深袭”和“气行游走”。“游走”是风痹（行痹）的特点，而本题强调“痛处固定不移”，是寒邪凝滞的表现。
        \item \textbf{C项}：精准地抓住了寒邪“收引”导致“经脉绌急”、“凝滞”导致“血行不畅”的要点。同时，它深入解释了剧烈疼痛的根源在于寒邪遏阻阳气，使营卫在局部产生激烈的郁遏交争，符合“痛者，寒气多也”的深度病理逻辑。
        \item \textbf{D项}：误导在于“水饮内停”和“关节肿胀沉重”。这描述的是湿痹（着痹）的特征，而非痛痹。
    \end{compactenum}
\end{compactenum}

\begin{center}
    $\boxed{\text{C}}$
\end{center}
\end{CJK}
\end{tcolorbox}

\clearpage
\section{Base Model Results by Language}
\begin{table}[H]
\setlength{\tabcolsep}{3pt}
\footnotesize
\setlength{\belowcaptionskip}{-0.35cm}
\centering
\resizebox{\linewidth}{!}{
\begin{tabular}{l|ccccccccccc}
\toprule
    Language & GlobalMMLU & MMMLU & MMLU-ProX-Lite & BELEBELE & mHellaSwag & mARC-Challenge & FLORES-200 & WMT24$++$ & MGSM & INCLUDE & Global-PIQA \\
    \midrule
    eng\_Latn & \cmark & \xmark & \cmark & \cmark & \xmark & \xmark & \xmark & \xmark & \cmark & \xmark & \cmark \\
    zho\_Hans & \cmark & \cmark & \cmark & \cmark & \cmark & \cmark & \cmark & \cmark & \cmark & \cmark & \cmark \\
    por\_Latn & \cmark & \cmark & \cmark & \cmark & \cmark & \cmark & \cmark & \cmark & \xmark & \cmark & \cmark \\
    fra\_Latn & \cmark & \cmark & \cmark & \cmark & \cmark & \cmark & \cmark & \cmark & \cmark & \cmark & \cmark \\
    rus\_Cyrl & \cmark & \xmark & \cmark & \cmark & \cmark & \cmark & \cmark & \cmark & \cmark & \cmark & \cmark \\
    deu\_Latn & \cmark & \cmark & \cmark & \cmark & \cmark & \cmark & \cmark & \cmark & \cmark & \cmark & \cmark \\
    ita\_Latn & \cmark & \cmark & \cmark & \cmark & \cmark & \cmark & \cmark & \cmark & \xmark & \cmark & \cmark \\
    spa\_Latn & \cmark & \cmark & \cmark & \cmark & \cmark & \cmark & \cmark & \cmark & \cmark & \cmark & \cmark \\
    jpn\_Jpan & \cmark & \cmark & \cmark & \cmark & \xmark & \xmark & \cmark & \cmark & \cmark & \cmark & \cmark \\
    kor\_Hang & \cmark & \cmark & \cmark & \cmark & \xmark & \xmark & \cmark & \cmark & \xmark & \cmark & \cmark \\
    arb\_Arab & \cmark & \cmark & \cmark & \cmark & \cmark & \cmark & \cmark & \cmark & \xmark & \cmark & \cmark \\
    ind\_Latn & \cmark & \cmark & \cmark & \cmark & \cmark & \cmark & \cmark & \cmark & \xmark & \cmark & \cmark \\
    vie\_Latn & \cmark & \xmark & \cmark & \cmark & \cmark & \cmark & \cmark & \cmark & \xmark & \cmark & \cmark \\
    ron\_Latn & \cmark & \xmark & \xmark & \cmark & \cmark & \cmark & \cmark & \cmark & \xmark & \xmark & \cmark \\
    tha\_Thai & \xmark & \xmark & \cmark & \cmark & \xmark & \xmark & \cmark & \cmark & \cmark & \xmark & \cmark \\
    zsm\_Latn & \cmark & \xmark & \xmark & \cmark & \xmark & \xmark & \cmark & \xmark & \xmark & \cmark & \cmark \\
    ces\_Latn & \cmark & \xmark & \cmark & \cmark & \xmark & \xmark & \cmark & \cmark & \xmark & \xmark & \cmark \\
    nld\_Latn & \cmark & \xmark & \xmark & \cmark & \cmark & \cmark & \cmark & \cmark & \xmark & \cmark & \cmark \\
    ukr\_Cyrl & \cmark & \xmark & \cmark & \cmark & \cmark & \cmark & \cmark & \cmark & \xmark & \cmark & \cmark \\
    pol\_Latn & \cmark & \xmark & \xmark & \cmark & \xmark & \xmark & \cmark & \cmark & \xmark & \cmark & \cmark \\
    tur\_Latn & \cmark & \xmark & \xmark & \cmark & \xmark & \xmark & \cmark & \cmark & \xmark & \cmark & \cmark \\
    ell\_Grek & \xmark & \xmark & \xmark & \cmark & \xmark & \xmark & \cmark & \cmark & \xmark & \cmark & \cmark \\
    hun\_Latn & \xmark & \xmark & \cmark & \cmark & \cmark & \cmark & \cmark & \cmark & \xmark & \cmark & \cmark \\
    heb\_Hebr & \cmark & \xmark & \xmark & \cmark & \xmark & \xmark & \cmark & \cmark & \xmark & \cmark & \cmark \\
    ben\_Beng & \cmark & \cmark & \cmark & \cmark & \cmark & \cmark & \cmark & \cmark & \cmark & \cmark & \cmark \\
    npi\_Deva & \cmark & \xmark & \xmark & \cmark & \cmark & \cmark & \cmark & \xmark & \xmark & \cmark & \cmark \\
    urd\_Arab & \xmark & \xmark & \cmark & \cmark & \xmark & \xmark & \cmark & \cmark & \xmark & \cmark & \cmark \\
    kaz\_Cyrl & \xmark & \xmark & \xmark & \cmark & \xmark & \xmark & \cmark & \xmark & \xmark & \cmark & \cmark \\
    azj\_Latn & \xmark & \xmark & \xmark & \cmark & \xmark & \xmark & \cmark & \xmark & \xmark & \cmark & \cmark \\
    \bottomrule
\end{tabular}}
\caption{Languages included in each of the evaluated benchmark.}
\label{tab:language_by_dataset}
\end{table}
\begin{table}[H]
\setlength{\tabcolsep}{3pt}
\footnotesize
\setlength{\belowcaptionskip}{-0.35cm}
\centering
\resizebox{\linewidth}{!}{
\begin{tabular}{l|ccccccccccc}
\toprule
    \multirow{2}{*}[0ex]{\bf Language} & \bf Qwen3 & \bf Trinity & \bf Granite4 & \bf Marco & \bf Llama3.2 & \bf SmolLM3 & \bf Gemma3 & \bf Tiny-Aya & \bf Qwen3 & \bf Trinity & \bf Marco \\
    & \bf 1.7B Base & \bf Nano Base & \bf Tiny Base & \bf Nano Base & \bf 3B Base & \bf 3B Base & \bf 4B Base & \bf 3.35B Base & \bf 4B Base & \bf Mini Base & \bf Mini Base \\
    \midrule
    arb\_Arab & 45.5 & 36.3 & 51.6 & 49.0 & 39.3 & 49.5 & 48.1 & 48.5 & \underline{58.2} & 44.0 & \bf 62.9 \\
    ben\_Beng & 39.4 & 33.0 & 47.2 & 43.1 & 34.0 & 31.6 & 46.1 & 44.9 & \underline{50.6} & 38.1 & \bf 55.6 \\
    ces\_Latn & 48.2 & 42.2 & 56.5 & 51.6 & 42.9 & 41.8 & 51.3 & 50.2 & \underline{61.8} & 51.7 & \bf 64.4 \\
    deu\_Latn & 52.6 & 47.5 & 58.1 & 54.7 & 45.8 & 53.3 & 52.8 & 51.6 & \bf 65.3 & 57.6 & \underline{64.9} \\
    eng\_Latn & 61.6 & 61.7 & 65.2 & 62.9 & 54.6 & 60.1 & 58.7 & 56.6 & \bf 71.8 & 68.5 & \underline{71.4} \\
    fra\_Latn & 54.0 & 51.0 & 59.6 & 55.6 & 47.1 & 54.4 & 53.3 & 51.8 & \bf 66.3 & 60.6 & \underline{66.1} \\
    heb\_Hebr & 41.2 & 35.5 & 48.4 & 47.2 & 35.7 & 33.5 & 46.7 & 47.6 & \underline{51.9} & 43.1 & \bf 60.8 \\
    ind\_Latn & 51.8 & 44.3 & 55.8 & 53.3 & 44.9 & 45.8 & 52.0 & 51.1 & \underline{63.9} & 52.8 & \bf 65.6 \\
    ita\_Latn & 53.9 & 48.2 & 59.0 & 55.6 & 47.0 & 54.8 & 54.1 & 52.5 & \underline{66.0} & 57.7 & \bf 66.8 \\
    jpn\_Jpan & 49.7 & 43.3 & 53.7 & 51.4 & 41.1 & 48.0 & 48.6 & 49.0 & \underline{60.7} & 53.2 & \bf 63.4 \\
    kor\_Hang & 47.5 & 40.0 & 52.2 & 49.9 & 40.2 & 46.9 & 48.1 & 49.6 & \underline{59.3} & 49.9 & \bf 61.8 \\
    nld\_Latn & 50.7 & 46.3 & 58.0 & 54.3 & 45.6 & 46.3 & 52.6 & 50.4 & \underline{63.9} & 54.6 & \bf 66.4 \\
    npi\_Deva & 37.5 & 32.4 & 47.8 & 43.6 & 34.5 & 35.9 & 45.2 & 45.5 & \underline{49.7} & 39.9 & \bf 54.7 \\
    pol\_Latn & 48.5 & 42.5 & 53.7 & 51.8 & 42.9 & 42.3 & 50.7 & 48.8 & \underline{60.9} & 51.3 & \bf 65.0 \\
    por\_Latn & 55.0 & 50.1 & 59.7 & 56.3 & 48.0 & 54.9 & 53.6 & 52.8 & \underline{66.2} & 60.3 & \bf 66.8 \\
    ron\_Latn & 50.1 & 45.4 & 56.4 & 53.4 & 44.7 & 45.1 & 52.4 & 50.4 & \underline{63.5} & 53.8 & \bf 65.3 \\
    rus\_Cyrl & 50.8 & 44.5 & 55.9 & 53.3 & 43.5 & 51.4 & 51.4 & 49.7 & \underline{64.1} & 55.3 & \bf 66.4 \\
    spa\_Latn & 55.0 & 50.5 & 60.1 & 56.3 & 48.2 & 55.3 & 54.3 & 53.0 & \underline{66.9} & 60.7 & \bf 67.5 \\
    tur\_Latn & 46.0 & 38.1 & 49.9 & 48.3 & 41.6 & 38.8 & 49.1 & 48.2 & \underline{57.7} & 46.9 & \bf 62.5 \\
    ukr\_Cyrl & 47.1 & 40.5 & 53.7 & 50.5 & 41.8 & 43.1 & 49.9 & 48.0 & \underline{60.7} & 50.0 & \bf 64.7 \\
    vie\_Latn & 49.9 & 39.8 & 49.7 & 51.7 & 43.7 & 46.8 & 49.8 & 48.9 & \underline{61.1} & 49.5 & \bf 63.9 \\
    zho\_Hans & 56.0 & 49.4 & 54.8 & 54.1 & 44.1 & 50.2 & 50.8 & 50.8 & \bf 66.0 & 58.6 & \underline{65.1} \\
    zsm\_Latn & 48.1 & 41.4 & 53.1 & 52.7 & 42.6 & 43.3 & 49.8 & 49.2 & \underline{60.6} & 50.6 & \bf 64.6 \\
    Avg & 49.6 & 43.6 & 54.8 & 52.2 & 43.2 & 46.7 & 50.8 & 50.0 & \underline{61.6} & 52.6 & \bf 64.2 \\
    \bottomrule
\end{tabular}}
\caption{Per-language model performance comparison on GlobalMMLU.}
\label{tab:global_mmlu_results}
\end{table}
\begin{table}
\setlength{\tabcolsep}{3pt}
\footnotesize
\setlength{\belowcaptionskip}{-0.35cm}
\centering
\resizebox{\linewidth}{!}{
\begin{tabular}{l|ccccccccccc}
\toprule
    \multirow{2}{*}[0ex]{\bf Language} & \bf Qwen3 & \bf Trinity & \bf Granite4 & \bf Marco & \bf Llama3.2 & \bf SmolLM3 & \bf Gemma3 & \bf Tiny-Aya & \bf Qwen3 & \bf Trinity & \bf Marco \\
    & \bf 1.7B Base & \bf Nano Base & \bf Tiny Base & \bf Nano Base & \bf 3B Base & \bf 3B Base & \bf 4B Base & \bf 3.35B Base & \bf 4B Base & \bf Mini Base & \bf Mini Base \\
    \midrule
    arb\_Arab & 42.9 & 32.7 & 47.9 & 48.6 & 39.7 & 47.1 & 43.7 & 42.8 & \underline{53.7} & 41.1 & \bf 60.3 \\
    ben\_Beng & 37.3 & 30.0 & 44.0 & 44.6 & 34.6 & 32.2 & 39.9 & 37.9 & \underline{46.3} & 36.8 & \bf 55.1 \\
    deu\_Latn & 50.5 & 43.2 & 54.0 & 54.0 & 46.8 & 50.4 & 49.7 & 46.1 & \underline{62.0} & 53.1 & \bf 63.9 \\
    fra\_Latn & 51.1 & 46.2 & 56.0 & 54.6 & 47.7 & 51.8 & 50.4 & 45.8 & \underline{62.8} & 57.6 & \bf 63.3 \\
    ind\_Latn & 49.2 & 41.4 & 52.0 & 53.9 & 45.1 & 43.9 & 49.1 & 45.2 & \underline{60.3} & 50.8 & \bf 62.4 \\
    ita\_Latn & 51.5 & 44.3 & 55.5 & 55.9 & 46.4 & 51.2 & 50.5 & 46.0 & \underline{63.2} & 54.7 & \bf 64.1 \\
    jpn\_Jpan & 47.4 & 39.2 & 50.4 & 51.2 & 41.6 & 46.3 & 42.7 & 44.0 & \underline{58.3} & 49.0 & \bf 61.2 \\
    kor\_Hang & 45.8 & 38.1 & 49.9 & 49.9 & 41.2 & 44.9 & 45.7 & 43.6 & \underline{55.3} & 46.3 & \bf 60.0 \\
    por\_Latn & 52.5 & 46.2 & 56.2 & 56.0 & 47.7 & 51.9 & 51.1 & 46.5 & \underline{61.9} & 56.9 & \bf 63.8 \\
    spa\_Latn & 52.1 & 47.6 & 56.2 & 55.9 & 48.3 & 52.8 & 52.0 & 46.1 & \underline{63.9} & 57.3 & \bf 65.1 \\
    zho\_Hans & 54.3 & 44.0 & 53.2 & 54.2 & 44.7 & 47.9 & 47.2 & 45.7 & \bf 64.0 & 56.1 & \underline{62.7} \\
    Avg & 48.6 & 41.2 & 52.3 & 52.6 & 44.0 & 47.3 & 47.5 & 44.5 & \underline{59.2} & 50.9 & \bf 62.0 \\
    \bottomrule
\end{tabular}}
\caption{Per-language model performance comparison on MMMLU.}
\label{tab:mmmlu_results}
\end{table}
\begin{table}
\setlength{\tabcolsep}{3pt}
\footnotesize
\setlength{\belowcaptionskip}{-0.35cm}
\centering
\resizebox{\linewidth}{!}{
\begin{tabular}{l|ccccccccccc}
\toprule
    \multirow{2}{*}[0ex]{\bf Language} & \bf Qwen3 & \bf Trinity & \bf Granite4 & \bf Marco & \bf Llama3.2 & \bf SmolLM3 & \bf Gemma3 & \bf Tiny-Aya & \bf Qwen3 & \bf Trinity & \bf Marco \\
    & \bf 1.7B Base & \bf Nano Base & \bf Tiny Base & \bf Nano Base & \bf 3B Base & \bf 3B Base & \bf 4B Base & \bf 3.35B Base & \bf 4B Base & \bf Mini Base & \bf Mini Base \\
    \midrule
    arb\_Arab & 69.1 & 52.2 & 63.4 & 73.2 & 60.6 & 62.9 & 66.3 & 63.8 & \underline{81.8} & 69.2 & \bf 82.9 \\
    azj\_Latn & 55.7 & 37.3 & 47.9 & 65.3 & 45.1 & 35.6 & 55.4 & 44.2 & \underline{69.6} & 48.3 & \bf 73.7 \\
    ben\_Beng & 52.7 & 38.0 & 54.4 & 63.8 & 46.8 & 33.2 & 58.0 & 60.7 & \underline{70.1} & 53.0 & \bf 71.7 \\
    ces\_Latn & 68.9 & 55.2 & 62.6 & 78.7 & 61.0 & 52.8 & 66.6 & 67.7 & \bf 83.2 & 69.4 & \underline{82.4} \\
    deu\_Latn & 73.1 & 64.8 & 66.1 & 80.3 & 64.6 & 64.3 & 69.1 & 71.0 & \bf 87.0 & 76.0 & \underline{83.9} \\
    ell\_Grek & 66.6 & 43.2 & 61.3 & 75.0 & 62.0 & 63.2 & 68.6 & 69.7 & \underline{80.2} & 68.1 & \bf 82.2 \\
    eng\_Latn & 81.4 & 76.9 & 75.1 & 81.9 & 74.8 & 71.0 & 74.8 & 82.2 & \bf 89.2 & 84.3 & \underline{87.1} \\
    fra\_Latn & 76.3 & 66.9 & 68.8 & 79.3 & 68.2 & 63.8 & 70.6 & 72.1 & \bf 87.2 & 77.3 & \underline{84.2} \\
    heb\_Hebr & 62.0 & 44.7 & 57.7 & 71.2 & 49.2 & 44.4 & 64.3 & 65.3 & \underline{74.3} & 62.6 & \bf 78.2 \\
    hun\_Latn & 63.3 & 48.1 & 56.1 & 71.6 & 59.2 & 42.4 & 64.6 & 61.7 & \bf 79.9 & 61.2 & \underline{78.2} \\
    ind\_Latn & 71.9 & 59.7 & 60.6 & 75.1 & 62.2 & 57.2 & 67.0 & 68.2 & \bf 83.1 & 70.1 & \underline{79.2} \\
    ita\_Latn & 73.7 & 63.6 & 66.8 & 78.0 & 67.2 & 62.7 & 68.0 & 65.7 & \bf 85.3 & 73.7 & \underline{80.6} \\
    jpn\_Jpan & 69.6 & 55.7 & 59.7 & 69.8 & 58.1 & 55.4 & 63.2 & 65.6 & \bf 79.0 & 73.3 & \underline{76.1} \\
    kaz\_Cyrl & 50.9 & 34.2 & 45.0 & 61.6 & 42.6 & 32.1 & 56.1 & 33.7 & \underline{67.0} & 39.9 & \bf 71.7 \\
    kor\_Hang & 71.4 & 56.9 & 65.4 & 74.4 & 63.9 & 61.4 & 68.4 & 69.1 & \bf 83.6 & 74.9 & \underline{80.1} \\
    nld\_Latn & 69.3 & 60.3 & 64.1 & 77.0 & 64.1 & 54.6 & 68.6 & 69.7 & \bf 84.0 & 71.6 & \underline{81.7} \\
    npi\_Deva & 48.1 & 37.4 & 51.6 & 60.2 & 43.6 & 36.4 & 56.7 & 58.9 & \underline{67.1} & 50.4 & \bf 70.6 \\
    pol\_Latn & 68.4 & 53.1 & 56.1 & 74.4 & 59.1 & 51.9 & 64.9 & 63.0 & \bf 83.3 & 67.1 & \underline{80.6} \\
    por\_Latn & 75.1 & 68.0 & 69.4 & 77.9 & 68.4 & 66.6 & 71.2 & 70.4 & \underline{84.0} & 75.2 & \bf 84.1 \\
    ron\_Latn & 69.1 & 60.0 & 63.3 & 79.2 & 60.6 & 55.1 & 68.7 & 67.0 & \bf 84.0 & 74.1 & \underline{83.4} \\
    rus\_Cyrl & 73.9 & 65.4 & 68.9 & 79.7 & 67.2 & 65.3 & 70.4 & 71.0 & \bf 85.7 & 75.7 & \underline{82.8} \\
    spa\_Latn & 74.6 & 67.8 & 68.4 & 77.2 & 68.8 & 66.3 & 68.6 & 68.6 & \bf 85.2 & 75.8 & \underline{82.9} \\
    tha\_Thai & 66.0 & 44.3 & 50.6 & 70.1 & 57.9 & 58.4 & 62.0 & 65.3 & \bf 78.2 & 60.8 & \underline{76.1} \\
    tur\_Latn & 64.7 & 50.1 & 56.4 & 72.0 & 60.6 & 44.9 & 65.7 & 64.7 & \bf 77.3 & 62.1 & \underline{76.8} \\
    ukr\_Cyrl & 67.9 & 52.0 & 62.7 & 75.2 & 61.3 & 55.4 & 69.3 & 67.4 & \bf 83.9 & 68.3 & \underline{82.4} \\
    urd\_Arab & 53.7 & 39.4 & 55.1 & 65.6 & 46.3 & 39.3 & 56.1 & 56.1 & \underline{72.3} & 56.1 & \bf 74.0 \\
    vie\_Latn & 71.6 & 55.0 & 64.7 & 75.9 & 65.2 & 57.8 & 66.0 & 68.8 & \bf 83.8 & 71.6 & \underline{80.9} \\
    zho\_Hans & 79.9 & 71.1 & 72.7 & 79.2 & 70.9 & 63.7 & 68.4 & 76.6 & \bf 86.6 & 80.3 & \underline{82.6} \\
    zsm\_Latn & 70.1 & 58.7 & 61.0 & 76.8 & 63.7 & 55.9 & 68.0 & 67.4 & \underline{81.8} & 70.1 & \bf 82.2 \\
    Avg & 67.6 & 54.5 & 61.2 & 73.8 & 60.1 & 54.3 & 65.7 & 65.4 & \bf 80.6 & 67.6 & \underline{79.8} \\
    \bottomrule
\end{tabular}}
\caption{Per-language model performance comparison on BELEBELE.}
\label{tab:belebele_results}
\end{table}
\begin{table}
\setlength{\tabcolsep}{3pt}
\footnotesize
\setlength{\belowcaptionskip}{-0.35cm}
\centering
\resizebox{\linewidth}{!}{
\begin{tabular}{l|ccccccccccc}
\toprule
    \multirow{2}{*}[0ex]{\bf Language} & \bf Qwen3 & \bf Trinity & \bf Granite4 & \bf Marco & \bf Llama3.2 & \bf SmolLM3 & \bf Gemma3 & \bf Tiny-Aya & \bf Qwen3 & \bf Trinity & \bf Marco \\
    & \bf 1.7B Base & \bf Nano Base & \bf Tiny Base & \bf Nano Base & \bf 3B Base & \bf 3B Base & \bf 4B Base & \bf 3.35B Base & \bf 4B Base & \bf Mini Base & \bf Mini Base \\
    \midrule
    ben\_Beng & 40.0 & 18.0 & 37.2 & 51.2 & 10.4 & 12.0 & 31.2 & 28.8 & \bf 66.0 & 36.8 & \underline{64.8} \\
    deu\_Latn & 58.0 & 45.6 & 60.0 & 66.4 & 24.0 & 60.0 & 35.6 & 38.8 & \underline{76.0} & 63.6 & \bf 79.2 \\
    eng\_Latn & 74.0 & 58.4 & 70.4 & 78.4 & 34.0 & 74.0 & 46.0 & 52.4 & \bf 85.2 & 60.4 & \underline{83.2} \\
    fra\_Latn & 62.0 & 52.0 & 63.2 & 69.6 & 27.2 & 61.2 & 42.0 & 48.8 & \underline{75.6} & 63.2 & \bf 77.6 \\
    jpn\_Jpan & 46.4 & 23.6 & 46.0 & 49.6 & 14.0 & 34.8 & 25.6 & 23.6 & \underline{64.0} & 46.4 & \bf 68.8 \\
    rus\_Cyrl & 66.0 & 46.8 & 64.0 & 74.0 & 22.8 & 59.2 & 36.8 & 44.0 & \bf 84.0 & 63.6 & \underline{83.2} \\
    spa\_Latn & 69.6 & 57.6 & 66.0 & 71.2 & 27.6 & 64.4 & 42.4 & 44.4 & \bf 83.2 & 67.6 & \underline{76.4} \\
    tha\_Thai & 52.8 & 20.0 & 42.8 & 62.0 & 20.4 & 49.2 & 36.0 & 30.0 & \bf 74.0 & 45.2 & \underline{70.4} \\
    zho\_Hans & 60.4 & 43.2 & 60.8 & 65.6 & 21.6 & 42.8 & 34.0 & 35.2 & \underline{75.6} & 67.6 & \bf 76.4 \\
    Avg & 58.8 & 40.6 & 56.7 & 65.3 & 22.4 & 50.8 & 36.6 & 38.4 & \bf 76.0 & 57.2 & \underline{75.6} \\
    \bottomrule
\end{tabular}}
\caption{Per-language model performance comparison on MGSM.}
\label{tab:mgsm_results}
\end{table}
\begin{table}
\setlength{\tabcolsep}{3pt}
\footnotesize
\setlength{\belowcaptionskip}{-0.35cm}
\centering
\resizebox{\linewidth}{!}{
\begin{tabular}{l|ccccccccccc}
\toprule
    \multirow{2}{*}[0ex]{\bf Language} & \bf Qwen3 & \bf Trinity & \bf Granite4 & \bf Marco & \bf Llama3.2 & \bf SmolLM3 & \bf Gemma3 & \bf Tiny-Aya & \bf Qwen3 & \bf Trinity & \bf Marco \\
    & \bf 1.7B Base & \bf Nano Base & \bf Tiny Base & \bf Nano Base & \bf 3B Base & \bf 3B Base & \bf 4B Base & \bf 3.35B Base & \bf 4B Base & \bf Mini Base & \bf Mini Base \\
    \midrule
    arb\_Arab & 39.7 & 34.8 & \underline{51.1} & 44.1 & 41.6 & 50.7 & 50.3 & 49.7 & 46.7 & 43.1 & \bf 55.2 \\
    ben\_Beng & 30.4 & 29.7 & \underline{36.8} & 33.4 & 32.3 & 29.0 & 36.7 & 36.4 & 34.7 & 32.6 & \bf 38.2 \\
    deu\_Latn & 46.1 & 45.3 & \underline{59.7} & 51.5 & 53.3 & \underline{59.7} & 59.0 & 57.4 & 56.4 & 56.5 & \bf 63.5 \\
    fra\_Latn & 51.3 & 53.8 & \underline{64.5} & 56.2 & 57.5 & 63.9 & 63.8 & 62.3 & 61.1 & 64.3 & \bf 67.3 \\
    hun\_Latn & 35.8 & 33.1 & 42.1 & 41.5 & 42.9 & 32.2 & \underline{49.4} & 47.4 & 44.0 & 39.6 & \bf 49.9 \\
    ind\_Latn & 45.9 & 43.1 & 52.1 & 51.9 & 51.7 & 45.2 & \underline{58.3} & 57.8 & 53.9 & 53.3 & \bf 61.9 \\
    ita\_Latn & 48.9 & 47.6 & 60.9 & 54.6 & 55.0 & \underline{62.3} & 61.4 & 59.8 & 58.6 & 58.5 & \bf 65.3 \\
    nld\_Latn & 44.1 & 43.0 & 59.0 & 51.7 & 53.8 & 45.0 & \underline{60.8} & 56.5 & 54.0 & 54.1 & \bf 64.3 \\
    npi\_Deva & 30.2 & 29.4 & \underline{34.8} & 31.9 & 32.1 & 31.2 & 34.7 & 34.5 & 32.6 & 31.9 & \bf 35.5 \\
    por\_Latn & 50.8 & 51.4 & 62.9 & 55.7 & 56.8 & 63.3 & \underline{63.5} & 61.7 & 60.0 & 61.5 & \bf 67.1 \\
    ron\_Latn & 41.2 & 39.9 & 51.3 & 50.3 & 48.2 & 40.4 & \underline{58.2} & 55.7 & 51.3 & 49.7 & \bf 60.3 \\
    rus\_Cyrl & 46.8 & 45.7 & 54.6 & 50.5 & 51.0 & 56.4 & \underline{57.0} & 54.4 & 54.4 & 56.0 & \bf 62.0 \\
    spa\_Latn & 52.4 & 52.9 & 65.4 & 56.7 & 59.3 & \underline{65.8} & 64.9 & 62.4 & 62.1 & 64.3 & \bf 68.3 \\
    ukr\_Cyrl & 40.7 & 38.4 & 51.6 & 46.8 & 46.3 & 42.5 & \underline{53.8} & 51.6 & 49.4 & 48.8 & \bf 58.1 \\
    vie\_Latn & 45.5 & 39.9 & 48.0 & 48.0 & 49.4 & 49.0 & \underline{54.7} & 53.6 & 52.2 & 49.3 & \bf 57.5 \\
    zho\_Hans & 53.0 & 52.6 & 56.7 & 52.6 & 51.9 & 56.3 & 57.1 & 54.4 & 59.8 & \underline{60.7} & \bf 62.4 \\
    Avg & 43.9 & 42.5 & 53.2 & 48.6 & 48.9 & 49.6 & \underline{55.2} & 53.5 & 52.0 & 51.5 & \bf 58.5 \\
    \bottomrule
\end{tabular}}
\caption{Per-language model performance comparison on mHellaSwag.}
\label{tab:mhellaswag_results}
\end{table}
\begin{table}
\setlength{\tabcolsep}{3pt}
\footnotesize
\setlength{\belowcaptionskip}{-0.35cm}
\centering
\resizebox{\linewidth}{!}{
\begin{tabular}{l|ccccccccccc}
\toprule
    \multirow{2}{*}[0ex]{\bf Language} & \bf Qwen3 & \bf Trinity & \bf Granite4 & \bf Marco & \bf Llama3.2 & \bf SmolLM3 & \bf Gemma3 & \bf Tiny-Aya & \bf Qwen3 & \bf Trinity & \bf Marco \\
    & \bf 1.7B Base & \bf Nano Base & \bf Tiny Base & \bf Nano Base & \bf 3B Base & \bf 3B Base & \bf 4B Base & \bf 3.35B Base & \bf 4B Base & \bf Mini Base & \bf Mini Base \\
    \midrule
    arb\_Arab & 32.5 & 25.4 & 37.6 & 34.6 & 29.6 & 36.4 & \underline{39.0} & 35.6 & 38.9 & 30.8 & \bf 43.4 \\
    ben\_Beng & 27.2 & 22.7 & \underline{31.1} & 28.7 & 27.1 & 24.3 & 29.7 & 29.1 & \underline{31.1} & 25.1 & \bf 33.8 \\
    deu\_Latn & 36.2 & 32.3 & 44.4 & 38.1 & 35.5 & 42.3 & \underline{45.5} & 38.7 & 44.9 & 41.5 & \bf 48.8 \\
    fra\_Latn & 37.4 & 37.4 & 44.5 & 41.2 & 38.3 & 44.1 & 45.7 & 41.4 & \underline{45.9} & 44.1 & \bf 51.2 \\
    hun\_Latn & 29.7 & 25.7 & 33.8 & 32.9 & 32.4 & 25.5 & \underline{39.0} & 35.5 & 38.7 & 31.5 & \bf 41.2 \\
    ind\_Latn & 37.8 & 30.9 & 41.4 & 38.3 & 35.5 & 33.8 & \underline{46.9} & 41.1 & 45.1 & 39.0 & \bf 47.8 \\
    ita\_Latn & 38.8 & 35.9 & \underline{45.3} & 41.4 & 36.8 & \underline{45.3} & 44.1 & 40.5 & \underline{45.3} & 42.7 & \bf 52.6 \\
    nld\_Latn & 33.4 & 31.2 & 41.9 & 39.3 & 33.8 & 31.7 & \underline{43.5} & 35.8 & 41.2 & 37.8 & \bf 48.4 \\
    npi\_Deva & 23.9 & 23.6 & 26.4 & 25.7 & 23.8 & 24.6 & 26.9 & 26.3 & \underline{28.1} & 25.5 & \bf 30.0 \\
    por\_Latn & 40.8 & 37.0 & 47.4 & 41.0 & 38.8 & 45.8 & \underline{47.6} & 42.6 & 45.9 & 45.8 & \bf 50.8 \\
    ron\_Latn & 35.0 & 29.6 & 37.2 & 39.3 & 35.1 & 30.8 & \underline{41.4} & 38.8 & 41.0 & 38.4 & \bf 45.2 \\
    rus\_Cyrl & 36.9 & 31.8 & 41.6 & 39.6 & 35.6 & 40.4 & 40.4 & 36.9 & \underline{43.6} & 40.7 & \bf 48.7 \\
    spa\_Latn & 37.3 & 36.2 & 46.8 & 39.4 & 39.0 & 46.8 & \underline{49.3} & 42.6 & 46.0 & 43.4 & \bf 51.3 \\
    ukr\_Cyrl & 33.8 & 30.4 & 38.9 & 36.9 & 35.3 & 30.3 & \underline{40.8} & 34.6 & 39.5 & 36.0 & \bf 44.5 \\
    vie\_Latn & 34.9 & 27.7 & 36.7 & 34.4 & 33.0 & 35.2 & 40.6 & 36.2 & \underline{42.2} & 33.5 & \bf 42.4 \\
    zho\_Hans & 39.6 & 36.6 & 43.7 & 40.1 & 37.4 & 40.7 & 44.4 & 39.0 & \bf 48.2 & 44.8 & \underline{46.6} \\
    Avg & 34.7 & 30.9 & 39.9 & 36.9 & 34.2 & 36.1 & 41.5 & 37.2 & \underline{41.6} & 37.5 & \bf 45.4 \\
    \bottomrule
\end{tabular}}
\caption{Per-language model performance comparison on mARC-Challenge.}
\label{tab:marc_challenge_results}
\end{table}
\begin{table}
\setlength{\tabcolsep}{3pt}
\footnotesize
\setlength{\belowcaptionskip}{-0.35cm}
\centering
\resizebox{\linewidth}{!}{
\begin{tabular}{l|ccccccccccc}
\toprule
    \multirow{2}{*}[0ex]{\bf Language} & \bf Qwen3 & \bf Trinity & \bf Granite4 & \bf Marco & \bf Llama3.2 & \bf SmolLM3 & \bf Gemma3 & \bf Tiny-Aya & \bf Qwen3 & \bf Trinity & \bf Marco \\
    & \bf 1.7B Base & \bf Nano Base & \bf Tiny Base & \bf Nano Base & \bf 3B Base & \bf 3B Base & \bf 4B Base & \bf 3.35B Base & \bf 4B Base & \bf Mini Base & \bf Mini Base \\
    \midrule
    arb\_Arab & 24.1 & 16.3 & 26.2 & 24.3 & 23.3 & 29.4 & 21.3 & 22.8 & \underline{34.2} & 27.7 & \bf 39.5 \\
    ben\_Beng & 21.6 & 13.3 & 25.3 & 23.6 & 19.9 & 19.2 & 20.7 & 23.8 & \underline{31.8} & 15.1 & \bf 34.2 \\
    ces\_Latn & 28.2 & 21.4 & 31.8 & 29.6 & 20.9 & 25.7 & 23.3 & 23.1 & \underline{39.8} & 32.0 & \bf 40.8 \\
    deu\_Latn & 26.9 & 22.1 & 30.8 & 30.6 & 22.3 & 30.3 & 24.7 & 24.3 & \underline{38.8} & 35.7 & \bf 40.5 \\
    eng\_Latn & 33.7 & 32.3 & 35.9 & 34.7 & 27.2 & 36.7 & 28.4 & 27.0 & \underline{44.6} & 42.3 & \bf 45.4 \\
    fra\_Latn & 29.4 & 23.6 & 31.8 & 31.6 & 23.3 & 32.5 & 25.3 & 24.5 & \underline{40.3} & 38.8 & \bf 41.5 \\
    hun\_Latn & 26.4 & 16.2 & 26.7 & 27.0 & 22.3 & 24.3 & 24.0 & 22.1 & \bf 36.6 & 27.2 & \underline{35.9} \\
    ind\_Latn & 28.6 & 21.9 & 29.4 & 27.9 & 22.4 & 26.4 & 26.0 & 25.5 & \underline{40.8} & 31.8 & \bf 41.2 \\
    ita\_Latn & 29.9 & 22.3 & 32.8 & 32.0 & 23.8 & 31.0 & 27.4 & 24.0 & \underline{40.8} & 36.1 & \bf 41.2 \\
    jpn\_Jpan & 26.5 & 19.4 & 30.1 & 27.7 & 20.2 & 28.9 & 23.8 & 24.3 & \bf 37.6 & 33.3 & \underline{37.1} \\
    kor\_Hang & 26.7 & 18.5 & 29.8 & 27.6 & 21.6 & 27.0 & 22.3 & 23.3 & \bf 37.6 & 30.8 & \underline{35.7} \\
    por\_Latn & 27.7 & 22.4 & 30.3 & 30.6 & 22.6 & 31.3 & 25.7 & 27.0 & \bf 40.6 & 39.5 & \underline{40.3} \\
    rus\_Cyrl & 26.7 & 21.3 & 31.1 & 32.1 & 24.1 & 28.9 & 25.5 & 25.9 & \bf 40.8 & 35.9 & \underline{40.0} \\
    spa\_Latn & 29.3 & 24.7 & 32.8 & 31.1 & 23.5 & 31.6 & 25.0 & 24.7 & \underline{41.0} & 38.9 & \bf 42.0 \\
    tha\_Thai & 25.9 & 14.1 & 26.9 & 24.7 & 21.3 & 27.2 & 21.9 & 22.1 & \underline{33.3} & 21.8 & \bf 37.1 \\
    ukr\_Cyrl & 26.5 & 16.5 & 31.8 & 28.9 & 20.9 & 25.5 & 24.5 & 24.1 & \underline{38.3} & 30.3 & \bf 38.9 \\
    urd\_Arab & 21.9 & 17.5 & 28.6 & 25.0 & 20.9 & 24.7 & 22.8 & 22.6 & \underline{34.4} & 25.9 & \bf 35.9 \\
    vie\_Latn & 29.3 & 20.1 & 28.2 & 29.6 & 23.1 & 28.2 & 24.7 & 26.5 & \bf 40.3 & 32.8 & \bf 40.3 \\
    zho\_Hans & 27.4 & 21.4 & 31.3 & 30.4 & 21.8 & 27.9 & 24.7 & 24.1 & \bf 39.3 & 35.7 & \underline{38.1} \\
    Avg & 27.2 & 20.3 & 30.1 & 28.9 & 22.4 & 28.2 & 24.3 & 24.3 & \underline{38.5} & 32.2 & \bf 39.2 \\
    \bottomrule
\end{tabular}}
\caption{Per-language model performance comparison on MMLU-ProX-Lite.}
\label{tab:mmlu_prox_results}
\end{table}
\begin{table}
\setlength{\tabcolsep}{3pt}
\footnotesize
\setlength{\belowcaptionskip}{-0.35cm}
\centering
\resizebox{\linewidth}{!}{
\begin{tabular}{l|ccccccccccc}
\toprule
    \multirow{2}{*}[0ex]{\bf Language} & \bf Qwen3 & \bf Trinity & \bf Granite4 & \bf Marco & \bf Llama3.2 & \bf SmolLM3 & \bf Gemma3 & \bf Tiny-Aya & \bf Qwen3 & \bf Trinity & \bf Marco \\
    & \bf 1.7B Base & \bf Nano Base & \bf Tiny Base & \bf Nano Base & \bf 3B Base & \bf 3B Base & \bf 4B Base & \bf 3.35B Base & \bf 4B Base & \bf Mini Base & \bf Mini Base \\
    \midrule
    arb\_Arab & 49.1 & 38.6 & 51.1 & 50.4 & 43.5 & 52.0 & 57.2 & 56.9 & \underline{61.8} & 49.3 & \bf 62.1 \\
    azj\_Latn & 47.8 & 38.3 & 45.1 & 47.1 & 38.5 & 33.0 & 46.2 & 41.6 & \underline{51.8} & 43.6 & \bf 56.6 \\
    ben\_Beng & 40.5 & 34.9 & 49.5 & 44.0 & 34.3 & 32.1 & 48.4 & 45.6 & \bf 53.3 & 42.0 & \underline{50.4} \\
    deu\_Latn & 42.4 & 43.2 & 47.5 & 46.0 & 36.0 & 39.6 & 41.0 & 41.7 & \bf 56.1 & 48.9 & \underline{51.8} \\
    ell\_Grek & 44.9 & 38.0 & 45.5 & 48.2 & 38.2 & 44.4 & 45.1 & 49.3 & \underline{55.6} & 40.8 & \bf 60.0 \\
    fra\_Latn & 54.4 & 48.7 & 63.0 & 55.1 & 54.9 & 61.6 & 55.8 & 58.7 & \underline{65.9} & 63.0 & \bf 68.0 \\
    heb\_Hebr & 56.9 & 53.3 & 57.8 & 63.3 & 48.4 & 47.8 & 59.1 & \underline{66.0} & 65.3 & 61.8 & \bf 69.1 \\
    hun\_Latn & 41.6 & 32.2 & 43.5 & 46.4 & 38.5 & 34.5 & 42.5 & 49.8 & \bf 52.7 & 38.5 & \underline{50.2} \\
    ind\_Latn & 59.5 & 53.8 & 52.9 & 57.5 & 53.5 & 51.5 & 58.2 & 61.3 & \underline{65.5} & 56.5 & \bf 68.5 \\
    ita\_Latn & 65.1 & 56.2 & 68.1 & 63.7 & 58.9 & 63.3 & 64.2 & 64.4 & \bf 73.0 & 65.7 & \underline{72.1} \\
    jpn\_Jpan & 63.1 & 52.5 & 65.3 & 61.5 & 47.1 & 55.3 & 58.7 & 61.7 & \bf 78.2 & 67.9 & \underline{77.4} \\
    kaz\_Cyrl & 40.0 & 33.8 & 41.0 & \underline{48.6} & 39.2 & 35.0 & 41.4 & 33.2 & 45.0 & 35.6 & \bf 54.6 \\
    kor\_Hang & 58.0 & 43.0 & 56.4 & 56.4 & 46.6 & 52.0 & 55.2 & 60.8 & \underline{63.8} & 54.6 & \bf 64.2 \\
    nld\_Latn & 52.8 & 48.6 & 61.0 & 58.1 & 53.7 & 47.0 & 56.4 & 58.6 & \underline{66.4} & 59.2 & \bf 71.0 \\
    npi\_Deva & 37.0 & 32.6 & 47.4 & 49.4 & 37.6 & 36.8 & 48.6 & \underline{51.6} & 49.0 & 42.4 & \bf 54.2 \\
    pol\_Latn & 46.7 & 40.1 & 48.7 & 47.6 & 41.8 & 41.2 & 50.7 & 51.1 & \bf 60.4 & 51.3 & \underline{58.9} \\
    por\_Latn & 54.1 & 49.9 & 53.9 & 52.3 & 48.1 & 53.2 & 47.4 & 51.5 & \bf 66.2 & 56.1 & \underline{61.9} \\
    rus\_Cyrl & 57.6 & 47.1 & 51.1 & 53.8 & 48.4 & 51.4 & 52.4 & 53.1 & \bf 63.0 & 53.8 & \underline{60.7} \\
    spa\_Latn & 59.8 & 53.3 & 64.5 & 60.5 & 55.8 & 58.9 & 59.8 & 61.1 & \underline{69.3} & 62.4 & \bf 70.9 \\
    tur\_Latn & 46.4 & 40.9 & 51.3 & 53.6 & 47.4 & 40.0 & 57.5 & 53.3 & \underline{58.6} & 45.6 & \bf 58.8 \\
    ukr\_Cyrl & 57.1 & 48.7 & 54.9 & 58.9 & 51.3 & 51.5 & 62.5 & \underline{63.6} & \bf 66.5 & 54.2 & \underline{63.6} \\
    urd\_Arab & 36.6 & 38.9 & 37.5 & 37.2 & 36.6 & 32.7 & \bf 44.6 & 39.8 & \bf 44.6 & 43.8 & 42.3 \\
    vie\_Latn & 52.9 & 38.5 & 42.7 & 57.8 & 44.7 & 45.3 & 54.4 & 55.5 & \underline{64.4} & 52.9 & \bf 65.6 \\
    zho\_Hans & \underline{65.5} & 49.9 & 51.6 & 56.0 & 49.4 & 54.7 & 52.8 & 59.6 & \bf 76.1 & 56.7 & 64.0 \\
    zsm\_Latn & 49.5 & 42.5 & 50.5 & 55.5 & 45.3 & 40.3 & 54.7 & 56.5 & \underline{62.3} & 50.7 & \bf 65.3 \\
    Avg & 51.2 & 43.9 & 52.1 & 53.2 & 45.5 & 46.2 & 52.6 & 53.9 & \underline{61.4} & 51.9 & \bf 61.7 \\
    \bottomrule
\end{tabular}}
\caption{Per-language model performance comparison on INCLUDE.}
\label{tab:include_results}
\end{table}
\begin{table}
\setlength{\tabcolsep}{3pt}
\footnotesize
\setlength{\belowcaptionskip}{-0.35cm}
\centering
\resizebox{\linewidth}{!}{
\begin{tabular}{l|ccccccccccc}
\toprule
    \multirow{2}{*}[0ex]{\bf Language} & \bf Qwen3 & \bf Trinity & \bf Granite4 & \bf Marco & \bf Llama3.2 & \bf SmolLM3 & \bf Gemma3 & \bf Tiny-Aya & \bf Qwen3 & \bf Trinity & \bf Marco \\
    & \bf 1.7B Base & \bf Nano Base & \bf Tiny Base & \bf Nano Base & \bf 3B Base & \bf 3B Base & \bf 4B Base & \bf 3.35B Base & \bf 4B Base & \bf Mini Base & \bf Mini Base \\
    \midrule
    arb\_Arab & 53.0 & 47.0 & 58.0 & 53.0 & 55.0 & 50.0 & \underline{62.0} & \bf 65.0 & 60.0 & 48.0 & 59.0 \\
    azj\_Latn & 52.0 & 45.0 & \underline{59.0} & 48.0 & 52.0 & 53.0 & 54.0 & 53.0 & 57.0 & 50.0 & \bf 63.0 \\
    ben\_Beng & 45.0 & 36.0 & 59.0 & 55.0 & 49.0 & 38.0 & 60.0 & \bf 63.0 & 44.0 & 41.0 & \bf 63.0 \\
    ces\_Latn & 54.0 & 40.0 & 63.0 & 62.0 & 60.0 & 54.0 & \bf 71.0 & \underline{70.0} & 61.0 & 45.0 & 69.0 \\
    deu\_Latn & 66.0 & 63.0 & 64.0 & 61.0 & \underline{71.0} & 68.0 & \bf 73.0 & 70.0 & 65.0 & 70.0 & 66.0 \\
    ell\_Grek & \bf 64.0 & 58.0 & \bf 64.0 & 58.0 & 62.0 & 57.0 & 61.0 & 60.0 & 63.0 & 58.0 & 58.0 \\
    eng\_Latn & 74.0 & 71.0 & \underline{79.0} & 73.0 & 72.0 & \underline{79.0} & 78.0 & 75.0 & 76.0 & 72.0 & \bf 82.0 \\
    fra\_Latn & 65.5 & 59.5 & \underline{73.5} & 69.0 & 69.5 & \underline{73.5} & 72.5 & \underline{73.5} & 70.0 & 66.0 & \bf 77.0 \\
    heb\_Hebr & 48.0 & 44.0 & \bf 60.0 & 53.0 & 51.0 & 50.0 & 56.0 & 58.0 & 51.0 & 46.0 & \bf 60.0 \\
    hun\_Latn & 50.0 & 41.0 & 52.0 & 58.0 & 48.0 & 48.0 & \underline{61.0} & 54.0 & 53.0 & 42.0 & \bf 64.0 \\
    ind\_Latn & 70.0 & 53.0 & 61.0 & 72.0 & 68.0 & 52.0 & 75.0 & \underline{79.0} & 70.0 & 59.0 & \bf 80.0 \\
    ita\_Latn & 62.0 & 55.0 & 69.0 & 71.0 & 70.0 & \underline{80.0} & 79.0 & \underline{80.0} & 78.0 & 56.0 & \bf 83.0 \\
    jpn\_Jpan & 72.0 & 63.0 & 76.0 & 73.0 & 68.0 & 71.0 & \bf 81.0 & \underline{79.0} & 75.0 & 68.0 & 77.0 \\
    kaz\_Cyrl & 65.0 & 45.0 & 63.0 & 75.0 & 68.0 & 57.0 & \underline{78.0} & 55.0 & 64.0 & 49.0 & \bf 80.0 \\
    kor\_Hang & 53.0 & 47.0 & 56.0 & 56.0 & 52.0 & 50.0 & 51.0 & \underline{58.0} & 57.0 & 53.0 & \bf 64.0 \\
    nld\_Latn & 57.0 & 47.0 & 69.0 & 65.0 & 64.0 & 54.0 & \underline{75.0} & 70.0 & 61.0 & 51.0 & \bf 83.0 \\
    npi\_Deva & 54.0 & 44.0 & 59.0 & 63.0 & 54.0 & 47.0 & 66.0 & \underline{67.0} & 58.0 & 50.0 & \bf 69.0 \\
    pol\_Latn & 53.0 & 45.0 & 59.0 & 65.0 & 56.0 & 51.0 & \underline{71.0} & 70.0 & 64.0 & 50.0 & \bf 78.0 \\
    por\_Latn & 56.5 & 51.0 & 60.5 & 58.0 & 58.5 & 64.0 & \underline{68.0} & 65.0 & \bf 69.5 & 61.0 & \underline{68.0} \\
    ron\_Latn & 63.0 & 63.0 & 69.0 & 69.0 & 63.0 & 65.0 & \underline{71.0} & 70.0 & 68.0 & 68.0 & \bf 76.0 \\
    rus\_Cyrl & 71.0 & 59.0 & 71.0 & 73.0 & 77.0 & \underline{82.0} & 81.0 & 77.0 & 79.0 & 64.0 & \bf 85.0 \\
    spa\_Latn & 73.3 & 61.7 & 73.7 & 76.3 & 76.3 & 77.7 & \underline{80.7} & 80.3 & 78.3 & 70.7 & \bf 82.3 \\
    tha\_Thai & 60.0 & 47.0 & 53.0 & 62.0 & 58.0 & 65.0 & 61.0 & \underline{66.0} & 64.0 & 48.0 & \bf 68.0 \\
    tur\_Latn & 46.0 & 42.0 & 44.0 & 58.0 & 56.0 & 40.0 & \underline{68.0} & 56.0 & 53.0 & 47.0 & \bf 70.0 \\
    ukr\_Cyrl & 70.0 & 54.0 & 76.0 & 74.0 & 70.0 & 67.0 & \underline{83.0} & 76.0 & 78.0 & 61.0 & \bf 87.0 \\
    urd\_Arab & 57.0 & 52.0 & 70.0 & 71.0 & 63.0 & 55.0 & 76.0 & \underline{77.0} & 72.0 & 58.0 & \bf 81.0 \\
    vie\_Latn & 65.0 & 53.0 & 66.0 & 67.0 & 65.0 & 63.0 & \bf 79.0 & 75.0 & 72.0 & 52.0 & \underline{76.0} \\
    zho\_Hans & 56.0 & 54.0 & 56.5 & \underline{57.5} & 53.0 & 55.5 & 54.0 & 52.0 & \bf 59.5 & 56.0 & 56.0 \\
    zsm\_Latn & 49.0 & 49.0 & 55.0 & 53.0 & 51.0 & 57.0 & 57.0 & \underline{63.0} & 47.0 & 60.0 & \bf 67.0 \\
    Avg & 59.5 & 51.4 & 63.4 & 63.8 & 61.4 & 59.4 & \underline{69.1} & 67.5 & 64.4 & 55.9 & \bf 72.1 \\
    \bottomrule
\end{tabular}}
\caption{Per-language model performance comparison on Global-PIQA.}
\label{tab:global_piqa_results}
\end{table}
\begin{table}
\setlength{\tabcolsep}{3pt}
\footnotesize
\setlength{\belowcaptionskip}{-0.35cm}
\centering
\resizebox{\linewidth}{!}{
\begin{tabular}{l|ccccccccccc}
\toprule
    \multirow{2}{*}[0ex]{\bf Language} & \bf Qwen3 & \bf Trinity & \bf Granite4 & \bf Marco & \bf Llama3.2 & \bf SmolLM3 & \bf Gemma3 & \bf Tiny-Aya & \bf Qwen3 & \bf Trinity & \bf Marco \\
    & \bf 1.7B Base & \bf Nano Base & \bf Tiny Base & \bf Nano Base & \bf 3B Base & \bf 3B Base & \bf 4B Base & \bf 3.35B Base & \bf 4B Base & \bf Mini Base & \bf Mini Base \\
    \midrule
    arb\_Arab & 16.6 & 7.2 & 31.5 & 22.6 & 18.8 & 30.5 & \underline{34.0} & \bf 34.8 & 24.4 & 6.6 & 33.2 \\
    azj\_Latn & 1.1 & 0.3 & 1.9 & 10.7 & 6.0 & 0.3 & \underline{11.0} & 1.0 & 3.5 & 1.3 & \bf 16.3 \\
    ben\_Beng & 4.6 & 2.3 & 21.3 & 14.7 & 12.6 & 0.6 & \bf 26.0 & 21.3 & 13.3 & 2.9 & \underline{23.7} \\
    ces\_Latn & 16.4 & 9.2 & \underline{34.4} & 27.8 & 27.6 & 8.7 & 34.3 & \bf 35.8 & 25.7 & 13.3 & 33.3 \\
    deu\_Latn & 30.5 & 32.5 & 39.7 & 32.5 & 34.6 & 39.0 & \underline{41.1} & 39.3 & 37.0 & 25.9 & \bf 41.3 \\
    ell\_Grek & 10.0 & 5.0 & 17.9 & 21.7 & 22.2 & 28.3 & \underline{31.9} & \bf 33.0 & 19.9 & 8.3 & 28.1 \\
    fra\_Latn & 39.8 & 44.0 & 48.3 & 43.0 & 43.5 & 46.9 & \underline{49.3} & 48.4 & 46.1 & 33.4 & \bf 50.3 \\
    heb\_Hebr & 7.9 & 6.5 & 25.9 & 20.5 & 19.6 & 2.1 & \underline{32.9} & \bf 34.6 & 16.3 & 8.7 & 30.4 \\
    hun\_Latn & 9.2 & 3.4 & 13.2 & 18.9 & 20.6 & 1.8 & \bf 28.6 & 25.3 & 19.0 & 5.2 & \underline{25.5} \\
    ind\_Latn & 32.1 & 24.1 & 33.7 & 37.8 & 35.3 & 23.4 & 45.0 & \bf 46.2 & 39.2 & 23.5 & \underline{45.6} \\
    ita\_Latn & 25.9 & 23.8 & 31.7 & 27.9 & 29.4 & 32.1 & 32.9 & \underline{33.9} & 30.8 & 19.3 & \bf 34.1 \\
    jpn\_Jpan & 17.1 & 14.0 & 23.7 & 19.4 & 18.4 & 20.7 & \underline{26.0} & 22.8 & 23.8 & 8.1 & \bf 27.0 \\
    kaz\_Cyrl & 1.6 & 0.3 & 2.1 & 8.2 & 4.6 & 0.3 & \underline{15.1} & 0.3 & 6.2 & 1.4 & \bf 18.2 \\
    kor\_Hang & 9.4 & 6.5 & 16.1 & 13.2 & 12.6 & 14.8 & 21.7 & \underline{22.8} & 16.7 & 6.3 & \bf 23.4 \\
    nld\_Latn & 20.8 & 15.8 & 29.0 & 27.1 & 27.4 & 17.1 & \underline{31.4} & 29.6 & 26.5 & 16.8 & \bf 32.1 \\
    npi\_Deva & 2.0 & 1.7 & 13.5 & 9.4 & 6.2 & 2.0 & \underline{19.2} & \bf 23.7 & 9.1 & 3.8 & 17.9 \\
    pol\_Latn & 15.9 & 12.8 & 21.2 & 22.2 & 22.8 & 10.0 & \underline{28.7} & 26.6 & 20.9 & 12.3 & \bf 29.3 \\
    por\_Latn & 40.5 & 40.2 & 47.5 & 42.6 & 44.3 & 46.5 & \underline{49.9} & 49.6 & 47.5 & 32.3 & \bf 50.7 \\
    ron\_Latn & 26.9 & 29.9 & 37.5 & 36.3 & 31.9 & 12.3 & \bf 42.7 & 40.6 & 34.5 & 26.7 & \underline{41.0} \\
    rus\_Cyrl & 27.1 & 28.1 & 33.4 & 27.5 & 30.3 & 32.8 & \underline{36.0} & 33.6 & 31.9 & 21.7 & \bf 36.6 \\
    spa\_Latn & 26.4 & 26.1 & 29.4 & 27.2 & 29.3 & 30.2 & \bf 31.3 & \underline{31.2} & 29.9 & 20.3 & 31.1 \\
    tha\_Thai & 22.6 & 9.7 & 19.1 & 29.3 & 20.2 & 27.0 & \underline{35.4} & 27.3 & 30.3 & 12.8 & \bf 36.5 \\
    tur\_Latn & 12.3 & 4.3 & 12.2 & 18.7 & 16.8 & 4.8 & \bf 30.1 & 26.1 & 21.4 & 7.4 & \underline{29.1} \\
    ukr\_Cyrl & 15.0 & 12.1 & 27.9 & 24.8 & 25.8 & 13.7 & \underline{33.4} & 31.0 & 25.0 & 12.3 & \bf 33.6 \\
    urd\_Arab & 1.7 & 1.9 & 12.6 & 11.4 & 8.1 & 0.6 & \underline{16.4} & 15.5 & 7.3 & 3.6 & \bf 19.7 \\
    vie\_Latn & 32.5 & 16.9 & 28.4 & 33.8 & 32.5 & 30.0 & \underline{40.9} & \bf 41.1 & 37.8 & 16.9 & 40.8 \\
    zho\_Hans & 32.0 & 28.3 & 30.8 & 27.6 & 27.1 & 29.7 & 33.1 & 30.9 & \bf 36.7 & 18.2 & \underline{36.1} \\
    zsm\_Latn & 22.9 & 16.7 & 27.1 & 34.4 & 29.0 & 17.5 & \underline{39.9} & \bf 40.6 & 30.5 & 14.6 & 39.8 \\
    Avg & 18.6 & 15.1 & 25.4 & 24.7 & 23.5 & 18.7 & \underline{32.1} & 30.2 & 25.4 & 13.7 & \bf 32.3 \\
    \bottomrule
\end{tabular}}
\caption{Per-language model performance comparison on FLORES En-Xx translation direction.}
\label{tab:flores_en_xx_results}
\end{table}
\begin{table}
\setlength{\tabcolsep}{3pt}
\footnotesize
\setlength{\belowcaptionskip}{-0.35cm}
\centering
\resizebox{\linewidth}{!}{
\begin{tabular}{l|ccccccccccc}
\toprule
    \multirow{2}{*}[0ex]{\bf Language} & \bf Qwen3 & \bf Trinity & \bf Granite4 & \bf Marco & \bf Llama3.2 & \bf SmolLM3 & \bf Gemma3 & \bf Tiny-Aya & \bf Qwen3 & \bf Trinity & \bf Marco \\
    & \bf 1.7B Base & \bf Nano Base & \bf Tiny Base & \bf Nano Base & \bf 3B Base & \bf 3B Base & \bf 4B Base & \bf 3.35B Base & \bf 4B Base & \bf Mini Base & \bf Mini Base \\
    \midrule
    arb\_Arab & 33.3 & 30.3 & 40.3 & 35.1 & 35.7 & 40.6 & \underline{42.4} & 41.6 & 40.1 & 22.7 & \bf 42.9 \\
    azj\_Latn & 15.4 & 10.6 & 20.7 & 20.5 & 21.1 & 8.9 & \underline{25.8} & 17.0 & 22.2 & 8.3 & \bf 27.1 \\
    ben\_Beng & 22.7 & 19.6 & 32.4 & 25.6 & 25.5 & 5.8 & \bf 34.3 & 31.7 & 31.0 & 13.7 & \underline{33.4} \\
    ces\_Latn & 36.8 & 37.0 & 42.0 & 38.9 & 39.8 & 32.8 & \underline{43.3} & 42.3 & 41.2 & 26.6 & \bf 43.6 \\
    deu\_Latn & 42.6 & 45.4 & 46.9 & 43.2 & 45.1 & 46.3 & \underline{48.4} & 46.6 & 45.9 & 36.7 & \bf 49.2 \\
    ell\_Grek & 28.9 & 28.6 & 35.4 & 33.4 & 36.0 & 38.2 & \bf 40.0 & 39.8 & 35.5 & 20.9 & \underline{40.0} \\
    fra\_Latn & 44.4 & 48.5 & \underline{49.5} & 45.6 & 46.3 & 48.3 & 49.2 & 48.0 & 47.8 & 38.8 & \bf 50.5 \\
    heb\_Hebr & 30.8 & 32.4 & 40.5 & 37.4 & 38.3 & 21.4 & \underline{46.0} & 44.5 & 38.8 & 22.7 & \bf 46.2 \\
    hun\_Latn & 29.8 & 28.4 & 31.5 & 29.1 & 34.4 & 21.4 & \bf 38.5 & 37.3 & 36.4 & 21.2 & \underline{38.1} \\
    ind\_Latn & 40.6 & 40.9 & 44.0 & 41.7 & 42.2 & 39.2 & \underline{47.1} & 46.3 & 45.4 & 32.4 & \bf 47.1 \\
    ita\_Latn & 34.7 & 37.3 & 38.9 & 35.4 & 38.3 & 38.6 & \underline{40.0} & 39.1 & 38.1 & 28.8 & \bf 40.8 \\
    jpn\_Jpan & 25.2 & 26.9 & 29.4 & 25.3 & 25.4 & 27.5 & \underline{31.6} & 29.7 & 30.4 & 20.3 & \bf 31.8 \\
    kaz\_Cyrl & 19.6 & 7.3 & 22.2 & 25.5 & 22.9 & 5.6 & \underline{32.3} & 8.0 & 27.8 & 8.0 & \bf 33.8 \\
    kor\_Hang & 25.1 & 25.1 & 30.3 & 25.1 & 26.3 & 27.2 & \bf 32.9 & 31.0 & 31.4 & 20.4 & \underline{32.8} \\
    nld\_Latn & 32.1 & 33.7 & 36.5 & 33.9 & 34.4 & 31.2 & \underline{36.8} & 35.6 & 34.8 & 26.4 & \bf 36.9 \\
    npi\_Deva & 22.4 & 19.6 & 33.9 & 26.8 & 23.2 & 15.9 & \bf 38.1 & \underline{36.7} & 31.4 & 18.8 & 36.7 \\
    pol\_Latn & 30.0 & 30.1 & 33.7 & 30.6 & 31.7 & 27.9 & \underline{34.6} & 34.1 & 31.7 & 23.6 & \bf 35.8 \\
    por\_Latn & 46.1 & 50.4 & 52.5 & 48.6 & 50.1 & 50.6 & \underline{53.0} & 51.8 & 51.9 & 38.4 & \bf 53.7 \\
    ron\_Latn & 40.4 & 42.7 & 46.6 & 42.4 & 44.2 & 39.0 & \underline{47.5} & 45.8 & 43.5 & 32.1 & \bf 48.2 \\
    rus\_Cyrl & 33.9 & 36.2 & 38.4 & 34.7 & 36.3 & 38.0 & \underline{39.5} & 37.3 & 36.8 & 28.0 & \bf 41.5 \\
    spa\_Latn & 32.2 & 33.5 & 34.6 & 33.5 & 35.0 & 35.7 & \bf 37.2 & 36.5 & 35.1 & 25.1 & \underline{36.7} \\
    tha\_Thai & 26.9 & 20.9 & 27.6 & 28.8 & 29.7 & 31.7 & \underline{34.6} & 33.9 & 32.3 & 18.1 & \bf 36.0 \\
    tur\_Latn & 29.2 & 27.9 & 34.6 & 29.8 & 33.7 & 23.3 & \bf 40.2 & 37.8 & 37.1 & 19.6 & \underline{39.6} \\
    ukr\_Cyrl & 35.1 & 36.0 & 40.5 & 38.0 & 39.3 & 35.2 & \underline{43.1} & 41.5 & 40.0 & 28.0 & \bf 44.5 \\
    urd\_Arab & 21.4 & 17.9 & 31.7 & 25.7 & 26.1 & 10.8 & \bf 34.7 & 33.0 & 29.5 & 17.6 & \underline{34.3} \\
    vie\_Latn & 34.8 & 32.0 & 36.7 & 34.5 & 36.2 & 36.3 & \underline{40.0} & 38.9 & 38.3 & 24.0 & \bf 40.1 \\
    zho\_Hans & 30.0 & 30.8 & 32.5 & 28.9 & 30.0 & 30.6 & 32.9 & 31.9 & \underline{34.6} & 22.9 & \bf 34.9 \\
    zsm\_Latn & 39.1 & 39.8 & 44.2 & 41.7 & 42.4 & 39.1 & \underline{47.4} & 46.8 & 44.4 & 31.1 & \bf 48.1 \\
    Avg & 31.6 & 31.1 & 36.7 & 33.6 & 34.6 & 30.3 & \underline{39.7} & 37.3 & 36.9 & 24.1 & \bf 40.2 \\
    \bottomrule
\end{tabular}}
\caption{Per-language model performance comparison on FLORES Xx-En translation direction.}
\label{tab:flores_xx_en_results}
\end{table}
\begin{table}
\setlength{\tabcolsep}{3pt}
\footnotesize
\setlength{\belowcaptionskip}{-0.35cm}
\centering
\resizebox{\linewidth}{!}{
\begin{tabular}{l|ccccccccccc}
\toprule
    \multirow{2}{*}[0ex]{\bf Language} & \bf Qwen3 & \bf Trinity & \bf Granite4 & \bf Marco & \bf Llama3.2 & \bf SmolLM3 & \bf Gemma3 & \bf Tiny-Aya & \bf Qwen3 & \bf Trinity & \bf Marco \\
    & \bf 1.7B Base & \bf Nano Base & \bf Tiny Base & \bf Nano Base & \bf 3B Base & \bf 3B Base & \bf 4B Base & \bf 3.35B Base & \bf 4B Base & \bf Mini Base & \bf Mini Base \\
    \midrule
    arb\_Arab & 9.1 & 4.8 & 13.9 & 11.4 & 7.1 & 13.1 & 14.9 & \bf 15.7 & 11.5 & 3.2 & \underline{15.0} \\
    ben\_Beng & 6.4 & 2.8 & 16.0 & 12.7 & 9.5 & 1.0 & \bf 18.4 & 16.6 & 12.6 & 3.0 & \underline{17.8} \\
    ces\_Latn & 14.4 & 9.4 & 23.7 & 18.8 & 13.9 & 7.0 & \underline{27.2} & \bf 28.6 & 21.8 & 6.0 & 26.0 \\
    deu\_Latn & 23.1 & 24.5 & 28.8 & 24.5 & 20.2 & 28.9 & \underline{30.6} & 29.0 & 28.9 & 12.9 & \bf 32.0 \\
    ell\_Grek & 11.1 & 7.5 & 17.2 & 18.7 & 16.3 & 30.0 & \underline{35.4} & \bf 36.2 & 22.0 & 4.4 & 30.7 \\
    fra\_Latn & 32.4 & 28.5 & 34.3 & 31.8 & 29.0 & 38.0 & \underline{40.6} & 38.7 & 38.3 & 12.4 & \bf 41.6 \\
    heb\_Hebr & 8.0 & 7.7 & 18.5 & 15.9 & 10.9 & 3.4 & 24.4 & \underline{25.5} & 14.3 & 4.7 & \bf 25.8 \\
    hun\_Latn & 10.1 & 3.9 & 9.6 & 15.1 & 12.0 & 2.3 & \bf 22.3 & 19.7 & 16.7 & 4.0 & \underline{20.5} \\
    ind\_Latn & 23.3 & 17.7 & 23.0 & 28.0 & 18.5 & 15.8 & 33.1 & \underline{33.6} & 29.9 & 7.5 & \bf 33.7 \\
    ita\_Latn & 29.5 & 25.6 & 34.7 & 30.3 & 23.8 & 31.4 & \underline{37.0} & 34.2 & 34.3 & 12.1 & \bf 38.5 \\
    jpn\_Jpan & 12.0 & 8.2 & 14.6 & 10.7 & 6.6 & 12.8 & 15.4 & 12.6 & \underline{15.5} & 4.6 & \bf 18.1 \\
    kor\_Hang & 9.4 & 6.5 & 13.6 & 10.7 & 8.1 & 11.7 & 16.9 & \underline{17.8} & 15.8 & 5.1 & \bf 19.1 \\
    nld\_Latn & 22.2 & 18.9 & 29.4 & 25.6 & 20.5 & 16.6 & \underline{31.6} & 29.8 & 27.3 & 10.1 & \bf 34.2 \\
    pol\_Latn & 14.6 & 11.7 & 18.6 & 17.4 & 13.1 & 8.0 & \underline{22.6} & 21.1 & 18.9 & 5.9 & \bf 24.4 \\
    por\_Latn & 29.1 & 29.2 & 33.7 & 29.9 & 25.3 & 32.5 & \underline{35.2} & 33.1 & 33.8 & 12.0 & \bf 36.2 \\
    ron\_Latn & 22.0 & 22.1 & 28.6 & 30.0 & 20.8 & 9.3 & \bf 36.7 & 33.7 & 31.2 & 9.4 & \underline{36.6} \\
    rus\_Cyrl & 20.5 & 20.1 & 24.1 & 19.6 & 17.9 & 24.3 & \bf 26.5 & 24.5 & 24.2 & 8.3 & \underline{25.5} \\
    spa\_Latn & 33.2 & 34.7 & 34.5 & 33.5 & 33.3 & 38.5 & \underline{41.2} & 39.8 & 39.7 & 14.6 & \bf 42.0 \\
    tha\_Thai & 17.3 & 7.1 & 12.3 & 20.2 & 11.6 & 19.7 & \bf 27.0 & 20.6 & 23.6 & 5.1 & \underline{26.1} \\
    tur\_Latn & 10.8 & 4.1 & 8.0 & 12.8 & 8.6 & 3.5 & \bf 25.5 & 19.4 & 18.5 & 4.9 & \underline{24.2} \\
    ukr\_Cyrl & 13.4 & 10.1 & 21.0 & 19.4 & 15.3 & 9.2 & \bf 29.0 & 26.2 & 21.0 & 6.7 & \underline{28.5} \\
    urd\_Arab & 3.0 & 2.5 & 12.4 & 12.0 & 8.3 & 1.5 & \bf 22.3 & 18.0 & 10.2 & 2.5 & \underline{20.2} \\
    vie\_Latn & 26.6 & 16.2 & 22.4 & 26.1 & 19.7 & 21.4 & 31.3 & \underline{32.9} & 30.4 & 8.1 & \bf 33.4 \\
    zho\_Hans & 20.5 & 17.3 & 18.6 & 16.5 & 13.4 & 17.5 & 20.2 & 18.5 & \bf 22.9 & 7.6 & \underline{21.9} \\
    Avg & 17.6 & 14.2 & 21.3 & 20.5 & 16.0 & 16.6 & \underline{27.7} & 26.1 & 23.5 & 7.3 & \bf 28.0 \\
    \bottomrule
\end{tabular}}
\caption{Per-language model performance comparison on WMT24$++$ En-Xx translation direction.}
\label{tab:wmt24pp_en_xx_results}
\end{table}
\begin{table}
\setlength{\tabcolsep}{3pt}
\footnotesize
\setlength{\belowcaptionskip}{-0.35cm}
\centering
\resizebox{\linewidth}{!}{
\begin{tabular}{l|ccccccccccc}
\toprule
    \multirow{2}{*}[0ex]{\bf Language} & \bf Qwen3 & \bf Trinity & \bf Granite4 & \bf Marco & \bf Llama3.2 & \bf SmolLM3 & \bf Gemma3 & \bf Tiny-Aya & \bf Qwen3 & \bf Trinity & \bf Marco \\
    & \bf 1.7B Base & \bf Nano Base & \bf Tiny Base & \bf Nano Base & \bf 3B Base & \bf 3B Base & \bf 4B Base & \bf 3.35B Base & \bf 4B Base & \bf Mini Base & \bf Mini Base \\
    \midrule
    arb\_Arab & 17.1 & 15.2 & 18.4 & 15.1 & 18.4 & 19.7 & \bf 23.2 & \underline{22.8} & 21.3 & 6.3 & 21.6 \\
    ben\_Beng & 20.2 & 14.0 & 24.1 & 21.6 & 16.2 & 4.6 & \bf 30.7 & 27.7 & 27.1 & 7.3 & \underline{28.1} \\
    ces\_Latn & 30.4 & 30.3 & 34.2 & 30.0 & 31.7 & 25.3 & 35.4 & 34.9 & \underline{35.6} & 11.4 & \bf 36.9 \\
    deu\_Latn & 32.2 & 32.9 & 35.4 & 31.3 & 31.6 & 34.6 & 36.0 & 34.2 & \underline{36.1} & 12.0 & \bf 36.3 \\
    ell\_Grek & 31.9 & 29.4 & 35.1 & 33.0 & 36.0 & 39.8 & \bf 42.4 & 40.5 & 38.9 & 10.1 & \underline{41.1} \\
    fra\_Latn & 37.4 & 40.1 & 39.3 & 36.9 & 38.0 & 40.0 & \underline{41.0} & 40.1 & 40.5 & 15.1 & \bf 42.1 \\
    heb\_Hebr & 26.4 & 27.5 & 31.9 & 30.2 & 27.8 & 21.4 & \underline{35.2} & 34.6 & 32.0 & 9.5 & \bf 36.3 \\
    hun\_Latn & 24.8 & 22.3 & 27.3 & 24.3 & 26.1 & 16.5 & \underline{29.1} & 28.2 & \underline{29.1} & 9.4 & \bf 30.5 \\
    ind\_Latn & 30.3 & 30.7 & 32.9 & 30.8 & 32.4 & 29.7 & 34.6 & \underline{34.7} & 34.0 & 12.8 & \bf 35.8 \\
    ita\_Latn & 37.9 & 40.1 & 41.0 & 37.2 & 38.1 & 42.0 & \bf 43.5 & 42.1 & 42.3 & 14.6 & \underline{43.3} \\
    jpn\_Jpan & 18.9 & 19.5 & 18.9 & 17.4 & 18.4 & 19.5 & \bf 23.7 & 21.5 & \underline{23.2} & 6.2 & 23.1 \\
    kor\_Hang & 22.0 & 19.9 & 23.1 & 18.7 & 20.6 & 22.3 & \bf 26.9 & 25.5 & 25.8 & 6.6 & \underline{26.4} \\
    nld\_Latn & 33.3 & 33.6 & 35.3 & 33.2 & 33.0 & 31.6 & 35.6 & 37.1 & \underline{37.2} & 12.5 & \bf 38.8 \\
    pol\_Latn & 25.6 & 26.2 & 29.8 & 27.0 & 27.4 & 23.1 & \underline{32.1} & 30.4 & 30.3 & 10.5 & \bf 33.9 \\
    por\_Latn & 35.6 & 38.2 & \underline{39.7} & 35.5 & 36.5 & 36.3 & 39.3 & 39.3 & 39.6 & 14.2 & \bf 41.2 \\
    ron\_Latn & 34.0 & 37.4 & 38.4 & 36.3 & 36.8 & 32.4 & \bf 43.3 & 40.5 & 40.6 & 12.4 & \underline{42.7} \\
    rus\_Cyrl & 24.8 & 27.1 & 27.2 & 24.8 & 25.3 & 26.9 & \underline{29.7} & 27.7 & 28.5 & 10.7 & \bf 29.9 \\
    spa\_Latn & 39.4 & 41.1 & 41.2 & 38.1 & 41.7 & 42.5 & \underline{45.3} & 43.7 & 44.1 & 16.5 & \bf 45.6 \\
    tha\_Thai & 22.7 & 16.5 & 21.0 & 22.2 & 23.1 & 26.2 & \bf 29.2 & 27.6 & 27.8 & 7.6 & \underline{29.1} \\
    tur\_Latn & 24.1 & 21.1 & 25.8 & 23.5 & 24.6 & 16.5 & \underline{32.7} & 30.4 & 30.2 & 8.5 & \bf 32.8 \\
    ukr\_Cyrl & 29.6 & 27.8 & 33.1 & 29.4 & 28.3 & 28.0 & \underline{35.9} & 32.8 & 33.6 & 10.2 & \bf 36.9 \\
    urd\_Arab & 25.0 & 21.3 & 29.3 & 27.5 & 27.7 & 15.2 & \bf 36.7 & 34.1 & 32.0 & 8.1 & \underline{34.6} \\
    vie\_Latn & 27.2 & 25.8 & 27.7 & 27.7 & 27.3 & 26.2 & \underline{31.7} & 31.3 & 31.5 & 9.7 & \bf 32.8 \\
    zho\_Hans & 25.2 & 26.5 & 26.8 & 23.1 & 25.1 & 26.0 & 28.0 & 26.2 & \underline{29.9} & 8.9 & \bf 30.1 \\
    Avg & 28.2 & 27.7 & 30.7 & 28.1 & 28.8 & 26.9 & \underline{34.2} & 32.8 & 33.0 & 10.5 & \bf 34.6 \\
    \bottomrule
\end{tabular}}
\caption{Per-language model performance comparison on WMT24$++$ Xx-En translation direction.}
\label{tab:wmt24pp_xx_en_results}
\end{table}

\end{document}